\documentclass{article}
\usepackage{amsmath}

\PassOptionsToPackage{numbers}{natbib}

\usepackage[preprint]{neurips_2026}

\usepackage{array}
\usepackage{tabularx}

\usepackage{xcolor}
\colorlet{gray}{gray!75!black}

\title{Chem-PerturBridge: a harmonized compendium of small molecule perturbation transcriptomic effects}

\author{%
  Artur Szalata$^{1,2,*}$, Olga Novitskaia$^{1,*}$, Maiia Shulman$^{1,2}$, Matthew Mella, \\
  \textbf{Altynbek Zhubanchaliyev$^{3,4,5}$, Fabian J. Theis$^{1,2,6,\dag}$}\\
  \\
  $^1$Institute of Computational Biology, Helmholtz Center Munich, $^2$TUM School of Life Sciences\\
  Weihenstephan, Technical University of Munich,\\$^3$Institut Curie, INSERM U1331, Computational Oncology\\
  $^4$Centre National de Recherche en G\'enomique Humaine (CNRGH), CEA,\\
  Universit\'e Paris-Saclay, \'Evry, France, $^5$Universit\'e Paris Cit\'e, Paris, France,\\
  $^6$TUM School of Computation, Information and Technology, Technical University of Munich,\\
  $^{*}$Equal contribution, $^{\dag}$Corresponding author\\
  \texttt{fabian.theis@helmholtz-munich.de}
}

\begin{document}
\raggedbottom

\maketitle

\begin{abstract}
Large perturbation models require training data encompassing chemical, cellular, and assay diversity. Current transcriptomic resources for small-molecule modeling, however, are fragmented across technologies, metadata conventions, controls, doses, and preprocessing pipelines. We introduce Chem-PerturBridge, a harmonized multi-dataset resource comprising over 37k compounds, 136 cellular contexts, and 1.25M transcriptomic samples across eight assay types, with standardized identifiers, metadata, and replicate-aware condition-level effects. We use the resource to evaluate matched-condition agreement across datasets and replicate agreement within datasets. Matched same-compound conditions show weak absolute agreement in fine-grained $\mathrm{logFC}$ rankings and magnitudes across most dataset pairs. However, they generally exceed individual same-context different-compound signatures, while often falling below the smoothed centroid of such signatures. Same-compound retrieval is above chance in all pairs. $\mathrm{LogFC}$ direction agreement is high in absolute terms, and exceeds both baselines in most cases. We further evaluate Chem-PerturBridge as a pretraining resource for compound representation learning. Under a compound-held-out OP3 evaluation split, embeddings pretrained on Chem-PerturBridge improve over L1000-only embeddings, Morgan fingerprints, chemical language model embeddings, and the descriptor-free OP3 baseline across metrics. An extensive molecule-holdout evaluation across 11 datasets further shows that models trained on Chem-PerturBridge outperform or match those that are not. Chem-PerturBridge therefore supports both diagnostic evaluation of cross-dataset signature agreement and model-oriented reuse of heterogeneous perturbation transcriptomic data.
\end{abstract}

\section{Introduction}

Chemical perturbation profiling is a central experimental approach for connecting small molecules to cellular mechanisms, toxicity, and therapeutic response~\cite{lamb_connectivity_2006, subramanian_next_2017, keenan_library_2018}. Transcriptomic readouts are scalable, high-dimensional, and interpretable at the gene and pathway level. Transcriptomic perturbation datasets prompted the development of a growing class of perturbation models that primarily aim to predict cellular responses to unseen conditions, generalize across cellular contexts, and learn reusable representations~\cite{lotfollahi_scgen_2019, lotfollahi_predicting_2023, wu_perturbench_2025, miladinovic_silico_2025, ji_scalable_2025}. Such models benefit from large-scale training data spanning chemical, cellular, and assay diversity.

Current perturbation transcriptomic resources provide this diversity only in a fragmented form. Public and semi-public datasets differ in profiling technologies, measured genes, control design, doses, timepoints, cell systems, replicate structure, metadata quality and format, and preprocessing pipelines. Bulk RNA-seq~\cite{mortazavi_mapping_2008}, LINCS L1000~\cite{subramanian_next_2017, peck_method_2006}, DRUG-seq screens~\cite{ye_drug-seq_2018}, targeted count assays~\cite{xiang_high-throughput_2025}, and single-cell perturbation datasets~\cite{srivatsan_massively_2020} each provide useful but assay-dependent views of chemical response. Simple overlap in compound names or cell-line labels therefore does not imply that two resources measure comparable perturbation effects.

This creates two problems for perturbation modeling. The first is diagnostic: when two datasets contain the same compound in the same context, do the measured transcriptomic effects agree after harmonization? The second is constructive: can multi-dataset training extract a signal that improves downstream predictions across datasets? A harmonized resource should enable cross-dataset agreement to be measured under matched conditions, and it should provide a substrate for pretraining models that can learn across assay and context shifts.

We introduce Chem-PerturBridge, a harmonized multi-dataset resource designed to facilitate the training and evaluation of small-molecule transcriptomic perturbation models. Chem-PerturBridge integrates over 37k compounds, 136 cellular contexts, and over 1.25M transcriptomic samples across bulk RNA-seq, LINCS L1000, DRUG-seq screens, targeted count assays, and pseudobulked single-cell data. Chem-PerturBridge standardizes compounds, cellular contexts, doses, timepoints, controls, replicate labels, assays, and gene identifiers, and constructs condition-level perturbation effects using a shared replicate-aware layer and differential gene expression analysis~\cite{ritchie_limma_2015}.

To summarize, this paper makes three contributions:
\begin{itemize}
    \item \textbf{A model-ready harmonized perturbation resource.}
    We release a unified representation of small-molecule transcriptomic perturbation data with standardized compound, context, dose, timepoint, control, replicate, assay, and gene metadata, together with condition-level differential-effect tables and reproducible processing code.
    \item \textbf{A calibrated agreement benchmark.}
    We define a matched-condition benchmark for same-compound transcriptomic agreement across datasets and a replicate-agreement benchmark within datasets, both calibrated by same-context different-compound smoothed centroid and individual peer baselines.
    \item \textbf{A representation-learning utility evaluation.}
    We test whether embeddings pretrained on Chem-PerturBridge improve perturbation prediction for held-out compounds in OP3, relative to L1000-only embeddings, molecular fingerprints, chemical language model embeddings~\cite{chithrananda_chemberta_2020, ahmad_chemberta2_2022, ross_molformer_2022}, and descriptor-free compound encodings. We also evaluate direct multi-output LPM~\cite{miladinovic_silico_2025} prediction under molecule-holdout splits across all source datasets, comparing all-data training, target-dataset-only training, fine-tuning, fixed Morgan fingerprints, and Morgan-initialized molecule embeddings.
\end{itemize}

The main takeaway is that raw perturbation signatures are not directly interchangeable across assay families: fine-grained $\mathrm{logFC}$ agreement is weak, whereas direction agreement is more stable. At the same time, pooled training over the harmonized resource can recover transferable compound-level signal for downstream prediction, especially when cross-dataset training is combined with molecular priors and target-domain adaptation.

\section{Related work}

\paragraph{Harmonized perturbation-signature resources.}
Multiple initiatives have curated, reprocessed, or standardized public perturbation studies. iLINCS~\cite{pilarczyk_connecting_2022} aggregates processed omics datasets and precomputed signatures into an interactive analysis platform; however, it covers fewer than half as many compounds as Chem-PerturBridge and does not assess matched cross-dataset agreement or support model training. CREEDS~\cite{wang_extraction_2016} curates GEO-derived~\cite{barrett_ncbi_2013} metadata but omits recent large perturbation studies~\cite{zhang_tahoe-100m_2025, xiang_high-throughput_2025}. SigCom LINCS~\cite{evangelista_sigcom_2022} provides access to over one million perturbation signatures through a search interface, primarily from L1000, microarray, and bulk RNA-seq platforms, but lacks single-cell perturbation data. ToxicoDB~\cite{nair_toxicodb_2020} integrates toxicogenomics profiles, though its biological and assay scope is limited. PharmOmics~\cite{chen_pharmomics_2022} curates species- and tissue-specific drug signatures, yet includes only 941 chemicals. Lamin pertdata~\cite{sunny_pertdata_2026} harmonized a large collection of diverse perturbation datasets, but does not provide the post-processing necessary for cross-dataset evaluations. scPerturb~\cite{peidli_scperturb_2024} advances perturbation-modeling applications by harmonizing single-cell perturbation datasets and defining perturbation-effect statistics; however, it remains restricted to single-cell data and does not integrate large chemical perturbation resources across assay technologies. Chem-PerturBridge is distinct in harmonizing recent single-cell perturbation studies with large-scale L1000, DRUG-seq, targeted count, and bulk RNA-seq perturbation datasets. This integration enabled a cross-dataset agreement study covering multiple assay families and facilitated case studies demonstrating the corpus's utility in perturbation modeling.

\paragraph{Cross-dataset concordance and intra-dataset reproducibility.}
Large-scale transcriptomic benchmarking initiatives, including MAQC~\cite{shi_microarray_2006} and SEQC~\cite{su_comprehensive_2014}, have demonstrated that gene-expression measurements are highly reproducible across sites and platforms when using controlled reference-sample designs. In SEQC, identical RNA mixtures were distributed to multiple laboratories and sequencing technologies, enabling performance assessment based on engineered titration relationships and A-versus-A null comparisons. These studies confirm reproducibility for relative expression and filtered differential-expression calls, but also reveal limitations in absolute quantification due to library-construction and protocol-specific effects. In contrast, cross-dataset perturbation benchmarking presents greater challenges because perturbation datasets are generated independently in different laboratories. Even after matching for compound, cellular context, timepoint, and dose, residual variation remains, including differences in control design, operator technique, and assay implementation.

The LINCS L1000 project addressed replicate variability within a study through reproducibility-aware signature construction, including replicate-correlation-weighted MODZ consensus profiles, signature strength, and transcriptional activity scores~\cite{subramanian_next_2017}. Additional harmonization initiatives have targeted other aspects of perturbation-data integration. For example, scPerturb~\cite{peidli_scperturb_2024} harmonized public single-cell perturbation-response datasets and introduced E-statistics and E-distance to quantify perturbation effects and similarity prior to downstream modeling. In toxicogenomics, Sutherland et al.\ compared drug-induced transcriptional responses across DrugMatrix~\cite{ganter_development_2005}, Open TG-GATEs~\cite{igarashi_open_2015}, and GEO-derived liver and hepatocyte systems, finding that concordance is often more robust at the level of co-expression networks than at individual genes or gene sets~\cite{sutherland_assessing_2016}. Lim and Pavlidis directly evaluated the reproducibility of the
Connectivity Map (CMap)~\cite{lamb_connectivity_2006,
subramanian_next_2017} and found weak differential-expression
reproducibility both across CMap versions and within each version, with within-dataset replicate agreement falling below MAQC cross-laboratory benchmarks; reproducibility was associated with differential-expression strength, which itself was driven by the compound concentration and cell-line responsiveness. Additionally, they compared retrieval against a self-query control, in which CMap 2 was queried with its own profiles, serving as an upper bound on retrieval performance~\cite{lim_evaluation_2021}. Recent modeling approaches align L1000 and bulk RNA-seq perturbation data within shared latent spaces to facilitate transferability across readouts~\cite{li_aethercell_2026}. These findings underscore the importance of quantifying signal preservation across assay types. Collectively, these studies support the adoption of replicate-based calibration, context-aware baselines, and rigorous matched-condition evaluation in practical applications.

\paragraph{Position relative to existing resources.}
Chem-PerturBridge occupies a distinct position among perturbation-signature
resources. To our knowledge, it is the largest multi-source, multi-assay
small-molecule transcriptomic perturbation compendium by harmonized compound
coverage, while also providing a unified processing pipeline and
condition-level effect representation. Existing resources offer either very
large single-assay collections, broad signature-search interfaces, curated
toxicogenomic or drug-signature databases, or harmonized single-cell
perturbation datasets. Chem-PerturBridge differs by bringing broad chemical
coverage, cross-assay transcriptomic harmonization, reproducible effect
construction, and matched-condition cross-dataset evaluation into a single
model-ready resource. Relative to prior reproducibility evaluations of Affymetrix and L1000 within CMap~\cite{lim_evaluation_2021}, we extend agreement and retrieval analyses across multiple assay families, compare results against two kinds of same-context different-compound baselines (centroids and individual peers), and provide a model-ready data release.

\section{Resource design and scope}

\subsection{Design principles}

Chem-PerturBridge is designed to support two complementary uses. The first is evaluation: the resource provides matched-condition benchmarks for asking whether observed perturbation effects agree across datasets, as well as across replicates within a dataset. The second is pretraining: the same harmonized representation can be used as a diverse training set for perturbation models.

We follow four design principles to support these and other related applications.
\textbf{Explicit identifiers.} Compounds, cell contexts, genes, doses, timepoints, datasets, assays, plates, wells are represented in standardized fields, while source-specific names are retained for completeness.
\textbf{Replicate-aware effects.} We generate two sets of perturbation signatures: one estimated from sets of treated and control measurements and one from individual profiles of treated and sets of control measurements. The latter is used for studying replicate consistency.
\textbf{Assay-specific processing.} Count-based data and normalized L1000 profiles share a downstream limma contrast layer, but they do not share normalization steps.
\textbf{Calibrated evaluation.} Same-compound agreement is interpreted relative to same-context different-compound baselines and within-dataset replicate agreement.

\subsection{Datasets included}

Chem-PerturBridge integrates public and semi-public chemical perturbation resources spanning eight assay types. Table~\ref{tab:cpb_datasets} reports the current scope and Figure~\ref{fig:cpb_overlap} the cross-dataset overlap. The resource covers over 37k harmonized compounds, 136 cellular contexts, and over 1.25M transcriptomic samples after union-level deduplication across selected sources. The release is modular. Redistributable harmonized assets are hosted, while components under strict source licenses are documented and accompanied by reproduction scripts for users with source access.

\begin{table*}[t]
\caption{Overview of datasets included in Chem-PerturBridge. Per-dataset counts report harmonized dataset-level records. The Chem-PerturBridge row reports the union across selected datasets after harmonization; compounds (using the \texttt{pubchem\_cid} column) and contexts are deduplicated across datasets, and sample counts exclude controls. Dose ranges summarize non-zero recorded doses for non-control treatments.}
\label{tab:cpb_datasets}
\centering
\footnotesize
\setlength{\tabcolsep}{2pt}
\begin{tabular}{p{2.5cm} p{3.3cm} r r r p{1.4cm} p{1.8cm}}
\toprule
Dataset & Technology & \# compounds & \# contexts & \# samples & Timepoints (h) & Dose range ($\mu$M) \\
\midrule
L1000 Phase I~\cite{subramanian_next_2017, peck_method_2006} & LINCS L1000 assay & 20{,}212 & 70 & 699{,}276 & 6, 24, 48 & 0.0003--100000 \\
L1000 Phase II~\cite{subramanian_next_2017, peck_method_2006} & LINCS L1000 assay & 1{,}792 & 30 & 333{,}680 & 3, 6, 24 & 0.0041--40 \\
Tahoe-100M~\cite{zhang_tahoe-100m_2025} & Tahoe Mosaic / Parse Evercode Whole Transcriptome v3 & 379 & 50 & 67{,}018 & 24 & 0.05, 0.5, 5 \\
CIGS-MCE~\cite{xiang_high-throughput_2025} & HTS2~\cite{li_versatile_2012} & 11{,}261 & 2 & 63{,}722 & 24 & 10 \\
Novartis/DRUG-seq U2OS~\cite{hadjikyriacou_2025_14291446} & DRUG-seq~\cite{ye_drug-seq_2018} & 4{,}340 & 1 & 53{,}012 & 24 & 0.01, 0.1, 1, 10 \\
VCPI-0001~\cite{vcpi_0001_portal} & DRUG-seq~\cite{ye_drug-seq_2018} & 2{,}277 & 1 & 27{,}577 & 24 & 0.03--10 \\
CIGS-TCM~\cite{xiang_high-throughput_2025} & HiMAP-seq~\cite{xiang_high-throughput_2025} & 1{,}856 & 2 & 22{,}545 & 24 & 10, 20 \\
VCPI-0002~\cite{vcpi_0002_portal} & DRUG-seq~\cite{ye_drug-seq_2018} & 1{,}493 & 1 & 18{,}197 & 24 & 0.03--10 \\
GDPx2~\cite{baugh_mapping_2025} & DRUG-seq~\cite{ye_drug-seq_2018} & 86 & 4 & 9{,}504 & 24 & 0.0095--3 \\
sci-Plex~\cite{srivatsan_massively_2020} & sci-Plex single-cell RNA-seq & 187 & 3 & 4{,}974 & 24, 72 & 0.01, 0.1, 1, 10 \\
DILImap~\cite{bergen_large-scale_2025} & SMART-seq~\cite{zhu_reverse_2001} DE3~\cite{bergen_large-scale_2025} & 301 & 1 & 4{,}148 & 24 & 0.0037--1000 \\
OP3~\cite{szalata_benchmark_2024} & 10x Genomics Chromium Single Cell 3' v3.1 RNA-seq~\cite{zheng_massively_2017} & 138 & 4 & 1{,}813 & 24 & 1 \\
\midrule
\textbf{Chem-PerturBridge} & \textbf{all above} & \textbf{37{,}463} & \textbf{136} & \textbf{1{,}251{,}919} & \textbf{3, 6, 24, 48, 72} & \textbf{0.0003--100000} \\
\bottomrule
\end{tabular}
\end{table*}

\subsection{Data and code availability}

Harmonized data release is available at \url{https://huggingface.co/datasets/theislab/chem-perturbridge}. The release contains harmonized AnnData objects for redistributable sources,
condition-level differential-effect tables, matched-condition benchmark files,
model-training split files, a Croissant metadata file, and a license manifest.
The accompanying code repositories are:
(i) the harmonization and effect-construction pipeline at
\url{https://github.com/theislab/Chem-PerturBridge};
(ii) the cross-dataset and replicate agreement analyses at
\url{https://github.com/theislab/Chem-PerturBridge_analysis};
(iii) the representation-learning and molecule-holdout LPM code at
\url{https://github.com/theislab/chem-perturbridge_lpm}; and
(iv) the OP3 perturbation-prediction evaluation code at
\url{https://github.com/theislab/op3_signatures}.

\section{Harmonization and effect construction}

Figure~\ref{fig:data_processing} summarizes the harmonization workflow. All datasets are converted to a strict AnnData-based schema with standardized fields. The schema builds on Lamin pertdata~\cite{sunny_pertdata_2026}; for some datasets, metadata annotations are obtained from the same resource. Compounds are mapped to \texttt{pubchem\_cid} where possible using source identifiers, structures, CAS numbers, names, cached PubChem lookup, and manual resolution for cases that cannot be automatically mapped. Cellular contexts are standardized to Cellosaurus or Cell Ontology identifiers when available. Timepoints are represented in hours, doses are converted to micromolar units, and genes are indexed by Ensembl 115~\cite{dyer_ensembl_2025} identifiers where possible, while retaining gene symbols.

Processing differs by input modality. Tahoe-100M and sci-Plex are standardized at the cell level, filtered, and aggregated into pseudobulk count profiles before differential analysis. Count-based sample-level or pre-aggregated resources, including OP3, CIGS, Novartis DRUG-seq, VCPI, GDPx2, and DILImap, enter the common pseudobulk schema after source-specific quality control and metadata standardization. L1000 Phase I/II Level 3 profiles are assembled from normalized GCTX matrices and restricted to landmark genes by default; they therefore use a normalized-input limma branch rather than count normalization and voom.

For count-based datasets, perturbation effects are estimated with a shared edgeR~\cite{robinson_edger_2010}/limma workflow using gene filtering, TMM normalization, voom transformation, and moderated contrasts. For normalized L1000 profiles, the same limma contrast is fit directly to the normalized values. This yields gene-level log fold-changes, moderated \(t\)-statistics, \(p\)-values and adjusted \(p\)-values for grouped perturbation conditions. Appendix~\ref{app:representations} describes the further representations derived from these outputs, including per-gene population $z$-scores and $\mathrm{ashr}$ posterior means~\cite{stephens_ashr_2017}.

\section{Signature agreement evaluation}\label{sec:signature_eval}
\paragraph{Two evaluation objectives.}
Chem-PerturBridge performs two complementary evaluations.
This section evaluates \textbf{signature agreement} at two levels:
across datasets and within datasets across replicates. The cross-dataset
analysis asks whether the measured perturbation signature for a compound in
one dataset agrees with the signature of the same compound in another dataset
under matched cell context, time, and local dose. The within-dataset replicate
analysis asks the same question across technical and biological replicates of
the same condition within a single dataset, characterizing technical variation in each
assay family. Both are calibrated against two types of same-context,
different-compound baselines: the smoothed centroid of signatures and the individual peer signatures (source and target). The centroid baseline is reported from the source side. Section~\ref{sec:repr} evaluates \textbf{representation transfer}:
whether pooling harmonized data during pretraining yields compound embeddings
that improve unseen-compound prediction on external benchmarks. This reflects
how well the harmonized compendium serves as training data for
multi-dataset perturbation models.

\subsection{Perturbation-effect representation}

We study perturbation effects derived from bulk or pseudobulk transcriptomic measurements. A perturbation condition is indexed by
\(\xi = (a,p,c,d,t) \in \mathcal{A}\times\mathcal{P}_{\mathrm{pert}}\times\mathcal{C}\times\mathcal{D}\times\mathcal{T}\),
where \(a\) denotes dataset or assay domain, \(p\) perturbation identity, \(c\) cellular context, \(d\) dose, and \(t\) timepoint. For each condition we compare replicate-aware differential summaries
\(r_\xi = \Psi(Y_\xi^{\mathrm{pert}}, Y_\xi^{\mathrm{ctrl}})\),
where \(\Psi\) is the shared limma-based differential-expression layer applied to the corresponding sets of treated and control replicate measurements (samples for bulk RNA-seq, replicate-level pseudobulk profiles for single-cell data). For each gene \(g\), this yields a log fold-change \(L_{\xi,g}\) and moderated \(t\)-statistic \(T_{\xi,g}\). The benchmark compares these summaries rather than individual post-perturbation profiles. Appendix~\ref{app:formal_problem} gives the full formal setup~\cite{kovacevic_no_2025}, including the latent-state and assay observation operator framing.

The main representations used in the analyses are \(L_{\xi,g}\), per-gene population \(z\)-scores of \(L_{\xi,g}\), moderated \(t\)-statistics \(T_{\xi,g}\). We also evaluate $\mathrm{sign}(L_{\xi,g})$ on the retrieval task; signed significance, $\mathrm{ashr}$ posterior-shrunken $\mathrm{logFC}$, and the $p$-value are evaluated in a separate cross-dataset retrieval sweep (Appendix~\ref{app:representation_sweep}); Appendix~\ref{app:representations} defines each. We report agreement and retrieval under several representations rather than committing to one, since the choice affects both sets of results.

\subsection{Matched-condition construction}

The primary evaluation asks whether the same compound produces similar observed perturbation effects across resources when cell context, timepoint, and local dose are matched. Non-control differential-expression rows are indexed by normalized PubChem~\cite{kim_pubchem_2025} CID, harmonized \texttt{cell\_type}, time in hours, and positive dose in micromolar units. Two rows from datasets \(A\) and \(B\) are candidate matches only when compound, context, and time match exactly. Dose is then aligned by mutual nearest neighbors in \(\log_{10}\) dose, and a pair is retained only when \(|\Delta\log_{10} d|\leq 1\). Dataset-pair context strata are kept only if they contain at least 10 matched compounds. This filter prevents isolated one-off overlaps from driving agreement estimates and makes our baselines more reliable.

Agreement is measured on the overlap of the genes available in compared rows. We report whole-signature Spearman correlations on limma log fold-changes, moderated \(t\)-statistics and per-gene population \(z\)-scores; DEG-focused Spearman on the same representation families, and signed direction agreement using adjusted \(p<0.05\) genes; and retrieval statistics that ask whether the true same-compound partner can be identified among same-context target-side alternatives, using cosine similarity and negative Euclidean distance.

\subsection{Calibration by baselines and replicates}

A same-compound score is difficult to interpret without calibration because cell context, assay family, and batch structure can induce shared patterns. We therefore compare each matched condition to same-cellular-context, same-time, same-dose signatures from other compounds. Two references are used. A centroid baseline, formed by averaging other compounds in the same stratum, is retained as a hard competitor, while the individual peer baseline, formed by scoring the signature of each other compound in the stratum separately and averaging the resulting scores, is primary. Both are formed from the query's own dataset (source side) and from the partner dataset (target side); we report the centroid from the source side and peers from both sides. The baseline-adjusted scores subtract the scores of these same-context different-compound references from the observed same-compound agreement.

For within-dataset replicate agreement, replicate pairs are compared for the same compound-context-time-dose condition and evaluated against references of the same two kinds, formed within the dataset. Of these two references, only the centroid baseline is used for replicate retrieval.

\section{Representation-learning evaluation}\label{sec:repr}

Chem-PerturBridge is also designed as a training dataset for perturbation modeling. The representation-learning evaluation asks whether pooling heterogeneous, harmonized data can learn compound embeddings that transfer across datasets, to compounds withheld from the target dataset but observed elsewhere. We report two evaluations of this use.

The first supplies these embeddings to a downstream task. We train LPM-style~\cite{miladinovic_silico_2025} representation learners on different subsets of the
harmonized resource, comparing an L1000 model against the models trained on the Chem-PerturBridge dataset and on a version narrowed to L1000 compounds. Their small-molecule embeddings are fed to an OP3 downstream predictor, alongside Morgan fingerprints, embeddings from pretrained chemical language models~\cite{chithrananda_chemberta_2020, ahmad_chemberta2_2022, ross_molformer_2022}, and the descriptor-free baseline. The OP3 dataset is excluded from the corpus here to avoid ambiguity between resource construction and downstream evaluation. Appendix~\ref{app:op3_evaluation} reports the evaluation details and results.

The second trains the multi-output LPM directly, rather than supplying embeddings to a separate predictor. This experiment asks a complementary question: whether models trained on the harmonized resource improve direct perturbation effect prediction when molecules are held out within each target dataset, while held-out molecules remain represented in at least one other dataset. Table~\ref{tab:expanded_lpm_test_rmse} reports this evaluation across all 11 datasets, using molecule-disjoint splits over shared compounds and 10 seeds. We compare models trained under 8 different settings, which we describe in Table~\ref{tab:expanded_lpm_variants}.

\section{Results}

\begin{figure*}[!t]
\centering
\includegraphics[width=0.8\textwidth]{figures/retrieval_upd.pdf}
\caption{\textbf{$\mathbf{logFC}$ agreement and retrieval across Chem-PerturBridge.}
DEG-restricted $\mathrm{logFC}$ correlation is generally weak in absolute terms and often falls below the within-dataset same-context different-compound centroid, but exceeds individual peer baselines in most pairs. Cross-dataset $\mathrm{logFC}$ direction agreement is substantially stronger and usually outperforms all baselines. Cross-dataset retrieval ranks the matched compound above the target-peer baseline in all pairs, but below the averaged centroid, which itself ranks highly. Retrieval on signed directions is close to magnitudes (mean rank 0.67 vs. 0.68), exceeding the target-peer baseline in every pair. Each panel is a dataset-by-dataset matrix over the resources in Table~\ref{tab:cpb_datasets}. \textbf{a--d)} DEG-restricted (adj.\,$p < 0.05$) gene-wise Spearman correlation of log fold-changes. \textbf{e--h)}
Signed direction agreement on genes differentially expressed in both signatures. \textbf{i--l)} Cross-dataset
$\mathrm{logFC}$ retrieval of the matched partner against same-context different-compound distractors, under cosine similarity. \textbf{m--p)} The same retrieval on signed directions, $\mathrm{sign}(\mathrm{logFC})$. Retrieval is reported as normalized best-positive rank, where higher is better; 1.0 is the best. Within each row, panels report the cross-dataset score and the remaining three baselines: the source-side centroid baseline, individual source peers and individual target peers. Perturbagen-clustered bootstrap confidence intervals for
panels a--p are reported in
Tables~\ref{tab:cross_dataset_deg_lfc_spearman_bootstrap_ci},
\ref{tab:cross_dataset_direction_agreement_bootstrap_ci}, and
\ref{tab:cross_dataset_retrieval_cosine_bootstrap_ci}.}
\label{fig:cpb_overview}
\end{figure*}

\subsection{DEG-restricted and all-gene $\mathbf{logFC}$ agreements are weak across most dataset pairs}

We first report signature-level results on cross-dataset and replicate
agreement (Figures~\ref{fig:cpb_overview},~\ref{fig:cpb_overview_l2} and~\ref{fig:cpb_replicate}),
then representation-level results on cross-dataset compound transfer
(Table~\ref{tab:expanded_lpm_test_rmse}). Absolute agreement is generally weak across most dataset pairs, and whether same-compound cross-dataset scores fall below a baseline depends on which baseline is used. Against the within-dataset same-context different-compound centroid, the baseline-adjusted interval is below zero in every dataset pair, for both DEG-restricted and all-gene $\mathrm{logFC}$ Spearman (Tables~\ref{tab:cross_dataset_deg_lfc_spearman_bootstrap_ci} and~\ref{tab:cross_dataset_all_gene_logfc_spearman_bootstrap_ci}), so we treat the centroid as
a hard comparator. Comparisons against individual peers are more favorable, particularly against target peers. Standardizing each gene within a dataset shrinks the gap to the centroid in every dataset pair while the observed score moves much less. Weak absolute agreement, together with the centroid outscoring matched pairs, indicates that context-level, assay-level, or processing-induced structure can dominate rank or magnitude agreement among DEGs, while the positive margins against individual peers indicate that a compound-specific component persists on top of it.

A complementary analysis shows that cross-dataset agreement increases with
perturbation strength, measured by the mean absolute moderated $t$-statistic
across matched datasets, but with substantial residual variability
(Appendix~\ref{app:perturbation_strength_agreement},
Figure~\ref{fig:strength_vs_similarity}). This extends the strength-reproducibility association reported for Affymetrix-vs-L1000 CMap comparisons~\cite{lim_evaluation_2021} to the cross-assay setting.

\subsection{$\mathbf{LogFC}$ direction agreement is substantially more stable}

The weak $\mathrm{logFC}$ correlation results contrast with the direction-agreement results. Across the same matched conditions, $\mathrm{logFC}$ direction agreement is generally strong in absolute terms and usually outperforms both kinds of baselines. The perturbagen-clustered intervals show positive baseline-adjusted direction agreement for most dataset pairs, with weaker or negative margins in some Tahoe comparisons (Table~\ref{tab:cross_dataset_direction_agreement_bootstrap_ci}). This suggests that the sign of response among differentially expressed genes is more transferable than fine-grained $\mathrm{logFC}$ ranking or magnitude. The contrast between magnitude and direction is specific to agreement, both cross- and within-dataset. Under cosine similarity, which is scale-invariant, observed magnitude-based and direction-based retrieval perform comparably (Table~\ref{tab:cross_dataset_retrieval_cosine_bootstrap_ci}). For agreement-based evaluation, the distinction matters for benchmark design. A model or metric that is evaluated on signed response agreement may see meaningful cross-dataset signal, while a metric that requires precise $\mathrm{logFC}$ ranking may conclude that the same pair of datasets is poorly aligned.

\subsection{Retrieval signal is present but depends on context and similarity metric}

Retrieval-style evaluation asks whether a query signature can identify its
same-compound partner among same-context target-side alternatives. Under cosine, which is scale-invariant, magnitude-based and direction-based retrieval yield similar mean observed and baseline-adjusted scores; individual pairs differ by at most 0.045 in either direction for observed scores (Figure~\ref{fig:cpb_overview}, i and m).

Under negative $L_2$ (Euclidean) distance, cross-dataset retrieval on $\mathrm{logFC}$ places the matched partner above the target-peer baseline in most pairs, below the same-context centroid in every pair, and mixed against the source peer (Figure~\ref{fig:cpb_overview_l2}a--d). The centroid advantage follows from its own near-perfect rank, which averaging makes almost unbeatable. Signed directions outperform $\mathrm{logFC}$ in every pair for the observed score (by 0.035 to 0.239) and against the centroid and target peer, but not consistently against the source peer (Figure~\ref{fig:cpb_overview_l2}e--h, Table~\ref{tab:cross_dataset_retrieval_logfc_bootstrap_ci}).

The advantage of direction over magnitude is specific to the distance-based score, which is sensitive to signature scale while sign vectors are scale-free by construction. Thus, compound-specific retrieval signal is present, but context-level, assay-level, and dataset-pair structure remain strong confounders. Perturbagen-clustered bootstrap intervals for cross-dataset retrieval with cosine similarity and negative $L_2$ are reported in 
Tables~\ref{tab:cross_dataset_retrieval_cosine_bootstrap_ci} and~\ref{tab:cross_dataset_retrieval_logfc_bootstrap_ci}.

\subsection{Replicate agreement exposes limits of representation-dependent reproducibility}

Figure~\ref{fig:cpb_replicate} shows that DEG-restricted $\mathrm{logFC}$ Spearman agreement can fall below the same-context different-compound centroid baseline even within a single dataset. This means that some commonly used DEG-restricted $\mathrm{logFC}$ correlation metrics provide a low-reliability target for cross-dataset benchmarking. When the replicate-derived effect representation is already near or below a context-level baseline within a dataset, cross-dataset agreement on the same representation cannot be expected to be strong.

Direction agreement provides a contrasting internal reference on the centroid baseline: it is high in absolute terms and exceeds the centroid baseline across nearly all assays. The representation has a similar effect: the number of datasets with a centroid-adjusted interval above zero increases from 4 of 12 under $\mathrm{logFC}$ to 7 under moderated $t$ and 8 under cell-line $z$-scoring. DEG-restricted agreement exceeds the peer baseline in all 12 datasets under every representation, so selecting the comparator affects the baseline-adjusted scores. For all-gene agreement the counts change, and the peer-adjusted interval is above zero in 10 of 12 under every representation.

Together, the replicate analyses indicate that the apparent degree of reproducibility depends on the metric and effect representation, largely relative to the centroid; against individual peers DEG-restricted agreement is positive and all-gene agreement is mostly positive under every representation. Perturbagen-clustered bootstrap intervals for the replicate agreement summaries in Figure~\ref{fig:cpb_replicate}, and for their all-gene counterpart, are reported in Appendix~\ref{app:bootstrap_uncertainty} (Tables~\ref{tab:replicate_deg_lfc_spearman_bootstrap_ci},~\ref{tab:replicate_direction_agreement_bootstrap_ci}, and~\ref{tab:replicate_logfc_bootstrap_ci}).

\subsection{Multi-dataset training improves held-out-compound prediction}
Detailed evaluation strategies, including coverage-matched, step-matched and ablation studies, together with results for the OP3 downstream task, are described in Appendix~\ref{app:op3_evaluation} and Tables~\ref{tab:op3_unseen_compounds},~\ref{tab:op3_unseen_compounds_b},~\ref{tab:op3_ablation_covariates},~\ref{tab:op3_bootstrap},~\ref{tab:op3_bootstrap_b},~\ref{tab:op3_ablation_bootstrap}. Overall, pooled training over Chem-PerturBridge yields compound representations that lead on most metrics across multiple setups in the compound-held-out
OP3 evaluation, but part of the gain over the L1000-only model reflects differences in compound coverage. Together with the agreement analyses, these results still indicate that heterogeneous perturbation data can provide transferable compound-level information.

The molecule-holdout LPM experiments in Table~\ref{tab:expanded_lpm_test_rmse} support this conclusion while showing that transfer is dataset-dependent. Across the 11 datasets with test splits, the strategy that initializes molecule embeddings from Morgan fingerprints, trains on Chem-PerturBridge, and then fine-tunes on the target dataset with molecule embeddings frozen is best or near-best by mean $\pm$ standard deviation overlap on 8 datasets. Both LINCS phases are included as held-out targets and show some of the clearest gains, with RMSE differences of 0.011 and 0.007 in favor of fine-tuning over scratch. More broadly, a model trained on all of Chem-PerturBridge at some stage gives the lowest or tied-lowest mean test RMSE in 10 of 11 datasets under the reported rounded means; the exception is CIGS MCE, which performs better with scratch. Additionally, CIGS TCM, sci-Plex, Tahoe, and Ginkgo GDPx2 reach their best without fine-tuning; for GDPx2, All Morgan fixed and Scratch Morgan fixed tie at 0.4321. VCPI-0001 is uninformative.

Testing these differences across seeds and across held-out molecules (Tables~\ref{tab:lpm_molecule_welch_holm} and~\ref{tab:lpm_molecule_bootstrap}), fine-tuning helps beyond both on LINCS Phase I, LINCS Phase II, Novartis DRUG-seq and OP3, and hurts under both on CIGS MCE.

\begin{table*}[t]
\caption{\textbf{Molecule-holdout multi-output LPM test RMSE.}
Mean $\pm$ standard deviation across 10 seeds. Test scores are
reported from the checkpoint with the best validation RMSE. \textbf{Bold} marks the
lowest mean in each row and methods whose mean $\pm$ one-standard-deviation
interval overlaps the corresponding interval of the lowest-mean method.
VCPI-0002 has no test molecules under the current split.}

\label{tab:expanded_lpm_test_rmse}
\centering

\scriptsize

\setlength{\tabcolsep}{2.5pt}

\resizebox{\textwidth}{!}{%

\begin{tabular}{lcccccccc}

\toprule

Dataset

& All

& FT frozen mol

& Scratch

& All Morgan fixed

& FT Morgan fixed

& Scratch Morgan fixed

& All Morgan learned

& FT Morgan learned fixmol \\

\midrule

CIGS MCE

& $0.5761 \pm 0.0007$

& $0.5784 \pm 0.0019$

& $\mathbf{0.5714 \pm 0.0020}$

& $\mathbf{0.5731 \pm 0.0003}$

& $0.5867 \pm 0.0066$

& $\mathbf{0.5718 \pm 0.0022}$

& $0.5803 \pm 0.0054$

& $\mathbf{0.5750 \pm 0.0017}$ \\

CIGS TCM

& $0.4082 \pm 0.0003$

& $0.4077 \pm 0.0001$

& $0.4081 \pm 0.0003$

& $\mathbf{0.4072 \pm 0.0002}$

& $0.4076 \pm 0.0001$

& $0.4079 \pm 0.0001$

& $0.4082 \pm 0.0004$

& $\mathbf{0.4077 \pm 0.0006}$ \\

DILImap train

& $0.6640 \pm 0.0377$

& $\mathbf{0.6090 \pm 0.0015}$

& $\mathbf{0.6095 \pm 0.0014}$

& $0.6637 \pm 0.0128$

& $\mathbf{0.6088 \pm 0.0002}$

& $\mathbf{0.6096 \pm 0.0006}$

& $0.6813 \pm 0.0272$

& $\mathbf{0.6095 \pm 0.0008}$ \\

Ginkgo GDPx2

& $0.4392 \pm 0.0027$

& $0.4431 \pm 0.0036$

& $\mathbf{0.4331 \pm 0.0031}$

& $\mathbf{0.4321 \pm 0.0013}$

& $\mathbf{0.4378 \pm 0.0051}$

& $\mathbf{0.4321 \pm 0.0016}$

& $0.4374 \pm 0.0018$

& $0.4380 \pm 0.0023$ \\

LINCS Phase I

& $0.6284 \pm 0.0010$

& $0.6231 \pm 0.0003$

& $0.6326 \pm 0.0014$

& $0.6266 \pm 0.0007$

& $0.6247 \pm 0.0012$

& $0.6266 \pm 0.0011$

& $0.6271 \pm 0.0010$

& $\mathbf{0.6214 \pm 0.0011}$ \\

LINCS Phase II

& $0.4849 \pm 0.0014$

& $0.4830 \pm 0.0006$

& $0.4886 \pm 0.0004$

& $0.4861 \pm 0.0010$

& $0.4859 \pm 0.0006$

& $0.4861 \pm 0.0004$

& $0.4834 \pm 0.0008$

& $\mathbf{0.4814 \pm 0.0004}$ \\

Novartis DRUG-seq

& $0.5557 \pm 0.0004$

& $0.5394 \pm 0.0006$

& $0.5408 \pm 0.0006$

& $0.5553 \pm 0.0003$

& $0.5379 \pm 0.0001$

& $0.5395 \pm 0.0008$

& $0.5553 \pm 0.0003$

& $\mathbf{0.5371 \pm 0.0006}$ \\

OP3

& $0.3572 \pm 0.0009$

& $\mathbf{0.3481 \pm 0.0019}$

& $\mathbf{0.3478 \pm 0.0011}$

& $0.3545 \pm 0.0033$

& $\mathbf{0.3413 \pm 0.0054}$

& $\mathbf{0.3469 \pm 0.0008}$

& $0.3564 \pm 0.0016$

& $\mathbf{0.3431 \pm 0.0013}$ \\

sci-Plex

& $\mathbf{0.6841 \pm 0.0027}$

& $0.6896 \pm 0.0015$

& $0.6845 \pm 0.0006$

& $\mathbf{0.6819 \pm 0.0006}$

& $0.6861 \pm 0.0011$

& $0.6840 \pm 0.0008$

& $\mathbf{0.6837 \pm 0.0022}$

& $0.6899 \pm 0.0028$ \\

Tahoe-100M

& $0.7383 \pm 0.0026$

& $0.7431 \pm 0.0012$

& $0.7380 \pm 0.0004$

& $\mathbf{0.7288 \pm 0.0010}$

& $0.7349 \pm 0.0004$

& $0.7335 \pm 0.0015$

& $0.7319 \pm 0.0013$

& $0.7354 \pm 0.0013$ \\

VCPI-0001

& $0.5708 \pm 0.0000$

& $\mathbf{0.5707 \pm 0.0000}$

& $\mathbf{0.5707 \pm 0.0001}$

& $\mathbf{0.5708 \pm 0.0001}$

& $\mathbf{0.5707 \pm 0.0000}$

& $\mathbf{0.5707 \pm 0.0001}$

& $\mathbf{0.5709 \pm 0.0002}$

& $\mathbf{0.5706 \pm 0.0001}$ \\

\bottomrule

\end{tabular}%

}
\end{table*}

\section{Limitations and impact}
\label{sec:limitations}\label{sec:broader_impact}

Chem-PerturBridge is limited by the coverage and overlap of available source
datasets. Cross-dataset agreement can only be evaluated for compounds, cellular
contexts, timepoints, doses, and genes that can be matched after harmonization.
As a result, the agreement benchmark covers a biologically and technically
non-random subset of the full resource. Gene coverage also differs across assay
families, especially between targeted-expression panels and RNA-seq, and local
dose matching can bias agreement estimates when source dose grids are sparse or
very different. Control designs, replicate structures, batch annotations,
laboratory protocols, and upstream preprocessing differ by source. 

Pseudobulking makes a single limma effect layer comparable across bulk, targeted and single-cell assays, but collapses within-condition heterogeneity into an aggregate profile that may obscure partial responses, cell-state effects and composition changes, donor effects, or rare responding subpopulations. Replicate agreement varies among pseudobulked datasets, so these comparisons alone cannot separate aggregation effects from assay effects. 
The limma-based effect layer provides a common representation, but alternative effect estimators or
normalization strategies may produce different agreement patterns.

The representation-learning experiments are also limited in scope. The OP3
evaluation tests transfer under a compound-held-out split in four immune cell
contexts and a relatively narrow compound panel, and the multi-dataset gain is partly confounded with compound coverage. The molecule-holdout LPM evaluation covers all source datasets, but it remains tied to the selected
model architecture, split design, and RMSE objective. The observed
gains demonstrate useful transferable signal under these protocols, especially
when cross-dataset training is combined with molecular priors and target-dataset
fine-tuning. However, no single training strategy dominates all datasets:
target-only training, fixed Morgan fingerprints, and dataset-specific adaptation
remain competitive depending on the scenario.

Chem-PerturBridge is intended for research on perturbation modeling.

\section{Discussion}

Chem-PerturBridge was designed to evaluate cross-dataset agreement in
small-molecule perturbation transcriptomic resources and to test whether a
harmonized compendium benefits perturbation models. The main finding is that
signature-level agreement and representation-level utility are distinct. At the
signature level, DEG-restricted $\mathrm{logFC}$ ranking and magnitude agreement are weak
and below the centroid baseline across datasets, though they exceed individual peer signatures in most. $\mathrm{LogFC}$ direction agreement on differentially expressed genes is
more stable and usually exceeds both kinds of baselines. This contrast is specific to agreement: in retrieval, direction leads only under a distance-based score, while the two representations perform similarly under scale-invariant cosine similarity in this evaluation. At the representation level,
pooled pretraining over Chem-PerturBridge improves held-out-compound prediction
in OP3 relative to L1000-only embeddings, Morgan fingerprints, embeddings from chemical language models, and a
descriptor-free OP3 baseline; part of that advantage reflects compound coverage. The molecule-holdout LPM evaluation
shows a broader pattern: Chem-PerturBridge-trained models are often strongest
when all-data training is combined with molecular priors and target-dataset
fine-tuning, but the best method varies across datasets. These results suggest that assay, batch, context,
and processing effects can dominate fine-grained signature agreement, while
compound-level signal can still be recovered by models trained across diverse
and well-annotated data.

This distinction is important for the use of large perturbation resources. Signatures of seemingly similar samples from different assays should not be expected to agree with one another 
without matched-condition validation. However, weak direct $\mathrm{logFC}$ agreement does not rule out representation learning from heterogeneous resources. Chem-PerturBridge
therefore provides both a diagnostic benchmark for evaluating when perturbation
effects agree across datasets and a model-ready substrate for learning reusable
compound representations. Future work may incorporate additional resources,
evaluate stricter dose-matching and alternative effect estimators, train under direction-aware or significance-weighted objectives, expand to
additional downstream tasks, and support the development of foundation
transcriptomic perturbation models~\cite{ginkgo_gdpx1_hf, ginkgo_gdpx4_hf,
laise_model_2022, dapello_scgenescope_2025, dini_multiplex_2025,
rood_toward_2024, li_aethercell_2026, miladinovic_silico_2025}.

\begin{ack}
F.J.T. consults for Immunai Inc., Singularity Bio B.V., CytoReason Ltd, and Omniscope Ltd, and has ownership interest in Dermagnostix GmbH and Cellarity. M.M. is an employee at Valinor Discovery. A.Z. was supported by a doctoral contract from \'Ecole Doctorale 474 FIRE, Universit\'e Paris Cit\'e, Learning Planet Institute.
\end{ack}

\bibliographystyle{unsrtnat}
\bibliography{references}

\newpage
\appendix

\section{Formal problem setup}
\label{app:formal_problem}

We study perturbation effects derived from bulk or pseudobulk transcriptomic measurements. A perturbation condition is indexed by
\[
\xi = (a,p,c,d,t) \in \mathcal{A}\times\mathcal{P}_{\mathrm{pert}}\times\mathcal{C}\times\mathcal{D}\times\mathcal{T},
\]
where \(a\) denotes dataset or assay domain, \(p\) perturbation identity, \(c\) cellular context, \(d\) dose, and \(t\) timepoint.

Let \(Z\) denote a latent single-cell state space and let \(\mu \in \mathcal{P}(Z)\) denote a population-level latent state. The assay observation operator
\[
V : \mathcal{P}(Z) \to Y
\]
maps latent population state to an observed transcriptomic measurement space. Because \(V\) differs across technologies, assays may measure different views of the same underlying response.

For each condition \(\xi\), let
\[
Y_\xi^{\mathrm{pert}} = \{y_1^{\mathrm{pert}},\dots,y_{n_{\mathrm{pert}}(\xi)}^{\mathrm{pert}}\},
\qquad
Y_\xi^{\mathrm{ctrl}} = \{y_1^{\mathrm{ctrl}},\dots,y_{n_{\mathrm{ctrl}}(\xi)}^{\mathrm{ctrl}}\}
\]
denote treated and control measurements. A perturbation-effect representation is
\[
r_\xi = \Psi\!\left(Y_\xi^{\mathrm{pert}},Y_\xi^{\mathrm{ctrl}}\right),
\]
where \(\Psi\) compares sets of replicate measurements. In this paper, \(\Psi\) is the shared limma-based differential-expression pipeline. The resulting condition-level signature is represented through gene-level log fold-changes, moderated \(t\)-statistics, and Benjamini--Hochberg adjusted \(p\)-values:
\[
r_\xi = \big((L_{\xi,g},\,T_{\xi,g},\,Q_{\xi,g})\big)_{g \in G_\xi},
\]
where \(Q_{\xi,g}\) is used to define the differential-expression mask \(M_\xi=\{g\in G_\xi : Q_{\xi,g}<\alpha\}\) (with \(\alpha=0.05\) in the reported analyses) employed by the DEG-restricted metrics in Section~\ref{sec:signature_eval} and Appendix~\ref{app:evaluation_details}.

This set-based formulation may reflect both transcriptional changes within cells and compositional changes in the responding population.

\section{Data harmonization and processing}
\label{app:data_processing_details}

\begin{figure*}[t]
\centering
\includegraphics[width=\textwidth]{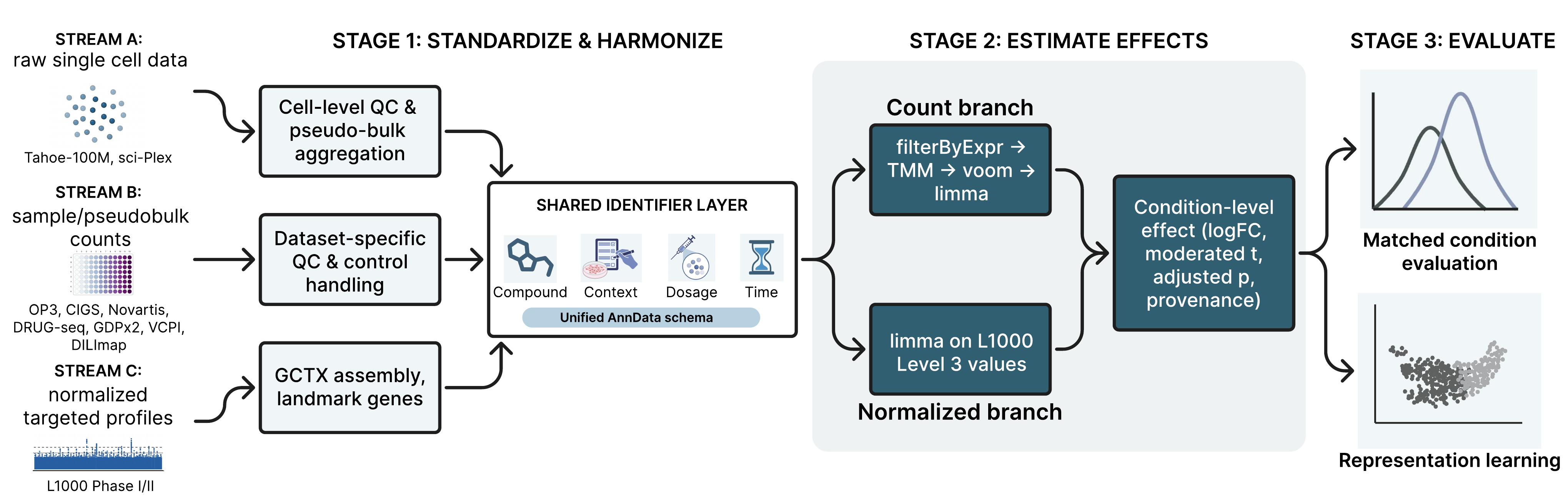}
\caption{\textbf{Three-stage harmonization and evaluation pipeline in Chem-PerturBridge.}
{Stage~1 (Standardize \& Harmonize):} heterogeneous source assays enter through three streams: raw single-cell data (Stream~A: Tahoe-100M, sci-Plex), sample or pseudobulk counts (Stream~B: OP3, CIGS, Novartis DRUG-seq, GDPx2, VCPI, DILImap), and normalized targeted profiles (Stream~C: L1000 Phase~I/II), each routed through assay-specific preprocessing before passing through a shared identifier layer that maps compounds to PubChem CIDs, contexts to Cellosaurus or Cell Ontology, doses to $\mu$M, timepoints to hours, and genes to a common Ensembl version, producing a unified AnnData schema.
{Stage~2 (Estimate Effects):} the unified schema feeds a two-branch limma layer: a count branch (filterByExpr, TMM normalization, voom, and limma) for pseudobulked and bulk count data, and a normalized branch (limma applied directly to L1000 Level~3 values), producing condition-level effects comprising log fold-changes, moderated $t$-statistics, adjusted $p$-values, and provenance fields. {Stage~3 (Evaluate):} the condition-level effects are used for matched-condition cross-dataset and replicate agreement evaluation and for pooled representation learning.
Per-dataset processing rules are detailed in Table~\ref{tab:processing_details}.}
\label{fig:data_processing}
\end{figure*}

Each source dataset was converted into a harmonized condition-level representation containing compound identity, cellular context, timepoint, dose, control assignment, replicate provenance, and assay provenance. The harmonized \texttt{obs} schema contains \texttt{sample\_id}, \texttt{plate}, \texttt{well}, \texttt{cell\_type}, \texttt{perturbagen}, \texttt{pert\_type}, \texttt{is\_control}, \texttt{pert\_dose\_uM}, \texttt{pert\_time\_h}, assay and donor metadata, \texttt{pubchem\_cid}, \texttt{psbulk\_cells}, and \texttt{psbulk\_counts}. Some datasets retain additional fields needed for stratification, such as \texttt{split} for DILImap and OP3 or \texttt{percent\_volume\_dmso} for GDPx2. The harmonized \texttt{var} schema uses Ensembl identifiers as the gene index where available and stores the corresponding gene symbol.

The processing pipeline depended on the upstream data, as indicated in Table~\ref{tab:processing_details}. Single-cell datasets were harmonized at the cell level and aggregated into pseudobulk count profiles. Count-based sample-level assays and already pseudobulked datasets were harmonized directly, with \texttt{psbulk\_cells} set to the sentinel value \(-666\) when no cell count is meaningful or available. L1000 Phase I/II required a normalized-input route because Level 3 contains processed log-scale expression rather than raw counts.

For Tahoe-100M and sci-Plex, pseudobulk sample identifiers were constructed from the available fields among \texttt{plate}, \texttt{well}, \texttt{perturbagen}, \texttt{cell\_type}, and \texttt{guide}. Cells with missing grouping fields were removed before aggregation. The single-cell filters used before pseudobulking were \(\mathrm{ngenes}\geq250\), \(\mathrm{ncounts}\geq700\), mitochondrial fraction \(\leq 0.2\), and a human \textit{MALAT1} expression filter using \(\log(1 + 10000\cdot f_{\mathrm{MALAT1}})\geq3.5\)~\cite{clarke_malat1_2024}. Decoupler~\cite{badia-i-mompel_decoupler_2022} pseudobulk aggregation summed cells within each composite \texttt{sample\_id} and stored both the count profile and the number of contributing cells. OP3 is downloaded as an already aggregated pseudobulk object with source-provided cell counts, whereas bulk, targeted count, and L1000 sample-level resources entered one row per assayed sample.

Differential-expression preprocessing created two replicate labelings for each source. In the \texttt{group\_all\_replicates} setting, treated samples were labeled by cell type, perturbagen, dose, and timepoint, while controls were labeled by cell type, perturbagen, and timepoint. In the \texttt{separate\_replicates} setting, treated labels additionally included well and plate so that within-dataset replicate agreement could be evaluated without collapsing technical replicates into one condition. Datasets were split by \texttt{cell\_type} before limma fitting; GDPx2 was split by both \texttt{cell\_type} and \texttt{percent\_volume\_dmso}. Tahoe-100M, sci-Plex, OP3, and L1000 analyses were restricted to the 24-hour configuration used in the cross-dataset evaluation; datasets without a configured time restriction retained all standardized timepoints.

\begin{table*}[t]
\caption{Dataset-specific harmonization rules used before condition-level effect estimation.}
\label{tab:processing_details}
\centering
\scriptsize
\setlength{\tabcolsep}{3pt}
\begin{tabular}{p{2.1cm} p{2.6cm} p{3.3cm} p{6.4cm}}
\toprule
Processing route & Sources & Input object & Dataset-specific operations before effect construction \\
\midrule
Raw single-cell to pseudobulk & Tahoe-100M, sci-Plex & Cell-level AnnData~\cite{virshup_anndata_2024} plus curated \texttt{obs}/\texttt{var} mainly from Lamin pertdata~\cite{sunny_pertdata_2026} & Standardize cell metadata; map cell contexts to Cellosaurus~\cite{bairoch_cellosaurus_2018} or Cell Ontology~\cite{diehl_cell_2016} IDs; map compounds to PubChem CIDs with manual resolution when needed; filter cells with low counts, low detected genes, high mitochondrial fraction, missing grouping fields, and low MALAT1 signal where enabled; aggregate with Decoupler using composite sample groups to obtain pseudobulk counts and \texttt{psbulk\_cells}. Tahoe plates are processed separately and concatenated; sci-Plex controls are renamed to DMSO and cell lines are mapped to Cellosaurus IDs. \\
Pre-aggregated single-cell & OP3 & Prepared pseudobulk AnnData & Download the prepared OP3 pseudobulk file; remove non-DMSO controls; map immune cell subsets to Cell Ontology IDs; merge donor metadata; annotate compounds with PubChem CIDs using names and SMILES; map gene symbols to Ensembl IDs with a manual fallback table. \\
Normalized targeted-expression & L1000 Phase I/II & Level 3 GCTX matrices and LINCS metadata & Download and assemble Level 3 matrices; keep compound treatments and DMSO vehicle controls; add metadata fields following Lamin pertdata curation; convert dose units to \(\mu\)M and time units to hours; annotate cell lines with Cellosaurus and donor metadata; merge LINCS compound tables and PubChem annotations; restrict to landmark genes by default and map Entrez IDs to Ensembl IDs. Because values are already normalized, downstream modeling uses limma without count filtering, TMM, or voom. \\
Count-based DRUG-seq and related bulk assays & Novartis DRUG-seq, VCPI-0001, VCPI-0002, GDPx2 & Raw count matrices from RData, VCPI payloads, or Lamin artifacts & Convert source payloads to AnnData; drop empty libraries and source-defined positive controls; for Novartis, retain robust DMSO reference controls and remove specified positive-control wells; parse concentration and time strings into \(\mu\)M and hours; map cell-line or primary-cell metadata; drop ERCC spike-ins where present; annotate compounds from InChIKey, SMILES, CAS, or names with PubChem lookup. GDPx2 additionally retains \texttt{percent\_volume\_dmso} for grouping. \\
CIGS targeted count panels & CIGS MCE, CIGS TCM & Excel count and metadata tables & Convert per-subset Excel files to AnnData and keep MCE and TCM separate; inner-join genes within each source; apply the paper-mirroring QC filter that drops BLANK/RNA wells and low-quality libraries; drop JQ1 samples; derive dose and time from the subset design; normalize MCE wells and TCM sample identifiers separately; canonicalize Excel/R-mangled gene symbols before Ensembl lookup. \\
DILImap bulk RNA-seq & DILImap train and train+validation & Count AnnData files & Keep the training-only and train+validation variants separate to avoid test-set contamination; apply LDH viability, RNA/mitochondrial, and cross-replicate-correlation QC; standardize near-identical dose values by log-scale clustering with training doses preferred as representatives; set primary hepatocyte metadata; annotate compounds with DILImap SMILES plus PubChem lookup; retain the original split label. \\
\bottomrule
\end{tabular}
\end{table*}

Additional exclusions were applied before the shared differential gene expression analysis. OP3 retained only DMSO vehicle controls. Novartis removed zero-count libraries, wells marked as empty, reference-control wells not present in the robust DMSO table of that dataset, and sample-analysis wells whose compound IDs were in the positive-control set \{\texttt{BD-11-DV28}, \texttt{EA-18-FP00}, \texttt{SE-15-AV21}\}. VCPI removed ERCC genes, zero-count libraries, and non-DMSO rows marked as controls. GDPx2 removed ERCC genes, zero-count libraries, and rows whose \texttt{sample\_type} began with \texttt{Ginkgo Pos Control}. CIGS first removed zero-count libraries, then applied the paper's R quality-control filter, which drops BLANK/RNA wells and per-subset low-read outliers, and finally removed JQ1 samples. DILImap retained only samples passing all three source QC checks: empty \texttt{LDH\_QC} and not \texttt{DMSO\_replaced}, log-total RNA greater than the dataset median minus \(2.5\) standard deviations with mitochondrial RNA fraction below \(9\%\), and the DILImap cross-replicate-correlation QC.

\subsection{Compound and context harmonization}

Compounds were harmonized to shared identifiers by mapping to \texttt{pubchem\_cid} when possible. The lookup strategy was source-dependent: L1000 used LINCS compound tables together with InChIKey, SMILES, compound names, and manual mappings; Novartis, VCPI, GDPx2, CIGS, OP3, and DILImap used the structural or catalog identifiers available in the source metadata, and dataset-specific manual mappings for unresolved cases. Source perturbagen names and PubChem CIDs were both retained. For L1000 differential-expression labels, PubChem CIDs were used when available to disambiguate compounds with the same name but different structures.

Cell lines and related contexts were harmonized using standardized metadata, including Cellosaurus identifiers for cell lines and, where available, Cell Ontology identifiers for primary cell types. Timepoints were represented in hours. Conditions were considered eligible for cross-dataset matching only when compound identity, cell context, and timepoint matched exactly after harmonization.

\subsection{Dose standardization and cross-dataset matching}

\begin{figure*}[!t]
\centering
\includegraphics[width=0.9\textwidth]{figures/cross_dataset.pdf}
\caption{\textbf{Chem-PerturBridge cross-dataset overlap.}
\textbf{a)} Number of shared compounds. \textbf{b)} Shared cellular contexts. \textbf{c)} Matched conditions, i.e.\ tuples of compound, line, time, and dose. Diagonals count each dataset's compounds (a) and contexts (b) shared with at least one other dataset. The overlap between the two L1000 phases is limited to the cell lines and compounds that appear in other datasets, to avoid the disproportionate L1000--L1000 overlap dominating the summary. Each panel is a dataset-by-dataset matrix over the resources in Table~\ref{tab:cpb_datasets}.}
\label{fig:cpb_overlap}
\end{figure*}

Dose values were converted to a common micromolar scale and transformed to \(\log_{10}\) dose for matching. Unit conversions were handled during standardization, including nanomolar and millimolar values in L1000 and Ginkgo-derived resources, percent vehicle entries in GDPx2, and fixed design doses in CIGS. Controls were assigned dose \(0\). We also applied clustering of near-identical DILImap doses on the log scale, so that numerically equivalent training, validation, and test dose values were represented by a single canonical value.

We used L1000 Phase I instance-level doses as recorded, including two
differences from signature-level metadata: 489 samples recorded
as 0 $\mu\mathrm{M}$ (signature-level \texttt{pert\_idose} label: 1 $\mathrm{nM}$)
and 835 KCl/NaCl samples recorded as 100 $\mathrm{mM}$
(signature-level numeric dose: 0.1 $\mu\mathrm{M}$). 

Both groups of samples were retained in the dataset. Zero-dose conditions were excluded from agreement and retrieval analyses but retained in the LPM input, with zero-dose log values set to $-5$, matching the maximum absolute finite log-dose in magnitude. Together, these samples account for 0.19\% of the 699,276 L1000 Phase I samples in Table~\ref{tab:cpb_datasets}; we did not assess alternative handling.

For each pair of datasets, candidate matches were constructed among conditions with the same compound, cell context, and timepoint. Within this candidate set, dose was matched by mutual nearest neighbors in \(\log_{10}\) dose. A matched condition pair was retained only when the dose mismatch was at most one order of magnitude.

To characterize how much dose mismatch the threshold actually admits, we report the fold-difference distribution across retained pairs. Table~\ref{tab:dose_mismatch} gives per-pair fold-difference quantiles and the number of pairs retained at successively tighter thresholds. Of the 76,717 retained pairs, 50,072 (65.3\%) survive at exact dose, 75,722 (98.7\%) at 2-fold, and 75,759 (98.8\%) at 3-fold; tightening from 10-fold to 3-fold removes 958 pairs (1.2\%), of which 927 fall in the L1000 Phase I--Phase II pair. The tolerance is therefore permissive, but almost no retained pair approaches the 10-fold threshold.

Additionally, a threshold sensitivity analysis shows that changing the threshold affects the two metrics differently (Table~\ref{tab:dose_threshold_sensitivity}). Tightening from 10-fold to exact dose moves DEG-restricted Spearman by up to 0.106 but direction agreement by at most 0.012, consistent with dose mismatch perturbing the magnitude of the response while preserving its direction; the CIGS-MCE--CIGS-TCM pair is the one exception, with DEG-restricted Spearman decreasing by 0.014. The 2-fold, 3-fold, and 10-fold results are nearly identical because 98.7\% of pairs are retained when the threshold is tightened to 2-fold. The baseline-adjusted DEG-restricted Spearman difference remains negative at exact dose in all 6 pairs that can be scored there, so the negative \(\Delta\) is not an artifact of dose mismatch. Figure~\ref{fig:dose_mismatch_vs_similarity} shows that dose mismatch has little association with cross-dataset signature similarity, although the largest-mismatch bins contain few units.

Three dataset pairs involving Tahoe cannot be evaluated at exact dose, since the Tahoe dose grid does not coincide with the L1000 or sci-Plex grids; the minimum attainable fold differences are 1.21, 1.21, and 2.00. A small number of matched pairs (110 of 76,717, 0.14\%) had no corresponding grouped result in the source line files and were not scored; these are concentrated in sci-Plex at 24 h; unresolved pairs are listed in the code repository.

\begin{table*}[t]
\caption{\textbf{Dose mismatch in matched cross-dataset pairs.}
Fold-difference quantiles over pairs retained at the $10\times$ threshold, and the number of pairs retained at successively tighter dose thresholds. The final column is the difference between the $10\times$ and $3\times$ counts, which includes pairs dropped when their context falls below the ten-compound minimum.}
\label{tab:dose_mismatch}
\centering
\footnotesize
\setlength{\tabcolsep}{3pt}
\begin{tabular}{lrrrrrrrrr}
\toprule
& \multicolumn{4}{c}{Fold difference} &
\multicolumn{5}{c}{\# Matched pairs by dose threshold} \\
\cmidrule(lr){2-5} \cmidrule(lr){6-10}
Dataset pair & min & median & q95 & max &
exact & 2$\times$ & 3$\times$ & 10$\times$ & $\#_{10\times}$--$\#_{3\times}$ \\
\midrule
L1000 Phase I--L1000 Phase II &
1.00 & 1.00 & 1.66 & 10.00 &
29{,}250 & 49{,}733 & 49{,}768 & 50{,}695 & 927 \\
CIGS-MCE--CIGS-TCM &
1.00 & 1.00 & 1.00 & 2.00 &
15{,}863 & 15{,}973 & 15{,}973 & 15{,}973 & 0 \\
L1000 Phase I--sci-Plex &
1.00 & 1.00 & 1.30 & 10.00 &
1{,}918 & 2{,}689 & 2{,}689 & 2{,}695 & 6 \\
L1000 Phase II--sci-Plex &
1.00 & 1.11 & 1.23 & 1.23 &
1{,}147 & 3{,}192 & 3{,}192 & 3{,}192 & 0 \\
L1000 Phase I--CIGS-MCE &
1.00 & 1.00 & 1.00 & 5.00 &
1{,}395 & 1{,}395 & 1{,}395 & 1{,}401 & 6 \\
L1000 Phase II--CIGS-MCE &
1.00 & 1.00 & 1.00 & 3.00 &
499 & 499 & 499 & 508 & 9 \\
L1000 Phase I--Tahoe-100M &
1.21 & 2.00 & 2.00 & 8.88 &
0 & 600 & 600 & 610 & 10 \\
L1000 Phase II--Tahoe-100M &
1.21 & 1.35 & 1.50 & 2.40 &
0 & 1{,}471 & 1{,}473 & 1{,}473 & 0 \\
Tahoe-100M--sci-Plex &
2.00 & 2.00 & 2.00 & 2.00 &
0 & 170 & 170 & 170 & 0 \\
\bottomrule
\end{tabular}
\end{table*}

\begin{table*}[t]
\caption{\textbf{Threshold sensitivity.}
Results are shown for DEG-restricted Spearman correlation and direction agreement metrics at exact, $2\times$ and $10\times$ dose thresholds. Each entry reports the observed score / observed-minus-baseline difference \(\Delta\). The baseline is the source centroid. Tahoe's dose grid does not coincide exactly with the L1000 or sci-Plex grids, so n/a marks pairs with no exact-dose matches.}
\label{tab:dose_threshold_sensitivity}
\centering
\scriptsize
\renewcommand{\arraystretch}{1.15}
\setlength{\tabcolsep}{5pt}
\begin{tabular}{>{\raggedright\arraybackslash}p{3.4cm}@{\hspace{0.8pt}}cccccc}
\toprule
& \multicolumn{3}{c}{DEG Spearman} & \multicolumn{3}{c}{Direction} \\
\cmidrule(lr){2-4} \cmidrule(lr){5-7}
Dataset pair & exact & 2$\times$ & 10$\times$ & exact & 2$\times$ & 10$\times$ \\
\midrule
L1000 Phase I--L1000 Phase II
  & 0.478 / -0.053 & 0.373 / -0.095 & 0.372 / -0.094
  & 0.929 / +0.152 & 0.920 / +0.158 & 0.917 / +0.156 \\
CIGS-MCE--CIGS-TCM
  & 0.351 / -0.142 & 0.365 / -0.129 & 0.365 / -0.129
  & 0.920 / +0.160 & 0.923 / +0.164 & 0.923 / +0.164 \\
L1000 Phase I--sci-Plex
  & 0.428 / -0.174 & 0.395 / -0.188 & 0.395 / -0.188
  & 0.899 / +0.130 & 0.888 / +0.129 & 0.890 / +0.131 \\
L1000 Phase II--sci-Plex
  & 0.538 / -0.115 & 0.491 / -0.107 & 0.491 / -0.107
  & 0.949 / +0.184 & 0.949 / +0.186 & 0.949 / +0.186 \\
L1000 Phase I--CIGS-MCE
  & 0.378 / -0.190 & 0.378 / -0.190 & 0.373 / -0.193
  & 0.878 / +0.055 & 0.878 / +0.055 & 0.873 / +0.053 \\
L1000 Phase II--CIGS-MCE
  & 0.482 / -0.155 & 0.482 / -0.155 & 0.479 / -0.163
  & 0.928 / +0.102 & 0.928 / +0.102 & 0.919 / +0.094 \\
L1000 Phase I--Tahoe-100M
  & n/a & 0.174 / -0.250 & 0.174 / -0.250
  & n/a & 0.736 / -0.018 & 0.736 / -0.018 \\
L1000 Phase II--Tahoe-100M
  & n/a & 0.084 / -0.305 & 0.084 / -0.305
  & n/a & 0.675 / -0.093 & 0.675 / -0.093 \\
Tahoe-100M--sci-Plex
  & n/a & 0.124 / -0.465 & 0.124 / -0.465
  & n/a & 0.896 / +0.083 & 0.896 / +0.083 \\
\bottomrule
\end{tabular}
\end{table*}

\subsection{Limma-based effect construction}

For every source dataset, perturbation effects were estimated with a limma-based differential-expression pipeline. Let \(g\) index genes; each grouped perturbation condition has associated treated and control replicates. Samples were first filtered by minimum contributing cell count when that quantity was available. If source QC flags were present and QC filtering was enabled, samples with \texttt{qc\_pass} set to false were removed. Redundant controls at timepoints with no treated samples were removed before model fitting, and each treated condition was compared only to controls from the same timepoint.

Each dataset was partitioned into strata before model fitting, and limma was fit independently within each stratum. A stratum $s$ corresponds to one group of samples produced by the dataset-specific splitting rule described above (\texttt{cell\_type} for most datasets; \texttt{cell\_type} and \texttt{percent\_volume\_dmso} for GDPx2). For stratum $s$, let $i$ index its samples, $\ell\in\mathcal{L}_s$ index its condition labels (compound, dose, and timepoint for treated samples; perturbagen and timepoint for controls), and $b\in\mathcal{B}_s$ index its plate labels. The design has no intercept. When the stratum spans multiple plates, the fitted model is
\[
y_{ig}
= \sum_{\ell\in\mathcal{L}_s}\mathbf{1}\{\mathrm{condition}_i=\ell\}\beta_{\ell g}
+ \sum_{b\in\mathcal{B}_s^\star}\mathbf{1}\{\mathrm{plate}_i=b\}\gamma_{bg}
+ \epsilon_{ig},
\]
corresponding to the R formula $\sim 0+\mathrm{condition}+\mathrm{plate}$. Because removing the intercept keeps all condition columns, identifiability requires dropping one plate level; $\mathcal{B}_s^\star\subset\mathcal{B}_s$ denotes the set of retained, non-reference plate levels. Under this parameterization, $\beta_{\ell g}$ is the mean expression of gene $g$ for condition $\ell$ on the reference plate, and $\gamma_{bg}$ is the additive offset of plate $b$ relative to that reference. When the stratum has a single plate, the plate term is omitted and the formula reduces to $\sim 0+\mathrm{condition}$. The condition coefficients $\beta_{\ell g}$ are the biological quantities of interest; the plate coefficients $\gamma_{bg}$ are nuisance terms.

For count-based datasets, genes were filtered with edgeR \texttt{filterByExpr}, library sizes were normalized with TMM, and voom-transformed expression values were passed to limma. For normalized L1000 Level 3 profiles, the pipeline skipped count filtering, TMM normalization, and voom, and fit limma directly to the normalized values. For each treated label $\ell_t$, contrasts were formed against every available control label $\ell_0$ at the same timepoint as
\[
\Delta_{\ell_t,\ell_0,g}=\beta_{\ell_t g}-\beta_{\ell_0 g},
\]
which compares the treated and control conditions on the same plate-adjusted model scale, so the explicit additive plate terms do not enter the contrast. Empirical Bayes moderation was then applied with robust estimation, and the normalized-input branch additionally used limma's trend option. For each contrast and gene, we retained the log fold-change, moderated $t$-statistic, average expression, residual standard deviation, unscaled standard error, $95\%$ confidence interval, nominal $p$-value, Benjamini--Hochberg adjusted $p$-values computed both within the contrast and across all contrasts in the fitted stratum, and the limma $B$-statistic.

\section{Cross-dataset and within-dataset agreement evaluation}
\label{app:evaluation_details}

The cross-dataset analyses used \texttt{group\_rep} differential-expression rows, in which all eligible technical or biological replicates for a compound-context-time-dose condition have already been grouped before limma fitting. The replicate analyses used \texttt{sep\_rep} differential-expression rows, in which replicate ID is preserved in the condition label so that replicate-level differential signatures can be compared after each replicate has been contrasted to the appropriate controls.

\subsection{Cross-dataset overlap filtering and match construction}

For each dataset, non-control rows were eligible for overlap matching only if they had a non-empty compound identifier, a non-empty harmonized context key, time, dose, and a normalized PubChem CID. The implementation used \texttt{cell\_type} as the harmonized context key, represented time as a formatted hour value, and represented dose by \(\log_{10}(\mathrm{pert\_dose\_uM})\). Rows were grouped by compound, context, and time. For each dataset pair, a candidate condition match was considered only when these three keys were shared.

Within a shared compound-context-time key, dose matching used all tied mutual nearest neighbors on the log-dose scale. If \(D_A\) and \(D_B\) are the positive dose grids for the same compound, context, and time in datasets \(A\) and \(B\), a dose pair \((d_a,d_b)\) was retained when
\[
|\log_{10}d_a-\log_{10}d_b|
= \min_{d\in D_B}|\log_{10}d_a-\log_{10}d|
= \min_{d\in D_A}|\log_{10}d-\log_{10}d_b|
\]
and \(|\log_{10}d_a-\log_{10}d_b|\leq 1\). Sparse dose grids therefore could not create matches to arbitrarily distant doses.

For overlap heatmaps and cross-dataset agreement scoring, a dataset-pair context stratum was retained only when at least 10 distinct matched compounds were present in that context. The overlap-building script also suppresses overlaps present only in the L1000 Phase I--Phase II pair by default, unless the matched context or compound is present in another dataset pair. This prevents the very large within-L1000 overlap from dominating the cross-dataset analysis. DMSO controls from retained context-time strata were kept in the overlap-filtered files for completeness, but the agreement and retrieval  removed controls before scoring matched perturbation effects.

\subsection{Effect representations}
\label{app:representations}
All agreement and retrieval metrics operate on a per-gene effect vector derived from the limma differential-expression output. In total, seven families of representation were evaluated: 
\begin{itemize}
    \item \(\mathrm{logFC}\): the log fold-change is the effect size itself.
    \item \(\mathrm{sign(logFC)}\): the sign of log fold-change, referred to as direction, discards magnitude.
    \item \(z\)-scores: a standardization of \(\mathrm{logFC}\), using one population mean and standard deviation per gene over all eligible non-control grouped-condition signatures, fitted at two scales: per dataset, and per dataset and cell line. This places sources with different normalization pipelines on a common scale.
    \item moderated \(t\): the moderated \(t\)-statistic divides the effect size by its moderated standard error, which limma stabilizes by borrowing variance information across genes.
    \item \(-\log_{10}(p\textrm{-value})\times\mathrm{sign}(\mathrm{logFC})\): signed significance combines the direction of regulation with the strength of evidence for it. 
    \item $\mathrm{ashr}$: the posterior-mean $\mathrm{logFC}$ under adaptive shrinkage, with one prior fitted per cell line within each dataset, pooling across batches. It shrinks uncertain effect estimates toward zero.
    \item $p$-value: the unsigned limma \(p\)-value, which measures the strength of evidence for a change but discards its direction.
    
\end{itemize}

Differential-expression status is defined by the two criteria given in Appendix~\ref{app:deg_focused_direct_metrics} and is fixed across representations.

\subsection{Gene universes and direct signature metrics}
\label{app:gene_direct_metrics}
For each resolved cross-dataset matched pair \((A,i;B,j)\), scores were computed on genes shared by the two source cell lines. Most reported metrics use this pairwise shared-gene set.

Let \(v_{i, g}\) denote a chosen representation from Appendix~\ref{app:representations} for condition-level signature \(i\) at gene \(g\); \(\mathrm{logFC}\), per-gene population \(z\)-scores, and moderated \(t\) are used for direct signature metric calculation.

The whole-signature direct agreement metric is Spearman correlation between the \(v\) vectors of the two signatures:
\[
\rho_{\mathrm{dir}}^{v}(i,j)
=
\mathrm{Spearman}_g\!\left(v_{i,g}^{A}, v_{j,g}^{B}\right),
\]
computed separately for each representation. Here \(\mathrm{Spearman}_g\) is taken over the pairwise shared-gene set.

The code also computes a signed top-\(k\) overlap score using moderated \(t\) rankings, with \(k=50\) in the cross-dataset signature analysis. Let \(U_i^{k_{\mathrm{eff}}}\) and \(D_i^{k_{\mathrm{eff}}}\) denote the $k_{\mathrm{eff}}$ genes with the largest and smallest moderated $t$ for signature \(i\), where \(k_{\mathrm{eff}}=\min(k,\lfloor |G|/2\rfloor)\) and $G$ is the subset of the pairwise shared-gene set with finite values in both signatures. The pairwise overlap score is
\[
\mathrm{Overlap@}k(i,j)
=
\frac{1}{2}
\left(
\frac{|U_i^{k_{\mathrm{eff}}} \cap U_j^{k_{\mathrm{eff}}}|}{k_{\mathrm{eff}}}
+
\frac{|D_i^{k_{\mathrm{eff}}} \cap D_j^{k_{\mathrm{eff}}}|}{k_{\mathrm{eff}}}
\right).
\]

\subsection{DEG-focused direct metrics}
\label{app:deg_focused_direct_metrics}
The DEG notebooks evaluate two definitions of differential expression: adjusted \(p<0.05\), and adjusted \(p<0.05\) with \(|\mathrm{logFC}|>0.2\). The main heatmaps use the adjusted-\(p<0.05\) definition. The adjusted \(p\)-value layer is \texttt{adj.P.Value.within\_one\_contrast}.

For a matched pair, let \(M_i\) and \(M_j\) be the DEG masks for the two signatures. The symmetric DEG-restricted \(v\)-representation Spearman score is
\[
\rho_{\mathrm{DEG}}^v(i,j)
=
\frac{1}{2}
\left[
\mathrm{Spearman}_{g\in M_i}(v_{i,g},v_{j,g})
+
\mathrm{Spearman}_{g\in M_j}(v_{i,g},v_{j,g})
\right],
\]
provided both terms are defined; \(\mathrm{logFC}\), per-gene population \(z\)-scores, and moderated \(t\) are used as representations for DEG-restricted correlation. 

DEG overlap was computed with the reference DEG-set size \(N_{\mathrm{DEG}}=|M_i|\) or \(N_{\mathrm{DEG}}=|M_j|\), symmetrized across the two references, and at fixed \(k\in\{50,100,200\}\) when both signatures had at least \(k\) DE genes. Direction agreement was the fraction of genes in \(M_i\cap M_j\) for which the two log fold-changes had the same sign.

\subsection{Centroid consensus baselines}
\label{app:centroid_baseline}
We retained the original centroid baseline as a hard comparator for direct signature and DEG agreement, alongside the primary individual-peer baselines of Appendix~\ref{app:peer_baseline}.

Let \(m\) denote any of the agreement metrics of Appendices~\ref{app:gene_direct_metrics} and~\ref{app:deg_focused_direct_metrics}, so that \(m(i, j)\) is the observed score for a matched pair and \(m(i, r)\) is the score of signature \(i\) against a reference signature \(r\).

For a matched signature \(i\) in dataset \(A\), the baseline vector \(\bar r^{A}_{-p(i),c,t,d}\) is the gene-wise mean over non-control rows in dataset \(A\) with the same cell context, time key, and dose key as \(i\), excluding rows with the same compound. The same construction is applied independently to the matched signature \(j\) in dataset \(B\). The construction above gives the source-side centroid; scoring each signature against the other dataset's centroid gives the target-side centroid. We report the source-side form by default.

The left and right baseline scores are \(m(i,\bar r^{A}_{-p(i),c,t,d})\) and \(m(j,\bar r^{B}_{-p(j),c,t,d})\), evaluated on the same gene universe as the corresponding observed score, with DEG-specific gene selection as described below. The pair baseline is their mean, and the baseline-adjusted score is
\[
\Delta m(i,j)_{\mathrm{cent}}
=
m(i,j)
-
\frac{1}{2}
\left[
m(i,\bar r^{A}_{-p(i),c,t,d})
+
m(j,\bar r^{B}_{-p(j),c,t,d})
\right].
\]
For DEG centroid baselines, the vector has no limma \(p\)-values of its own. The code therefore evaluates baseline DEG-restricted Spearman and direction agreement on the signature's DE genes, and evaluates baseline DE overlap by comparing the signature's ranked DE genes to the baseline genes ranked by absolute value of the chosen representation.

\subsection{Individual-peer baselines}
\label{app:peer_baseline}

The primary baseline for direct signature and DEG agreement is an individual same-context different-compound signature. For a matched signature \(i\) in dataset \(A\), a source peer is a non-control row from dataset $A$ with the same cell context, time key, and dose key as \(i\), excluding rows with the same compound. The same construction is applied independently to the matched signature \(j\) in dataset \(B\). Target peers for $i$ use the source-peer pool of $j$, and vice versa; thus, only the target-peer comparison crosses datasets as the observed score does.

Let \(R_i\) and \(R_j\) denote the peer sets for signatures \(i\) and \(j\), with sizes \(K_i=|R_i|\) and \(K_j=|R_j|\); for the source variant \(R_i \subset A\) and \(R_j \subset B\), and for the target variant the reverse. Each signature is scored against its peers separately rather than against their average, giving vectors of peer scores \(\{m(i, r)\}_{r\in R_i}\) and \(\{m(j, r)\}_{r\in R_j}\), where $m$ is defined in Appendix~\ref{app:centroid_baseline}. For DEG-focused peer comparisons, the DEG mask is taken from the signature being scored; peer genes are not required to be differentially expressed. When the number of qualifying peers exceeds a fixed cap, peers are subsampled deterministically; runs without a cap use all qualifying peers. The cap applies to the individual-peer baseline only. The baseline-adjusted score is
\[
\Delta m(i,j)_{\mathrm{peer}}
=
m(i,j)
-
\frac{1}{2}
\left[
\frac{1}{K_i}
\sum_{r\in R_i}
m(i, r) +
\frac{1}{K_j}
\sum_{r\in R_j}
m(j, r)
\right]
,
\]
and the peer percentile:

\[
\pi(i, j) = \frac{1}{2}\left[\frac{1 + \#\{r\in R_i: m(i, r) < m_i\}}{1 + K_i}
+ \frac{1 + \#\{r\in R_j: m(j, r) < m_j\}}{1 + K_j}\right]
,
\]

where the \(+1\) terms apply the standard finite-sample correction, keeping \(\pi\) in \((0, 1]\). Larger values indicate that the observed agreement exceeds a greater proportion of the peer scores. Here \(m_i\) and \(m_j\) denote the observed terms each side is referenced to. These equal \(m(i, j)\) except for DEG-restricted Spearman, whose observed score is itself symmetrized: there each side is referenced to the term computed on its own DEG mask. Both \(\Delta m(i,j)_{\mathrm{peer}}\) and \(\pi(i, j)\) are computed separately against source and target peers and reported as separate columns; where one side is undefined, baseline means and peer percentiles are averaged over the available side.

\subsection{Retrieval cross-dataset metrics}
\label{app:retrieval_cross_dataset}

\begin{figure*}[!t]
\centering
\includegraphics[width=0.8\textwidth]{figures/retrieval_upd_l2.pdf}
\caption{\textbf{$\mathbf{logFC}$ retrieval across Chem-PerturBridge, negative $L_2$.}
Cross-dataset retrieval ranks the matched compound above the target-peer baseline in most pairs, but below the averaged centroid, which is itself retrieved at rank nearly 1. Retrieval on signed directions is stronger than on magnitudes (mean rank 0.68 vs. 0.58) and exceeds the target-peer baseline in every pair, while remaining mixed against the source peer and below the centroid. Each panel is a dataset-by-dataset matrix over the resources in Table~\ref{tab:cpb_datasets}. \textbf{a--d)} Cross-dataset
$\mathrm{logFC}$ retrieval of the matched partner against same-context different-compound
distractors. \textbf{e--h)} The same retrieval on signed directions, $\mathrm{sign}(\mathrm{logFC})$. Retrieval is reported as normalized best-positive rank, where higher is better; 1.0 is the best. Within each row, panels report the cross-dataset score and the remaining three baselines: the source-side centroid baseline, individual source peers and individual target peers. Perturbagen-clustered bootstrap confidence intervals are reported in
Table~\ref{tab:cross_dataset_retrieval_logfc_bootstrap_ci}.}
\label{fig:cpb_overview_l2}
\end{figure*}

The retrieval analysis asks whether a query signature can identify its
same-compound partner among same-context target-side alternatives. For a fixed
dataset pair and cell-type-time stratum, every retained condition in one
dataset is used as a query against retained conditions in the other dataset;
scoring is performed in both directions. A stratum is eligible only if each side contains at least two unique compounds and at least five candidate conditions; each scored direction additionally requires at least five valid target candidates.

For each query, target candidates are restricted to the matched target-side
cell context and timepoint. In the compound-across-doses variant, positive candidates are target-side conditions
with the same PubChem CID as the query; all other target-side candidates are same-context different-compound distractors. In the strict matched-condition variant, used for the main retrieval heatmaps, positives additionally satisfy the mutual-nearest dose-matching criteria; all remaining candidates, including unmatched doses of the same compound, are treated as negatives. If more than one positive candidate is present, the normalized rank and observed score (defined below) use the best-ranked positive.

The following representation families were evaluated: $\mathrm{logFC}$, per-gene population \(z\)-scores of $\mathrm{logFC}$, moderated $t$, and $\mathrm{sign}(\mathrm{logFC})$. Signed significance and $\mathrm{ashr}$ representations were evaluated in an additional retrieval analysis (Appendix~\ref{app:representation_sweep}). Candidates and baseline references were scored by both cosine similarity and negative Euclidean distance on the shared gene set, so larger scores indicate more similar signatures. Unlike Euclidean distance, cosine similarity is invariant to uniform positive rescaling of either signature.

Let \(N\) be the number of target candidates and let \(\mathrm{rank}_{+}\) be the best positive
rank among the \(N_{+}\) positive candidates, with \(\mathrm{rank}_{+}=1\) denoting the most similar candidate. We report normalized
best-positive rank
\[
\mathrm{rank}_{\mathrm{norm}} = 1 - \frac{\mathrm{rank}_{+}-1}{N-1},
\]
so \(\mathrm{rank}_{\mathrm{norm}}=1\) is the best possible retrieval; the worst attainable score is 
\(\frac{N_{+}-1}{N-1}\), which equals zero when $N_{+}=1$. AUROC measures the probability that a randomly selected positive outranks a randomly chosen negative, with half credit for ties; it equals \(0.5\) at chance, and it is the statistic we use to assess above-chance retrieval.

The score-level statistic is computed using the individual peers of Appendix~\ref{app:peer_baseline}, selected from either dataset. Source-side references are formed from rows in the query's own dataset at its cell line, time, and dose, but with a different compound, whereas target-side references are formed from candidates in the target pool matching the cell line, time, and dose of the best observed positive, excluding the query compound. The observed score is the similarity to the best positive. For peers, the corrected percentile of Appendix~\ref{app:peer_baseline}, computed separately for each reference side, summarizes the fraction of peer similarities below the observed score with an add-one correction.

At the rank level, the centroid and peer baselines are compared with the candidate pool. For the centroid, the query-side same-context different-compound mean
signature is appended to the target candidate pool as an additional candidate.
The baseline score is the normalized rank of this injected baseline candidate
in the augmented pool. For peers, each peer similarity score is assigned the equivalent injected-candidate normalized rank, and the baseline is the mean of these ranks. The baseline-adjusted retrieval score is the observed same-compound normalized rank minus the baseline normalized rank. We report the source-side centroid baseline and individual-peer baselines at the rank level, together with the peer percentile, in Tables~\ref{tab:cross_dataset_retrieval_cosine_bootstrap_ci} and~\ref{tab:cross_dataset_retrieval_logfc_bootstrap_ci}.

Positive baseline-adjusted values indicate that the true same-compound partner exceeds the reference, and negative values the reverse.

\subsection{Within-dataset replicate agreement}

\begin{figure*}[t]
\centering
\includegraphics[width=0.85\textwidth]{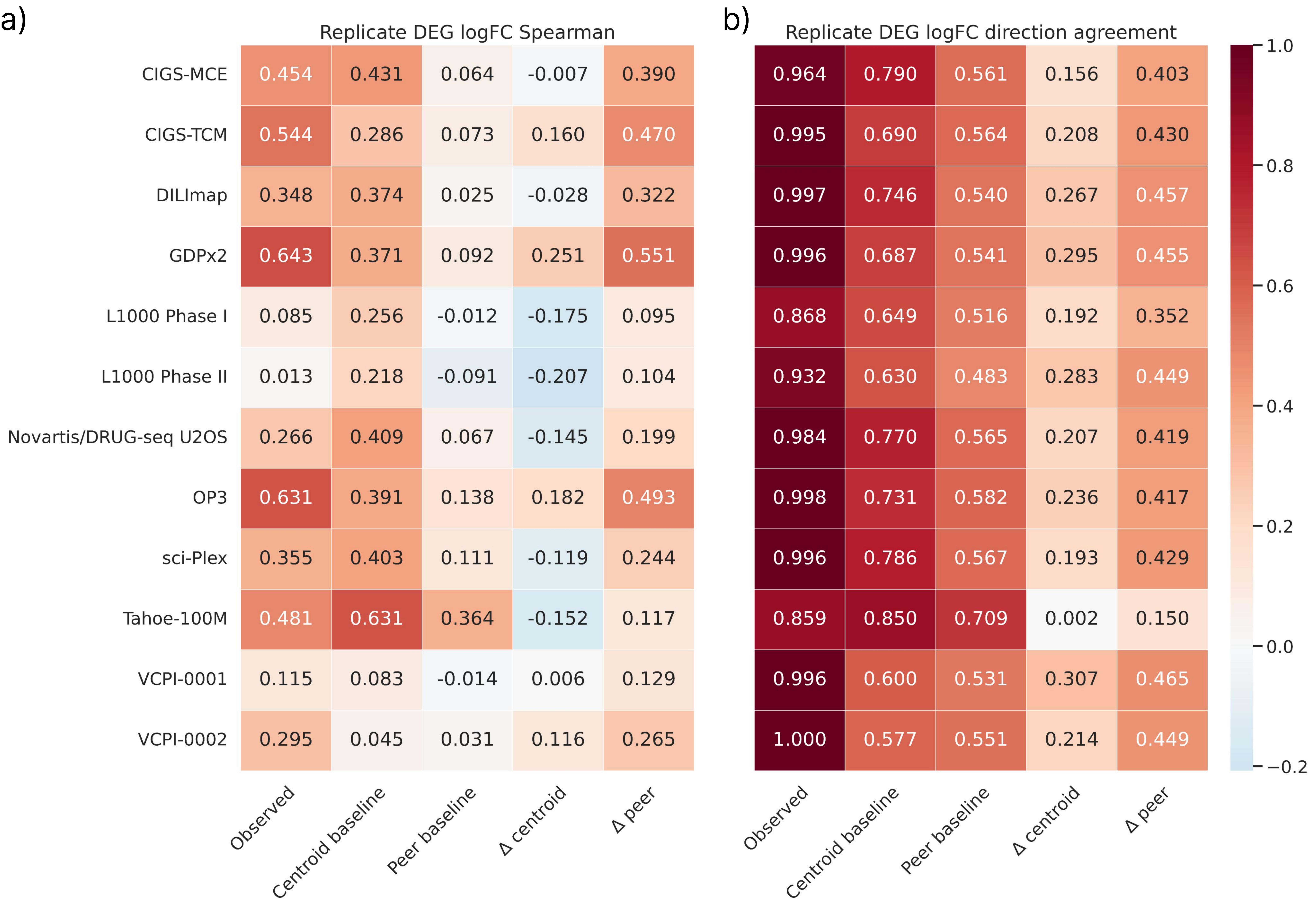}
\caption{\textbf{Within-dataset replicate agreement.}
Each row corresponds to a dataset from Table~\ref{tab:cpb_datasets}. Each panel reports the observed agreement on same-condition replicate pairs, the centroid baseline, the peer baseline, and two $\Delta$ columns giving condition-weighted means of baseline-adjusted scores. \textbf{a)} DEG-restricted (adj.\,$p < 0.05$) gene-wise Spearman correlation of log fold-changes.
\textbf{b)} DEG-restricted $\mathrm{logFC}$ direction agreement.
For DEG-restricted $\mathrm{logFC}$ Spearman, centroid baseline-adjusted replicate agreement is negative for several datasets, while peer baseline-adjusted replicate agreement is positive for all datasets. Direction agreement between replicate pairs is high in absolute terms and higher than both baselines for all assays, with the smallest margin in centroid baseline-adjusted scores observed on Tahoe-100M. Perturbagen-clustered bootstrap confidence intervals for panels a) and b) are
reported in Tables~\ref{tab:replicate_deg_lfc_spearman_bootstrap_ci}
and~\ref{tab:replicate_direction_agreement_bootstrap_ci}, respectively.}
\label{fig:cpb_replicate}
\end{figure*}

Within-dataset replicate agreement was computed from \texttt{sep\_rep} differential-expression outputs. Eligible rows were non-controls with non-empty \texttt{cell\_type}, normalized \texttt{pubchem\_cid}, time, and dose. Rows were grouped into conditions by \texttt{cell\_type}, \texttt{pubchem\_cid}, formatted time key, and formatted dose key. A condition was retained when it had at least two replicate differential-expression rows.

For a condition with two replicate rows, the single pair was scored. For a condition with \(n>2\) replicate rows, the implementation sampled \(n\) adjacent cyclic pairs after sorting by source path and source row position: \((1,2),(2,3),\ldots,(n,1)\). This keeps compute bounded while ensuring each replicate contributes to the internal reference.

Replicate metrics use the same main signature representations and gene handling as the cross-dataset signature analysis. Local metrics use genes shared by the source files needed for that condition. Global replicate metrics use the line-specific multi-dataset gene intersection. The replicate summary includes DEG-restricted Spearman and all-gene Spearman correlations on $\mathrm{logFC}$, two per-gene $z$-score scales and moderated $t$, and direction agreement.

Replicate agreement is referenced to the centroid baseline of Appendix~\ref{app:centroid_baseline} and the peer baseline of Appendix~\ref{app:peer_baseline}, formed within the dataset. DEG replicate centroid baselines use the same sample-side DEG convention as the cross-dataset DEG baseline.

Results for within-dataset strict retrieval are also reported in the code repository. Replicate retrieval uses the centroid baseline.

\subsection{Score aggregation}

Cross-dataset signature and DEG scores were first computed at the matched sample-pair level. When multiple matched doses were present for the same compound-cell-line-time case, those dose-level scores were averaged within the \texttt{pubchem\_cid}+\texttt{cell\_type}+\texttt{time\_key} group. Dataset-pair and dataset-pair-line summaries were then computed by averaging these drug-line-time summaries, so a compound with several matched doses does not dominate solely because it has a denser dose grid.

Cross-dataset retrieval was first averaged over queries within each dataset-pair, direction, cell-line, time, representation, and retrieval-variant stratum, separately for each similarity metric. Direction-specific dataset-pair summaries average over line-time strata; symmetric dataset-pair summaries then average the two directions. For the additional representation cross-dataset retrieval sweep, scores were averaged over queries within each directed dataset pair, then equally across directed pairs. Within-dataset replicate summaries were computed at the retained condition level and then averaged by dataset and by dataset-line. Heatmap cells are left blank when the corresponding pair or dataset lacks enough retained conditions or replicate-supported conditions to define the metric.

\subsection{Perturbation strength and cross-dataset agreement}
\label{app:perturbation_strength_agreement}

We additionally asked whether stronger perturbation effects are associated with
higher cross-dataset agreement. Starting from the matched sample-pair similarity
table, we aggregated rows to matched condition-dose units defined by dataset
pair, cell context, time, PubChem CID, matched dose pair, and matched condition
key. For each matched condition-dose unit, we averaged the mean absolute
moderated $t$-statistic across the two datasets and compared
$\log_{10}$ mean absolute moderated $t$ to the observed cross-dataset
Spearman correlations of $\mathrm{logFC}$ and moderated-$t$ signatures.

\begin{figure*}[t]
\centering
\includegraphics[width=\textwidth]{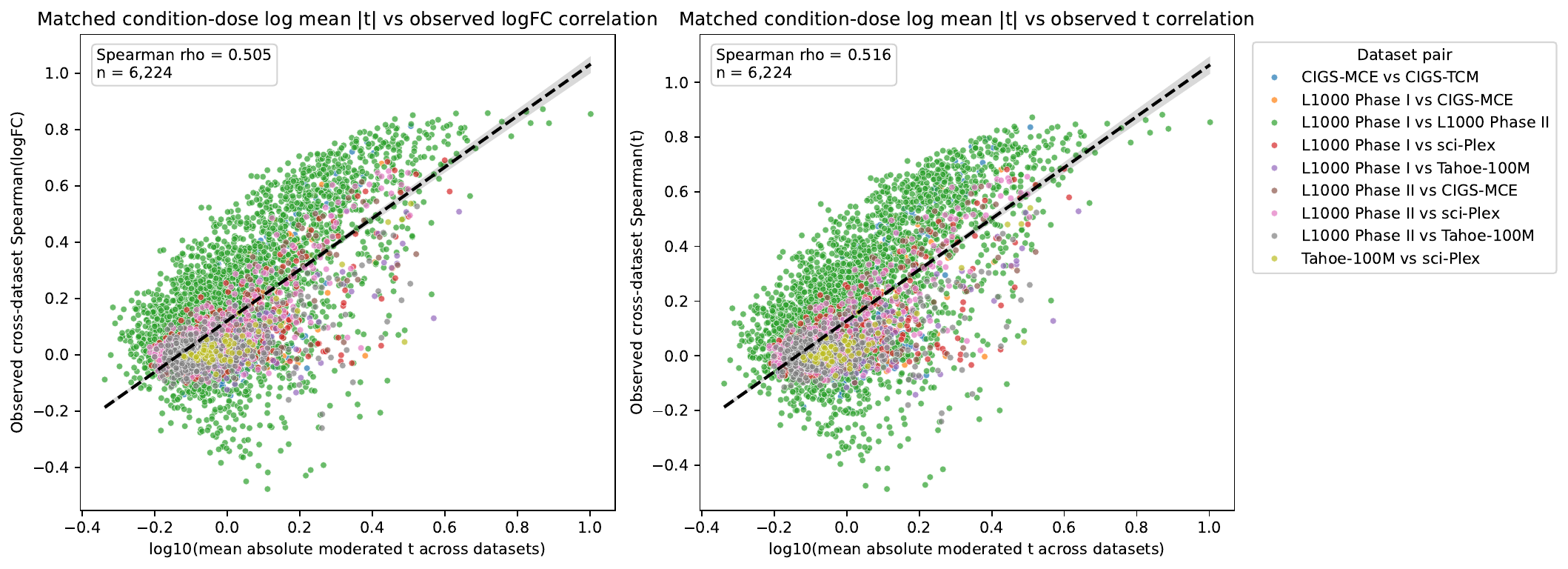}
\caption{\textbf{Perturbation strength is associated with cross-dataset signature similarity.}
Each point is a matched condition-dose unit aggregated from matched sample-pair
similarity scores, defined by dataset pair, cell context, time, PubChem CID,
matched dose pair, and matched condition key. The $x$-axis shows
$\log_{10}$ mean absolute moderated $t$ across the two matched datasets, used as
a proxy for perturbation strength. The left panel compares this quantity with
observed cross-dataset Spearman correlation of $\mathrm{logFC}$ signatures; the right panel
compares it with observed cross-dataset Spearman correlation of moderated-$t$
signatures. Dashed lines show linear trends for visual orientation. Stronger
perturbation effects are associated with higher cross-dataset similarity, but
the broad scatter indicates that perturbation strength is only a partial
explanation of agreement. This extends the previously reported strength-reproducibility association~\cite{lim_evaluation_2021} to the cross-assay setting.}
\label{fig:strength_vs_similarity}
\end{figure*}

\begin{figure*}[t]
\centering
\includegraphics[width=\textwidth]{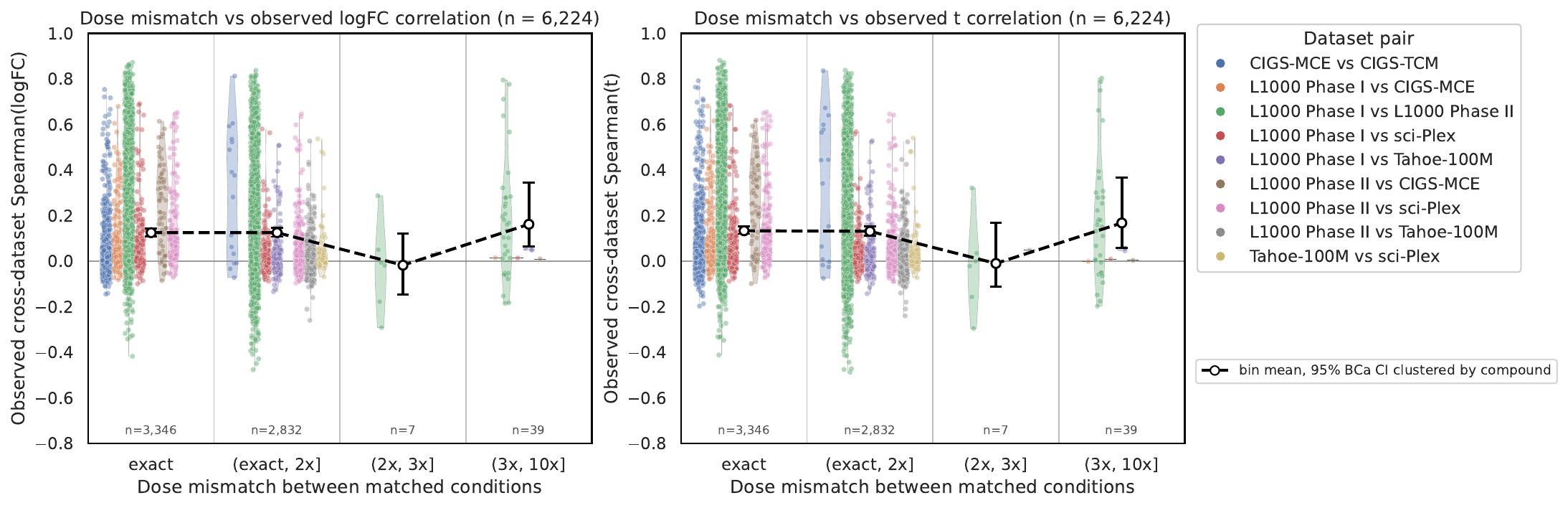}
\caption{\textbf{Dose mismatch versus cross-dataset signature agreement.} Each point
is a matched condition-dose unit aggregated from matched sample-pair similarity scores, defined by
dataset pair, cell context, time, PubChem CID, matched dose pair, and matched condition key. The $x$-axis groups these units by the fold difference between the two matched doses, averaged within a unit, and within each group points are offset by dataset pair. Shaded areas are kernel
density estimates per dataset pair, each scaled to equal width. The left panel shows observed cross-dataset Spearman correlation of $\mathrm{logFC}$ signatures; the right panel shows observed cross-dataset Spearman correlation of moderated-$t$ signatures. White circles connected by dashed lines give the bin mean with 95\% bias-corrected and accelerated bootstrap confidence intervals clustered by compound. Dose mismatch has little association with cross-dataset signature similarity, though the $(2\times, 3\times]$ and $(3\times, 10\times]$ bins contain few units.}
\label{fig:dose_mismatch_vs_similarity}
\end{figure*}

Figure~\ref{fig:strength_vs_similarity} shows that stronger perturbation
signals tend to have higher cross-dataset similarity, with Spearman
correlations of 0.505 for $\mathrm{logFC}$ similarity and 0.516 for moderated-$t$
similarity across 6{,}224 matched condition-dose points. The broad scatter
indicates that perturbation strength explains only part of the variability in cross-dataset agreement; assay, dataset, context, and
compound-specific effects likely also contribute. Figure~\ref{fig:dose_mismatch_vs_similarity} shows little overall association between dose-matching and cross-dataset signature similarity.

\subsection{Bootstrap uncertainty for cross-dataset agreement}

\label{app:cross_dataset_bootstrap_uncertainty}

For cross-dataset agreement summaries, we estimated uncertainty using a
perturbagen-clustered bootstrap applied separately to each evaluated signature representation. Within each dataset-pair stratum, PubChem CIDs
were sampled with replacement, and all associated matched condition rows for
each sampled compound were retained. For each bootstrap replicate, we recomputed
the same aggregate statistic used in the corresponding heatmap cell. Intervals
are 95\% BCa bootstrap confidence intervals computed from 2{,}000 bootstrap
iterations. For baseline-adjusted metrics, the adjusted score was bootstrapped
directly rather than bootstrapping observed and baseline scores independently; this applies to both the centroid and peer baselines, and the peer percentile of Appendix~\ref{app:peer_baseline} was bootstrapped in the same way.

These intervals quantify uncertainty over the observed matched compound panel
in each dataset pair, not over broader chemical space. Tables~\ref{tab:cross_dataset_deg_lfc_spearman_bootstrap_ci} and
\ref{tab:cross_dataset_direction_agreement_bootstrap_ci} report dataset-pair intervals for the cross-dataset agreement metrics, and Table~\ref{tab:cross_dataset_retrieval_cosine_bootstrap_ci} reports the same perturbagen-clustered bootstrap analysis for strict cross-dataset retrieval using cosine similarity; results on $\mathrm{logFC}$ correspond to Figure~\ref{fig:cpb_overview}a--p. Table~\ref{tab:cross_dataset_retrieval_logfc_bootstrap_ci} reports the same perturbagen-clustered bootstrap analysis for strict cross-dataset retrieval under negative Euclidean distance, giving intervals for the values shown in Figure~\ref{fig:cpb_overview_l2}. Retrieval is reported as normalized best-positive rank, where higher values indicate better same-compound recovery.

For completeness, Table~\ref{tab:cross_dataset_all_gene_logfc_spearman_bootstrap_ci}
reports the perturbagen-clustered bootstrap analysis for all-gene
cross-dataset Spearman agreement. This metric uses all shared genes in
each matched comparison rather than restricting to genes called differentially expressed.

\begin{table*}[t]
\caption{\textbf{Bootstrap intervals for cross-dataset DEG-restricted Spearman agreement.} Values for $\mathrm{logFC}$ correspond to Figure~\ref{fig:cpb_overview}a--d. The table also includes values for other representations: the two per-gene $z$-score scales and moderated $t$. Means are condition-weighted; intervals are 95\% BCa bootstrap confidence intervals clustered by PubChem CID. The DEG set is defined by adjusted $p<0.05$. Of the four $n_{\mathrm{cmpd}}$ columns, $\mathrm{obs}$ counts PubChem CID clusters with finite observed and baseline-adjusted scores; $\mathrm{cent_{src}}$, $\mathrm{peer_{src}}$ and $\mathrm{peer_{tgt}}$ count clusters with finite source-side centroid, source peer and target peer baseline scores. $\Delta$ columns are observed minus the named baseline; peer percentiles (pct.\ source peer and pct.\ target peer) give the corrected fraction of the corresponding peer distribution that the observed score exceeds. Each estimate is the point (first line) and its 95\% CI (\textcolor{gray}{gray}, second line). The peer cap is 256 for all representations.}
\label{tab:cross_dataset_deg_lfc_spearman_bootstrap_ci}
\centering
\tiny
\setlength{\tabcolsep}{2pt}
\renewcommand{\arraystretch}{0.95}
\resizebox{\textwidth}{!}{%
 \\

\bottomrule
\end{tabular}
}
\end{table*}

\begin{table*}[t]
\caption{\textbf{Bootstrap intervals for cross-dataset DEG-restricted $\mathbf{logFC}$ direction agreement.} Values correspond to Figure~\ref{fig:cpb_overview}e--h. Means are condition-weighted; intervals are 95\% BCa bootstrap confidence intervals clustered by PubChem CID. Direction agreement is computed on the DEG-restricted set defined by adjusted $p<0.05$. Of the two $n_{\mathrm{cmpd}}$ columns, $\mathrm{obs}$ / $\mathrm{peers}$ counts PubChem CID clusters with finite observed and baseline-adjusted scores, which are also the clusters with finite source peer and target
peer baseline scores; $\mathrm{cent_{src}}$ counts clusters with finite source-side centroid scores. The peer scores are finite on the same clusters as $\mathrm{obs}$, so the column is labeled $\mathrm{obs}$ / $\mathrm{peers}$. $\Delta$ columns are observed minus the named baseline; peer percentiles (pct.\ source peer and pct.\ target peer) give the corrected fraction of the corresponding peer distribution that the observed score exceeds. Each estimate is the point (first line) and its 95\% CI (\textcolor{gray}{gray}, second line). The peer cap is 256.}
\label{tab:cross_dataset_direction_agreement_bootstrap_ci}
\centering
\tiny
\setlength{\tabcolsep}{2pt}
\renewcommand{\arraystretch}{0.95}
\resizebox{\textwidth}{!}{%
\begin{tabular}{@{}>{\raggedright\arraybackslash}p{2.75cm}@{\hspace{1.5pt}}c@{\hspace{10pt}}c@{\hspace{14pt}}*{9}{>{\raggedright\arraybackslash}p{1.62cm}}@{}}
\toprule
Dataset pair & \multicolumn{2}{c@{\hspace{14pt}}}{$n_{\mathrm{cmpd}}$} & Observed & Source centroid & Source peer & Target peer & $\Delta$ source centroid & $\Delta$ source peer & $\Delta$ target peer & Pct.\ source peer & Pct.\ target peer \\
\cmidrule(l{2pt}r{16pt}){2-3}
 & $\mathrm{obs}$ / $\mathrm{peers}$ & $\mathrm{cent_{src}}$ & \textcolor{gray}{95\% CI} & \textcolor{gray}{95\% CI} & \textcolor{gray}{95\% CI} & \textcolor{gray}{95\% CI} & \textcolor{gray}{95\% CI} & \textcolor{gray}{95\% CI} & \textcolor{gray}{95\% CI} & \textcolor{gray}{95\% CI} & \textcolor{gray}{95\% CI} \\
\midrule
CIGS-MCE--CIGS-TCM & 161 & 846 & \begin{tabular}[t]{@{}l@{}}0.923\\\textcolor{gray}{[0.888, 0.947]}\end{tabular} & \begin{tabular}[t]{@{}l@{}}0.736\\\textcolor{gray}{[0.720, 0.753]}\end{tabular} & \begin{tabular}[t]{@{}l@{}}0.559\\\textcolor{gray}{[0.555, 0.565]}\end{tabular} & \begin{tabular}[t]{@{}l@{}}0.531\\\textcolor{gray}{[0.528, 0.535]}\end{tabular} & \begin{tabular}[t]{@{}l@{}}0.164\\\textcolor{gray}{[0.129, 0.198]}\end{tabular} & \begin{tabular}[t]{@{}l@{}}0.364\\\textcolor{gray}{[0.332, 0.389]}\end{tabular} & \begin{tabular}[t]{@{}l@{}}0.393\\\textcolor{gray}{[0.360, 0.416]}\end{tabular} & \begin{tabular}[t]{@{}l@{}}0.873\\\textcolor{gray}{[0.834, 0.903]}\end{tabular} & \begin{tabular}[t]{@{}l@{}}0.884\\\textcolor{gray}{[0.846, 0.912]}\end{tabular} \\
\addlinespace[1pt]
L1000 Phase I--CIGS-MCE & 55 & 121 & \begin{tabular}[t]{@{}l@{}}0.873\\\textcolor{gray}{[0.794, 0.916]}\end{tabular} & \begin{tabular}[t]{@{}l@{}}0.818\\\textcolor{gray}{[0.788, 0.841]}\end{tabular} & \begin{tabular}[t]{@{}l@{}}0.628\\\textcolor{gray}{[0.614, 0.643]}\end{tabular} & \begin{tabular}[t]{@{}l@{}}0.561\\\textcolor{gray}{[0.551, 0.570]}\end{tabular} & \begin{tabular}[t]{@{}l@{}}0.053\\\textcolor{gray}{[-0.029, 0.099]}\end{tabular} & \begin{tabular}[t]{@{}l@{}}0.245\\\textcolor{gray}{[0.174, 0.288]}\end{tabular} & \begin{tabular}[t]{@{}l@{}}0.312\\\textcolor{gray}{[0.240, 0.358]}\end{tabular} & \begin{tabular}[t]{@{}l@{}}0.864\\\textcolor{gray}{[0.768, 0.916]}\end{tabular} & \begin{tabular}[t]{@{}l@{}}0.902\\\textcolor{gray}{[0.816, 0.953]}\end{tabular} \\
\addlinespace[1pt]
L1000 Phase I--L1000 Phase II & 211 & 256 & \begin{tabular}[t]{@{}l@{}}0.917\\\textcolor{gray}{[0.902, 0.930]}\end{tabular} & \begin{tabular}[t]{@{}l@{}}0.724\\\textcolor{gray}{[0.711, 0.736]}\end{tabular} & \begin{tabular}[t]{@{}l@{}}0.571\\\textcolor{gray}{[0.565, 0.576]}\end{tabular} & \begin{tabular}[t]{@{}l@{}}0.545\\\textcolor{gray}{[0.540, 0.550]}\end{tabular} & \begin{tabular}[t]{@{}l@{}}0.156\\\textcolor{gray}{[0.141, 0.172]}\end{tabular} & \begin{tabular}[t]{@{}l@{}}0.347\\\textcolor{gray}{[0.333, 0.360]}\end{tabular} & \begin{tabular}[t]{@{}l@{}}0.373\\\textcolor{gray}{[0.360, 0.385]}\end{tabular} & \begin{tabular}[t]{@{}l@{}}0.928\\\textcolor{gray}{[0.912, 0.941]}\end{tabular} & \begin{tabular}[t]{@{}l@{}}0.940\\\textcolor{gray}{[0.924, 0.953]}\end{tabular} \\
\addlinespace[1pt]
L1000 Phase I--sci-Plex & 46 & 90 & \begin{tabular}[t]{@{}l@{}}0.890\\\textcolor{gray}{[0.845, 0.922]}\end{tabular} & \begin{tabular}[t]{@{}l@{}}0.724\\\textcolor{gray}{[0.693, 0.750]}\end{tabular} & \begin{tabular}[t]{@{}l@{}}0.570\\\textcolor{gray}{[0.562, 0.579]}\end{tabular} & \begin{tabular}[t]{@{}l@{}}0.535\\\textcolor{gray}{[0.529, 0.547]}\end{tabular} & \begin{tabular}[t]{@{}l@{}}0.131\\\textcolor{gray}{[0.075, 0.179]}\end{tabular} & \begin{tabular}[t]{@{}l@{}}0.320\\\textcolor{gray}{[0.272, 0.356]}\end{tabular} & \begin{tabular}[t]{@{}l@{}}0.355\\\textcolor{gray}{[0.315, 0.390]}\end{tabular} & \begin{tabular}[t]{@{}l@{}}0.873\\\textcolor{gray}{[0.809, 0.915]}\end{tabular} & \begin{tabular}[t]{@{}l@{}}0.905\\\textcolor{gray}{[0.845, 0.942]}\end{tabular} \\
\addlinespace[1pt]
L1000 Phase I--Tahoe-100M & 26 & 85 & \begin{tabular}[t]{@{}l@{}}0.736\\\textcolor{gray}{[0.609, 0.832]}\end{tabular} & \begin{tabular}[t]{@{}l@{}}0.731\\\textcolor{gray}{[0.690, 0.765]}\end{tabular} & \begin{tabular}[t]{@{}l@{}}0.589\\\textcolor{gray}{[0.575, 0.602]}\end{tabular} & \begin{tabular}[t]{@{}l@{}}0.509\\\textcolor{gray}{[0.502, 0.515]}\end{tabular} & \begin{tabular}[t]{@{}l@{}}-0.018\\\textcolor{gray}{[-0.157, 0.080]}\end{tabular} & \begin{tabular}[t]{@{}l@{}}0.147\\\textcolor{gray}{[0.016, 0.240]}\end{tabular} & \begin{tabular}[t]{@{}l@{}}0.227\\\textcolor{gray}{[0.095, 0.324]}\end{tabular} & \begin{tabular}[t]{@{}l@{}}0.742\\\textcolor{gray}{[0.579, 0.855]}\end{tabular} & \begin{tabular}[t]{@{}l@{}}0.813\\\textcolor{gray}{[0.647, 0.911]}\end{tabular} \\
\addlinespace[1pt]
L1000 Phase II--CIGS-MCE & 50 & 68 & \begin{tabular}[t]{@{}l@{}}0.919\\\textcolor{gray}{[0.877, 0.949]}\end{tabular} & \begin{tabular}[t]{@{}l@{}}0.836\\\textcolor{gray}{[0.813, 0.856]}\end{tabular} & \begin{tabular}[t]{@{}l@{}}0.647\\\textcolor{gray}{[0.633, 0.661]}\end{tabular} & \begin{tabular}[t]{@{}l@{}}0.586\\\textcolor{gray}{[0.574, 0.597]}\end{tabular} & \begin{tabular}[t]{@{}l@{}}0.094\\\textcolor{gray}{[0.051, 0.131]}\end{tabular} & \begin{tabular}[t]{@{}l@{}}0.272\\\textcolor{gray}{[0.233, 0.301]}\end{tabular} & \begin{tabular}[t]{@{}l@{}}0.333\\\textcolor{gray}{[0.291, 0.359]}\end{tabular} & \begin{tabular}[t]{@{}l@{}}0.909\\\textcolor{gray}{[0.829, 0.949]}\end{tabular} & \begin{tabular}[t]{@{}l@{}}0.936\\\textcolor{gray}{[0.859, 0.972]}\end{tabular} \\
\addlinespace[1pt]
L1000 Phase II--sci-Plex & 50 & 82 & \begin{tabular}[t]{@{}l@{}}0.949\\\textcolor{gray}{[0.911, 0.969]}\end{tabular} & \begin{tabular}[t]{@{}l@{}}0.683\\\textcolor{gray}{[0.647, 0.708]}\end{tabular} & \begin{tabular}[t]{@{}l@{}}0.574\\\textcolor{gray}{[0.566, 0.582]}\end{tabular} & \begin{tabular}[t]{@{}l@{}}0.543\\\textcolor{gray}{[0.537, 0.550]}\end{tabular} & \begin{tabular}[t]{@{}l@{}}0.186\\\textcolor{gray}{[0.151, 0.225]}\end{tabular} & \begin{tabular}[t]{@{}l@{}}0.375\\\textcolor{gray}{[0.340, 0.395]}\end{tabular} & \begin{tabular}[t]{@{}l@{}}0.406\\\textcolor{gray}{[0.374, 0.425]}\end{tabular} & \begin{tabular}[t]{@{}l@{}}0.912\\\textcolor{gray}{[0.859, 0.942]}\end{tabular} & \begin{tabular}[t]{@{}l@{}}0.932\\\textcolor{gray}{[0.887, 0.959]}\end{tabular} \\
\addlinespace[1pt]
L1000 Phase II--Tahoe-100M & 32 & 110 & \begin{tabular}[t]{@{}l@{}}0.675\\\textcolor{gray}{[0.565, 0.757]}\end{tabular} & \begin{tabular}[t]{@{}l@{}}0.686\\\textcolor{gray}{[0.656, 0.710]}\end{tabular} & \begin{tabular}[t]{@{}l@{}}0.592\\\textcolor{gray}{[0.578, 0.607]}\end{tabular} & \begin{tabular}[t]{@{}l@{}}0.502\\\textcolor{gray}{[0.494, 0.509]}\end{tabular} & \begin{tabular}[t]{@{}l@{}}-0.093\\\textcolor{gray}{[-0.204, -0.005]}\end{tabular} & \begin{tabular}[t]{@{}l@{}}0.083\\\textcolor{gray}{[-0.026, 0.171]}\end{tabular} & \begin{tabular}[t]{@{}l@{}}0.173\\\textcolor{gray}{[0.068, 0.265]}\end{tabular} & \begin{tabular}[t]{@{}l@{}}0.639\\\textcolor{gray}{[0.516, 0.748]}\end{tabular} & \begin{tabular}[t]{@{}l@{}}0.720\\\textcolor{gray}{[0.593, 0.829]}\end{tabular} \\
\addlinespace[1pt]
Tahoe-100M--sci-Plex & 8 & 23 & \begin{tabular}[t]{@{}l@{}}0.896\\\textcolor{gray}{[0.830, 0.946]}\end{tabular} & \begin{tabular}[t]{@{}l@{}}0.832\\\textcolor{gray}{[0.791, 0.878]}\end{tabular} & \begin{tabular}[t]{@{}l@{}}0.607\\\textcolor{gray}{[0.588, 0.626]}\end{tabular} & \begin{tabular}[t]{@{}l@{}}0.517\\\textcolor{gray}{[0.500, 0.537]}\end{tabular} & \begin{tabular}[t]{@{}l@{}}0.083\\\textcolor{gray}{[0.003, 0.132]}\end{tabular} & \begin{tabular}[t]{@{}l@{}}0.289\\\textcolor{gray}{[0.211, 0.346]}\end{tabular} & \begin{tabular}[t]{@{}l@{}}0.379\\\textcolor{gray}{[0.313, 0.418]}\end{tabular} & \begin{tabular}[t]{@{}l@{}}0.920\\\textcolor{gray}{[0.781, 0.989]}\end{tabular} & \begin{tabular}[t]{@{}l@{}}0.957\\\textcolor{gray}{[0.787, 1.000]}\end{tabular} \\
\bottomrule
\end{tabular}
}
\end{table*}

\begin{table*}[t]
\caption{\textbf{Bootstrap intervals for cross-dataset all-gene Spearman agreement.} Values are condition-weighted means of Spearman correlations between cross-dataset signatures, computed over all shared genes in each matched comparison rather than restricting to differentially expressed genes. Intervals are 95\% BCa bootstrap confidence intervals clustered by PubChem CID. $n_{\mathrm{cmpd}}$ is the number of PubChem CID clusters with a finite score, which is the same for the observed, source-side centroid, source peer and target peer panels. Baseline values compare each matched signature to the corresponding same-context different-compound baseline; $\Delta$ columns are observed minus the named baseline; peer percentiles (pct.\ source peer and pct.\ target peer) give the corrected fraction of the corresponding peer distribution that the observed score exceeds. Each estimate is the point (first line) and its 95\% CI (\textcolor{gray}{gray}, second line). The peer cap is 256 for all representations.}
\label{tab:cross_dataset_all_gene_logfc_spearman_bootstrap_ci}
\centering
\tiny
\setlength{\tabcolsep}{2pt}
\renewcommand{\arraystretch}{0.95}
\resizebox{\textwidth}{!}{%
 \\
\bottomrule
\end{tabular}
}
\end{table*}

\subsection{Representation sweep}
\label{app:representation_sweep}

We additionally compared signature representations in a separate
cross-dataset retrieval sweep, which covers 18 directed dataset pairs and evaluates candidates by cosine similarity, Pearson correlation, and MRRMSE (Table~\ref{tab:cross_dataset_retrieval_representation_ranking_min_support}). The table presents 7 representation families listed in Appendix~\ref{app:representations} with each representation reported at its best metric.

Unlike the primary retrieval described in Appendix~\ref{app:retrieval_cross_dataset}, candidates include all target cell lines, timepoints, and doses. For each query, a single positive is selected with the same PubChem CID and cell line, at the closest timepoint and dose; all other candidates are negatives. Retrieval is scored as the normalized rank of this positive.

Moderated \(t\) and signed significance achieve the highest mean retrieval scores, 0.669 and 0.667, respectively, compared with 0.656 for $\mathrm{logFC}$. They exceed $\mathrm{logFC}$ on 14 and 13 of the 18 directed pairs. The $\mathrm{ashr}$ posterior mean ranks third at 0.659. Direction scores slightly below $\mathrm{logFC}$, while unsigned \(p\)-values are the worst.

\begin{table*}[t]
\caption{\textbf{Cross-dataset retrieval by signature representation.}
18 directed pairs, matching 9 combinations of Table~\ref{tab:cross_dataset_deg_lfc_spearman_bootstrap_ci}. Each representation is at its best metric. Mean retrieval score is an unweighted mean over directed pairs. Beats $\mathrm{logFC}$ counts the directed pairs on which the representation exceeds $\mathrm{logFC}$ at their best metrics.}
\label{tab:cross_dataset_retrieval_representation_ranking_min_support}
\centering
\footnotesize
\setlength{\tabcolsep}{6pt}
\begin{tabular}{lrcc}
\toprule
Representation & Best metric & Mean retrieval score & Beats $\mathrm{logFC}$ \\
\midrule
moderated $t$ & Pearson & 0.669 & 14 / 18 \\
$-\log_{10}(p\text{-value})\times\mathrm{sign}(\mathrm{logFC})$ & cosine & 0.667 & 13 / 18 \\
$\mathrm{ashr}$ posterior mean$(\mathrm{logFC})$, per cell line & cosine & 0.659 & 9 / 18 \\
$\mathrm{logFC}$ & cosine & 0.656 & -- \\
dataset $z$-score & cosine & 0.655 & 10 / 18 \\
$\mathrm{sign}(\mathrm{logFC})$ & MRRMSE & 0.648 & 6 / 18 \\
dataset $\times$ cell line $z$-score & cosine & 0.646 & 9 / 18 \\
$p$-value & Pearson & 0.579 & 0 / 18 \\
\bottomrule
\end{tabular}
\end{table*}

\subsection{Bootstrap uncertainty for replicate agreement}
\label{app:bootstrap_uncertainty}

For within-dataset replicate agreement summaries, we estimated uncertainty
using a perturbagen-clustered bootstrap applied separately to each evaluated signature representation. Within each dataset stratum, PubChem
CIDs were sampled with replacement, and all associated replicate-supported
condition rows for each sampled compound were retained. For each bootstrap
replicate, we recomputed the same condition-weighted summary statistic used in
the corresponding aggregate table. Intervals are 95\% BCa bootstrap confidence
intervals computed from 2{,}000 bootstrap iterations. For baseline-adjusted
metrics, the adjusted score was bootstrapped directly rather than bootstrapping
observed and baseline scores independently; both centroid and peer baselines were used as references for replicate evaluations and reported alongside the peer percentile. These intervals quantify uncertainty over the observed compound panel in each
dataset, not over broader chemical space. The number of PubChem CID clusters
and replicate-supported condition rows is reported alongside each interval.

\begin{figure*}[t]
\centering
\begin{minipage}{0.95\textwidth}
\centering
\includegraphics[width=\textwidth]{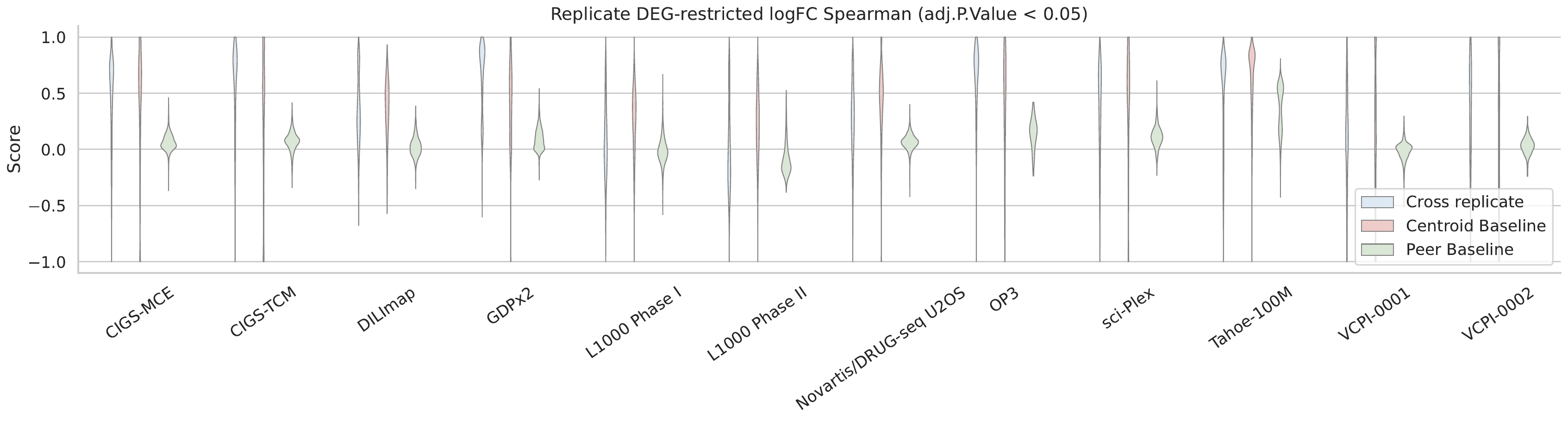}
\end{minipage}

\vspace{0.8em}

\begin{minipage}{0.95\textwidth}
\centering
\includegraphics[width=\textwidth]{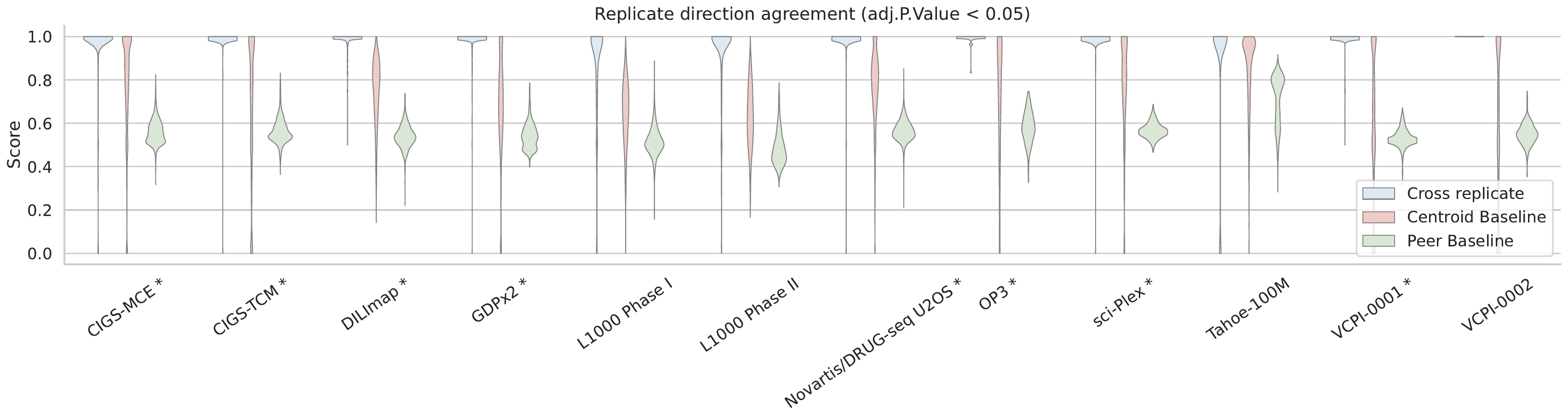}
\end{minipage}

\caption{\textbf{Distribution of within-dataset replicate agreement for the DEG-restricted metrics in Figure~\ref{fig:cpb_replicate}.}
\textbf{Top:} replicate DEG-restricted $\mathrm{logFC}$ Spearman agreement using genes with adjusted $p<0.05$.
\textbf{Bottom:} replicate $\mathrm{logFC}$ direction agreement in the same DEG-restricted setting.
For each dataset, cross-replicate scores are shown alongside the centroid and peer baseline distributions.
These distributions complement the aggregate heatmap summaries in Figure~\ref{fig:cpb_replicate}; perturbagen-clustered bootstrap confidence intervals for the corresponding condition-weighted means are reported in Tables~\ref{tab:replicate_deg_lfc_spearman_bootstrap_ci} and~\ref{tab:replicate_direction_agreement_bootstrap_ci}. Violin width is proportional to the kernel density per unit score, on a scale shared with Figure~\ref{fig:replicate_logfc_spearman_violin}. Datasets marked $^{*}$ contain at least one violin whose width is capped at the limit of that scale.}
\label{fig:replicate_deg_metric_violins}
\end{figure*}

Figure~\ref{fig:replicate_deg_metric_violins} shows the distributions of
condition-level replicate, centroid and peer baseline scores underlying the two replicate
agreement heatmaps in Figure~\ref{fig:cpb_replicate}.

Tables~\ref{tab:replicate_deg_lfc_spearman_bootstrap_ci}
and~\ref{tab:replicate_direction_agreement_bootstrap_ci}
include bootstrap intervals for the two replicate metrics shown in
Figure~\ref{fig:cpb_replicate}.

\begin{figure*}[t]
\centering
\includegraphics[width=0.95\textwidth]{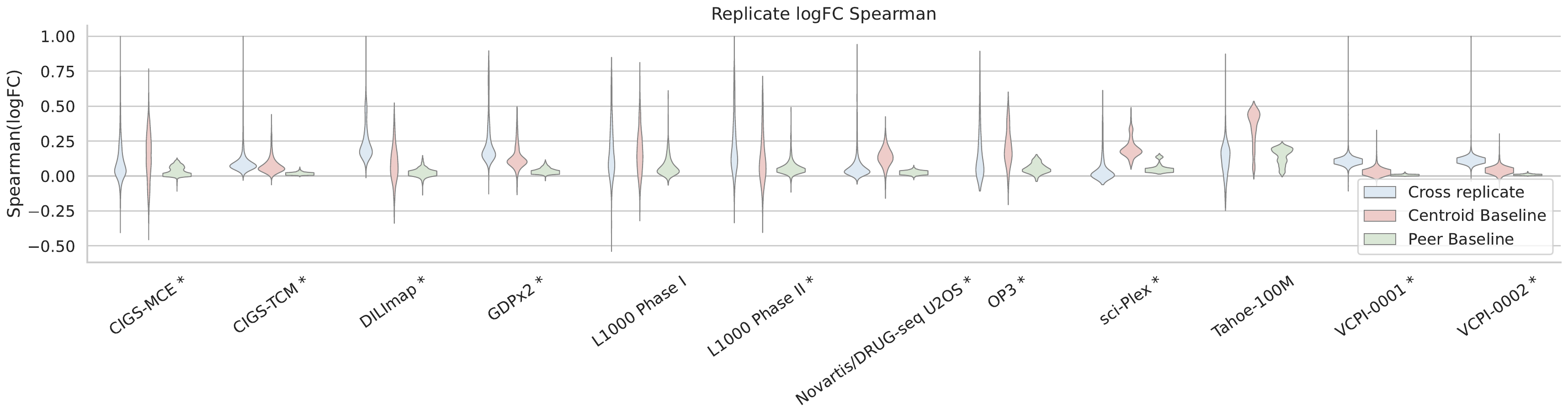}
\caption{\textbf{Distribution of within-dataset all-gene replicate $\mathbf{logFC}$ Spearman agreement.}
For each dataset, the distribution of Spearman correlations between replicate-level
$\mathrm{logFC}$ signatures is shown alongside the centroid and peer baseline distributions. This analysis uses all shared genes in
the replicate comparison rather than restricting to differentially expressed
genes. The plot complements Figure~\ref{fig:cpb_replicate}, which reports
DEG-restricted $\mathrm{logFC}$ Spearman and $\mathrm{logFC}$ direction agreement. Aggregate
condition-weighted means and perturbagen-clustered bootstrap confidence
intervals for this all-gene Spearman metric are reported in
Table~\ref{tab:replicate_logfc_bootstrap_ci}. Violin width is proportional to the kernel density per unit score, on a scale shared with Figure~\ref{fig:replicate_deg_metric_violins}. Datasets marked $^{*}$ contain at least one violin whose width is capped at the limit of that scale.}
\label{fig:replicate_logfc_spearman_violin}
\end{figure*}

Table~\ref{tab:replicate_logfc_bootstrap_ci} reports the same
bootstrap uncertainty analysis for all-gene replicate Spearman agreement; the $\mathrm{logFC}$ results correspond to Figure~\ref{fig:replicate_logfc_spearman_violin}.

\begin{table*}[t]
\caption{\textbf{Bootstrap intervals for within-dataset DEG-restricted replicate Spearman agreement.}
The four blocks give $\mathrm{logFC}$, two per-gene $z$-score scales and moderated
$t$; $\mathrm{logFC}$ values correspond to Figure~\ref{fig:cpb_replicate}a. Means are
condition-weighted; intervals are 95\% BCa bootstrap confidence intervals
clustered by PubChem CID. The DEG set is defined by adjusted $p<0.05$. $\Delta$ columns are the condition-weighted means of baseline-adjusted scores; peer percentiles (pct. peer) give the corrected fraction of the peer distribution that the observed score exceeds. Each estimate is the point (first line) and its 95\% CI (\textcolor{gray}{gray}, second line). The peer cap is 256 for all representations.}
\label{tab:replicate_deg_lfc_spearman_bootstrap_ci}
\centering
\tiny
\setlength{\tabcolsep}{2pt}
\renewcommand{\arraystretch}{0.95}
\resizebox{0.95\textwidth}{!}{%
 \\
\bottomrule
\end{tabular}
}
\end{table*}

\begin{table*}[t]
\caption{\textbf{Bootstrap intervals for within-dataset DEG-restricted replicate $\mathbf{logFC}$ direction agreement.}
Values correspond to Figure~\ref{fig:cpb_replicate}b. Means are
condition-weighted; intervals are 95\% BCa bootstrap confidence intervals
clustered by PubChem CID. Direction agreement is computed on the
DEG-restricted set defined by adjusted $p<0.05$. $\Delta$ columns are the condition-weighted means of baseline-adjusted scores; peer percentiles (pct. peer) give the corrected fraction of the peer distribution that the observed score exceeds. Each estimate is the point (first line) and its 95\% CI (\textcolor{gray}{gray}, second line). The peer cap is 256.}
\label{tab:replicate_direction_agreement_bootstrap_ci}
\centering
\tiny
\setlength{\tabcolsep}{2pt}
\renewcommand{\arraystretch}{0.95}
\resizebox{\textwidth}{!}{%
\begin{tabular}{@{}>{\raggedright\arraybackslash}p{2.75cm}@{\hspace{1.5pt}}c@{\hspace{1.5pt}}c@{\hspace{14pt}}*{6}{>{\raggedright\arraybackslash}p{2.1cm}}@{}}
\toprule
Dataset & $n_{\mathrm{cmpd}}$ & $n_{\mathrm{cond}}$ & Observed & Centroid & Peer & $\Delta$ centroid & $\Delta$ peer & Pct. peer \\
 &  &  & \textcolor{gray}{95\% CI} & \textcolor{gray}{95\% CI} & \textcolor{gray}{95\% CI} & \textcolor{gray}{95\% CI} & \textcolor{gray}{95\% CI} & \textcolor{gray}{95\% CI} \\
\midrule
\addlinespace[2pt]
CIGS-MCE & 4{,}065 & 21{,}426 & \begin{tabular}[t]{@{}l@{}}0.964\\\textcolor{gray}{[0.959, 0.968]}\end{tabular} & \begin{tabular}[t]{@{}l@{}}0.790\\\textcolor{gray}{[0.786, 0.794]}\end{tabular} & \begin{tabular}[t]{@{}l@{}}0.561\\\textcolor{gray}{[0.559, 0.562]}\end{tabular} & \begin{tabular}[t]{@{}l@{}}0.156\\\textcolor{gray}{[0.151, 0.161]}\end{tabular} & \begin{tabular}[t]{@{}l@{}}0.403\\\textcolor{gray}{[0.398, 0.407]}\end{tabular} & \begin{tabular}[t]{@{}l@{}}0.917\\\textcolor{gray}{[0.912, 0.923]}\end{tabular} \\
\addlinespace[1pt]
CIGS-TCM & 898 & 7{,}020 & \begin{tabular}[t]{@{}l@{}}0.995\\\textcolor{gray}{[0.991, 0.997]}\end{tabular} & \begin{tabular}[t]{@{}l@{}}0.690\\\textcolor{gray}{[0.681, 0.700]}\end{tabular} & \begin{tabular}[t]{@{}l@{}}0.564\\\textcolor{gray}{[0.562, 0.567]}\end{tabular} & \begin{tabular}[t]{@{}l@{}}0.208\\\textcolor{gray}{[0.198, 0.219]}\end{tabular} & \begin{tabular}[t]{@{}l@{}}0.430\\\textcolor{gray}{[0.426, 0.434]}\end{tabular} & \begin{tabular}[t]{@{}l@{}}0.910\\\textcolor{gray}{[0.901, 0.918]}\end{tabular} \\
\addlinespace[1pt]
DILImap & 296 & 1{,}184 & \begin{tabular}[t]{@{}l@{}}0.997\\\textcolor{gray}{[0.994, 0.998]}\end{tabular} & \begin{tabular}[t]{@{}l@{}}0.746\\\textcolor{gray}{[0.734, 0.757]}\end{tabular} & \begin{tabular}[t]{@{}l@{}}0.540\\\textcolor{gray}{[0.536, 0.543]}\end{tabular} & \begin{tabular}[t]{@{}l@{}}0.267\\\textcolor{gray}{[0.255, 0.279]}\end{tabular} & \begin{tabular}[t]{@{}l@{}}0.457\\\textcolor{gray}{[0.453, 0.461]}\end{tabular} & \begin{tabular}[t]{@{}l@{}}0.998\\\textcolor{gray}{[0.996, 0.999]}\end{tabular} \\
\addlinespace[1pt]
GDPx2 & 86 & 2{,}064 & \begin{tabular}[t]{@{}l@{}}0.996\\\textcolor{gray}{[0.992, 0.998]}\end{tabular} & \begin{tabular}[t]{@{}l@{}}0.687\\\textcolor{gray}{[0.670, 0.701]}\end{tabular} & \begin{tabular}[t]{@{}l@{}}0.541\\\textcolor{gray}{[0.536, 0.546]}\end{tabular} & \begin{tabular}[t]{@{}l@{}}0.295\\\textcolor{gray}{[0.280, 0.312]}\end{tabular} & \begin{tabular}[t]{@{}l@{}}0.455\\\textcolor{gray}{[0.449, 0.460]}\end{tabular} & \begin{tabular}[t]{@{}l@{}}0.972\\\textcolor{gray}{[0.964, 0.978]}\end{tabular} \\
\addlinespace[1pt]
L1000 Phase I & 265 & 2{,}517 & \begin{tabular}[t]{@{}l@{}}0.868\\\textcolor{gray}{[0.851, 0.882]}\end{tabular} & \begin{tabular}[t]{@{}l@{}}0.649\\\textcolor{gray}{[0.639, 0.658]}\end{tabular} & \begin{tabular}[t]{@{}l@{}}0.516\\\textcolor{gray}{[0.509, 0.521]}\end{tabular} & \begin{tabular}[t]{@{}l@{}}0.192\\\textcolor{gray}{[0.174, 0.208]}\end{tabular} & \begin{tabular}[t]{@{}l@{}}0.352\\\textcolor{gray}{[0.335, 0.367]}\end{tabular} & \begin{tabular}[t]{@{}l@{}}0.887\\\textcolor{gray}{[0.871, 0.902]}\end{tabular} \\
\addlinespace[1pt]
L1000 Phase II & 218 & 2{,}729 & \begin{tabular}[t]{@{}l@{}}0.932\\\textcolor{gray}{[0.919, 0.943]}\end{tabular} & \begin{tabular}[t]{@{}l@{}}0.630\\\textcolor{gray}{[0.621, 0.641]}\end{tabular} & \begin{tabular}[t]{@{}l@{}}0.483\\\textcolor{gray}{[0.478, 0.490]}\end{tabular} & \begin{tabular}[t]{@{}l@{}}0.283\\\textcolor{gray}{[0.270, 0.296]}\end{tabular} & \begin{tabular}[t]{@{}l@{}}0.449\\\textcolor{gray}{[0.436, 0.460]}\end{tabular} & \begin{tabular}[t]{@{}l@{}}0.948\\\textcolor{gray}{[0.935, 0.958]}\end{tabular} \\
\addlinespace[1pt]
Novartis/DRUG-seq U2OS & 3{,}342 & 15{,}276 & \begin{tabular}[t]{@{}l@{}}0.984\\\textcolor{gray}{[0.981, 0.986]}\end{tabular} & \begin{tabular}[t]{@{}l@{}}0.770\\\textcolor{gray}{[0.767, 0.773]}\end{tabular} & \begin{tabular}[t]{@{}l@{}}0.565\\\textcolor{gray}{[0.564, 0.567]}\end{tabular} & \begin{tabular}[t]{@{}l@{}}0.207\\\textcolor{gray}{[0.203, 0.212]}\end{tabular} & \begin{tabular}[t]{@{}l@{}}0.419\\\textcolor{gray}{[0.416, 0.421]}\end{tabular} & \begin{tabular}[t]{@{}l@{}}0.974\\\textcolor{gray}{[0.972, 0.977]}\end{tabular} \\
\addlinespace[1pt]
OP3 & 78 & 541 & \begin{tabular}[t]{@{}l@{}}0.998\\\textcolor{gray}{[0.994, 0.999]}\end{tabular} & \begin{tabular}[t]{@{}l@{}}0.731\\\textcolor{gray}{[0.689, 0.768]}\end{tabular} & \begin{tabular}[t]{@{}l@{}}0.582\\\textcolor{gray}{[0.567, 0.596]}\end{tabular} & \begin{tabular}[t]{@{}l@{}}0.236\\\textcolor{gray}{[0.197, 0.292]}\end{tabular} & \begin{tabular}[t]{@{}l@{}}0.417\\\textcolor{gray}{[0.403, 0.431]}\end{tabular} & \begin{tabular}[t]{@{}l@{}}0.965\\\textcolor{gray}{[0.947, 0.977]}\end{tabular} \\
\addlinespace[1pt]
sci-Plex & 103 & 2{,}224 & \begin{tabular}[t]{@{}l@{}}0.996\\\textcolor{gray}{[0.982, 0.999]}\end{tabular} & \begin{tabular}[t]{@{}l@{}}0.786\\\textcolor{gray}{[0.774, 0.797]}\end{tabular} & \begin{tabular}[t]{@{}l@{}}0.567\\\textcolor{gray}{[0.561, 0.572]}\end{tabular} & \begin{tabular}[t]{@{}l@{}}0.193\\\textcolor{gray}{[0.178, 0.212]}\end{tabular} & \begin{tabular}[t]{@{}l@{}}0.429\\\textcolor{gray}{[0.418, 0.435]}\end{tabular} & \begin{tabular}[t]{@{}l@{}}0.989\\\textcolor{gray}{[0.977, 0.994]}\end{tabular} \\
\addlinespace[1pt]
Tahoe-100M & 114 & 7{,}226 & \begin{tabular}[t]{@{}l@{}}0.859\\\textcolor{gray}{[0.793, 0.908]}\end{tabular} & \begin{tabular}[t]{@{}l@{}}0.850\\\textcolor{gray}{[0.822, 0.874]}\end{tabular} & \begin{tabular}[t]{@{}l@{}}0.709\\\textcolor{gray}{[0.688, 0.732]}\end{tabular} & \begin{tabular}[t]{@{}l@{}}0.002\\\textcolor{gray}{[-0.049, 0.042]}\end{tabular} & \begin{tabular}[t]{@{}l@{}}0.150\\\textcolor{gray}{[0.102, 0.193]}\end{tabular} & \begin{tabular}[t]{@{}l@{}}0.846\\\textcolor{gray}{[0.771, 0.893]}\end{tabular} \\
\addlinespace[1pt]
VCPI-0001 & 383 & 13{,}283 & \begin{tabular}[t]{@{}l@{}}0.996\\\textcolor{gray}{[0.992, 0.998]}\end{tabular} & \begin{tabular}[t]{@{}l@{}}0.600\\\textcolor{gray}{[0.593, 0.608]}\end{tabular} & \begin{tabular}[t]{@{}l@{}}0.531\\\textcolor{gray}{[0.528, 0.535]}\end{tabular} & \begin{tabular}[t]{@{}l@{}}0.307\\\textcolor{gray}{[0.287, 0.329]}\end{tabular} & \begin{tabular}[t]{@{}l@{}}0.465\\\textcolor{gray}{[0.461, 0.469]}\end{tabular} & \begin{tabular}[t]{@{}l@{}}0.952\\\textcolor{gray}{[0.940, 0.962]}\end{tabular} \\
\addlinespace[1pt]
VCPI-0002 & 248 & 8{,}825 & \begin{tabular}[t]{@{}l@{}}1.000\\\textcolor{gray}{[1.000, 1.000]}\end{tabular} & \begin{tabular}[t]{@{}l@{}}0.577\\\textcolor{gray}{[0.566, 0.587]}\end{tabular} & \begin{tabular}[t]{@{}l@{}}0.551\\\textcolor{gray}{[0.546, 0.556]}\end{tabular} & \begin{tabular}[t]{@{}l@{}}0.214\\\textcolor{gray}{[0.191, 0.241]}\end{tabular} & \begin{tabular}[t]{@{}l@{}}0.449\\\textcolor{gray}{[0.444, 0.454]}\end{tabular} & \begin{tabular}[t]{@{}l@{}}0.925\\\textcolor{gray}{[0.909, 0.940]}\end{tabular} \\
\bottomrule
\end{tabular}
}
\end{table*}

\begin{table*}[t]
\caption{\textbf{Bootstrap intervals for cross-dataset strict retrieval, cosine similarity.} Blocks give $\mathrm{logFC}$, the two per-gene $z$-score scales, moderated $t$ and $\mathrm{sign}(\mathrm{logFC})$; $\mathrm{logFC}$ values correspond to Figure~\ref{fig:cpb_overview}i--l and $\mathrm{sign}(\mathrm{logFC})$ corresponds to Figure~\ref{fig:cpb_overview}m--p. Retrieval is reported as normalized best-positive rank, where higher values indicate better recovery of the matched same-compound partner; 1.0 is best possible retrieval. Values are averaged over queries within cell-line-time strata, then equally across strata within each retrieval direction, and then across the two directions; intervals are 95\% BCa bootstrap confidence intervals clustered by PubChem CID. The source-side centroid values report the normalized rank of the within-dataset same-context different-compound baseline candidate; the source and target peer baselines score individual peers on the respective side. $n_{\mathrm{cmpd}}$ is the number of PubChem CID clusters with a finite score and $n_{\mathrm{rows}}$ the number of scored query conditions; both are the same for the observed, centroid and peer panels. $\Delta$ is the directly bootstrapped observed-minus-baseline score; peer percentiles (pct.\ source peer and pct.\ target peer) give the corrected fraction of the corresponding peer distribution that the observed score exceeds. Each estimate is the point (first line) and its 95\% CI (\textcolor{gray}{gray}, second line). For the dataset $\times$ cell-line $z$-score block, peer baselines are uncapped; for the rest, the peer cap is 256.  AUROC measures how well strictly matched positive candidates rank above the remaining candidates; $0.5$ indicates chance performance.}
\label{tab:cross_dataset_retrieval_cosine_bootstrap_ci}
\centering
\tiny
\setlength{\tabcolsep}{2pt}
\renewcommand{\arraystretch}{0.95}
\resizebox{\textwidth}{!}{%
 \\
\bottomrule
\end{tabular}
}
\end{table*}

\begin{table*}[t]
\caption{\textbf{Bootstrap intervals for cross-dataset strict retrieval, negative $L_2$ similarity.} Blocks give $\mathrm{logFC}$, the two per-gene $z$-score scales, moderated $t$ and $\mathrm{sign}(\mathrm{logFC})$; $\mathrm{logFC}$ values correspond to Figure~\ref{fig:cpb_overview_l2}a--d and $\mathrm{sign}(\mathrm{logFC})$ corresponds to Figure~\ref{fig:cpb_overview_l2}e--h. Retrieval is reported as normalized best-positive rank, where higher values indicate better recovery of the matched same-compound partner; 1.0 is best possible retrieval. Values are averaged over queries within cell-line-time strata, then equally across strata within each retrieval direction, and then across the two directions; intervals are 95\% BCa bootstrap confidence intervals clustered by PubChem CID. The source centroid values report the normalized rank of the within-dataset same-context different-compound baseline candidate; the source and target peer baselines score individual peers on the respective side, uncapped. $n_{\mathrm{cmpd}}$ is the number of PubChem CID clusters with a finite score and $n_{\mathrm{rows}}$ the number of scored query conditions; both are the same for the observed, centroid and peer panels. $\Delta$ is the directly bootstrapped observed-minus-baseline score; peer percentiles (pct.\ source peer and pct.\ target peer) give the corrected fraction of the corresponding peer distribution that the observed score exceeds. Each estimate is the point (first line) and its 95\% CI (\textcolor{gray}{gray}, second line). AUROC measures how well strictly matched positive candidates rank above the remaining candidates; $0.5$ indicates chance performance.}
\label{tab:cross_dataset_retrieval_logfc_bootstrap_ci}
\centering
\tiny
\setlength{\tabcolsep}{2pt}
\renewcommand{\arraystretch}{0.95}
\resizebox{\textwidth}{!}{%
 \\
\bottomrule
\end{tabular}
}
\end{table*}

\begin{table*}[t]
\caption{\textbf{Perturbagen-clustered bootstrap intervals for within-dataset all-gene replicate Spearman agreement.}
The four blocks give $\mathrm{logFC}$, the two per-gene $z$-score scales and moderated
$t$. Values are condition-weighted means of Spearman correlations between replicate-level signatures, computed over all shared genes in each comparison. Intervals are 95\%
BCa bootstrap confidence intervals clustered by PubChem CID.
$n_{\mathrm{cmpd}}$ is the number of PubChem CID clusters and
$n_{\mathrm{cond}}$ is the number of replicate-supported condition rows.
$\Delta$ columns are the directly bootstrapped observed-minus-baseline scores; peer percentiles (pct. peer) give the corrected fraction of the peer distribution that the observed score exceeds. Intervals quantify uncertainty over the observed compound panel. Each estimate is the point (first line) and its 95\% CI (\textcolor{gray}{gray}, second line). The peer cap is 256 for all representations.}
\label{tab:replicate_logfc_bootstrap_ci}
\centering
\tiny
\setlength{\tabcolsep}{2pt}
\renewcommand{\arraystretch}{0.95}
\resizebox{0.95\textwidth}{!}{%
 \\
\bottomrule
\end{tabular}
}
\end{table*}

\section{Representation-learning experimental details}
\label{app:additional_experiments}
The representation-learning evaluations cover two complementary tasks: compound embeddings pretrained on subsets of Chem-PerturBridge and supplied to a downstream OP3 predictor (Appendix~\ref{app:op3_evaluation}), and the multi-output model trained directly under per-dataset molecule holdout (Appendix~\ref{app:expanded_molecule_holdout_lpm}).

\subsection{Compound-embedding evaluation on the OP3 dataset}
\label{app:op3_evaluation}
The evaluation has two steps: pretraining large perturbation model-style representation learners on different slices of Chem-PerturBridge, then supplying their compound-level embeddings to the OP3 downstream predictor alongside the other compound representations. Predictions are made for OP3-held-out compounds.

\textbf{Pretraining setup.} The primary comparison involves two data slices. One slice includes only L1000 Phase I/II signatures, while the other, Chem-PerturBridge (All), incorporates perturbation-induced responses from more datasets: Tahoe, CIGS-MCE, CIGS-TCM, GDPx2, L1000 Phase I, L1000 Phase II, Novartis, VCPI-0001, VCPI-0002, and DILImap train.

To match the LPM data format and train the model, we reshape the collected datasets from AnnData into long-format Parquet tables, and filter out missing differential expression values. The rows in the resulting tables correspond to information on a perturbation experiment and a single gene response: \texttt{dataset}, \texttt{context}, \texttt{perturbation}, \texttt{log\textsubscript{10}dose}, \texttt{time}, \texttt{readout}, \texttt{value}.

We apply a rough splitting strategy for each dataset independently: 30\% of the unique (\texttt{context}, \texttt{perturbation}) pairs are randomly held out; tuples whose perturbation would otherwise appear only in the hold-out pool are swapped, where possible, with same-dataset training tuples whose perturbation is also represented in the training set of at least one other dataset to preserve learning of perturbagen embeddings. The remaining hold-out tuples are partitioned into validation and test by perturbation, with each perturbation assigned in its entirety to one of the two splits. Finally, data is partitioned into split-pure, $\leq$ 200 K-row Parquet shards, grouped under one folder per \texttt{dataset\_context}.

The model architectures for two endpoints (L1000-based and multi-dataset) are similar and are inspired by the large perturbation model. In addition to categorical inputs associated with \texttt{readout}, \texttt{perturbation} and \texttt{context} embeddings introduced in the original large perturbation model, we also added \texttt{dataset}-related embeddings and the information for \texttt{log\textsubscript{10}dose} values and perturbation \texttt{time} represented as numerical features and projected into 128-dimensional vectors, matching the dimensionality of the categorical embeddings.

We construct a readout vocabulary from Ensembl-based gene identifiers and a perturbation vocabulary from PubChem CIDs, while context categories are derived from cell-type IDs; all these identifiers are obtained within the standardization part of the Chem-PerturBridge pipeline. 

Beyond the embedding and projection layers (which produce a 768-dim representation), the architecture is a 2-hidden-layer MLP with a scalar head: 768 - 256 - 256 - 1. In our setup, the head predicts $\mathrm{logFC}$ for each perturbation sample determined by the input features mentioned above, minimizing MSE loss.

We follow the hyperparameters proposed by the authors of the LPM paper for their model on the LINCS dataset (learning rate, learning rate decay, optimizer, dropout ratio), increasing the batch size to 16{,}384 to accelerate training and reducing the number of epochs to 25. Additionally, training is distributed over 16 randomly chosen GPUs from the Helmholtz HPC cluster, with the training state checkpointed at the end of every epoch.

The code release records the training objective, architecture, hyperparameters, compound split construction, leakage checks, assay and dose metadata handling, and checkpoint selection procedure that we use for each representation-learning experiment.

The molecule-holdout experiment in
Appendix~\ref{app:expanded_molecule_holdout_lpm} uses exactly the same hyperparameters but a multi-output
training format in which each row corresponds to one perturbation sample and
the model predicts all observed genes for that sample with a masked loss.

\textbf{Downstream evaluations.} The downstream transfer evaluation uses the OP3 NeurIPS-2023 benchmark~\cite{szalata_benchmark_2024} under a compound-held-out split. The release records train, validation, and test split definitions; leakage checks against the harmonized pretraining pool; model-selection rules; and metric aggregation procedures.

Benchmark evaluation employs the top-performing method as the default reference, alongside its variant adapted to leverage embeddings. The original approach is an ensemble of small feed-forward networks that treats each sample as a categorical pair $(\mathrm{cell}\,\mathrm{type},\,\mathrm{small}\,\mathrm{molecule})$. A single shared label encoder maps both fields into one 152-token vocabulary and each network learns its own embedding table end-to-end: the cell and compound indices are looked up in the same table, concatenated, and passed through a stack of MLP blocks ending in a linear head. Targets are compressed per model by a truncated SVD, so the network regresses SVD coefficients that are mapped back to gene space at prediction time. The original version does not use any additional information on molecular features.

The method consists of two stages, each comprising an ensemble run. During the first stage, the ensemble makes predictions on the held-out $(\mathrm{cell}\,\mathrm{type},\,\mathrm{small}\,\mathrm{molecule})$ pairs; these predictions are kept as pseudolabels. In the next stage, the second ensemble is trained on the union of the training set and pseudolabels, finally releasing the output predictions. 

The extension keeps the general architecture of the original two-stage method unchanged and modifies only the embedding head of each individual network. First, the original shared 152-token embedding table is split into two separate tables, one for cell types and one for compounds. Second, the precomputed compound descriptors are combined with the learnable compound embedding by initializing that table from the descriptors and fine-tuning it end-to-end, with a learned linear projection to match the per-network embedding dimension. The cell-type embedding remains a freely learned table.

We evaluate five families of compound descriptors, each stored as a per-compound vector indexed by small-molecule name, together with a descriptor-free configuration: 
\begin{itemize}
    \item \textbf{Morgan fingerprints.} Morgan fingerprints (ECFP:2) are extended-connectivity fingerprints, computed deterministically from each compound's SMILES string. 
    \item \textbf{Chemical language models.} Embeddings from pretrained chemical models, ChemBERTa-1, ChemBERTa-2, MolFormer-XL, are used as the baseline embeddings along with Morgan fingerprints.
    \item \textbf{LPM on Chem-PerturBridge (All).} Perturbation embeddings that the LPM-based model learns when it is trained jointly on all datasets of the Chem-PerturBridge collection, excluding OP3 and sci-Plex.
    \item \textbf{LPM on Chem-PerturBridge (narrowed).} Perturbation embeddings that the multi-dataset LPM-based model learns when it is trained on the Chem-PerturBridge collection but restricted to the L1000 molecule set.
    \item \textbf{LPM on L1000.} Perturbation embeddings that the LPM-based model learns when it is trained only on the L1000 subset of Chem-PerturBridge.
    \item \textbf{Descriptor-free.} A descriptor-free configuration, in which the network sees the categorical compound indices and learns its own internal embeddings, is included as the default reference.
\end{itemize}

The OP3 dataset is not included in either pretraining slice, so no OP3 measurement is seen when the LPM-based embeddings are learned. The OP3 training and test sets are concatenated and re-split by compound with a fixed seed: 25\% of the 140 unique compounds are drawn without replacement as the new test set, and the rest are used as the new training set, so that no compound appears in both splits. Excluding OP3 from pretraining removes only OP3 measurements. Most OP3 compounds still appear in other pretraining datasets and other cellular contexts; see Table~\ref{tab:op3_vocab_coverage}. The evaluation tests whether perturbation signal learned in other datasets transfers to compounds held out from the OP3 training split.

Although the perturbation vocabulary in the LPM-style model is large (37{,}441 compounds for the multi-dataset-trained model and 21{,}005 for both the L1000-based model and the multi-dataset-trained model narrowed to L1000 molecules), not every OP3 compound has a corresponding entry. We modify the input head so that it preserves the default "learn its own embedding" behavior for the compounds without an external descriptor while incorporating the external information for the compounds that do have one. The results are reported in Table~\ref{tab:op3_unseen_compounds}.

As methods differ in the number of compounds for which embeddings exist (Table~\ref{tab:op3_vocab_coverage}), additional experiments on the OP3 evaluation with descriptor coverage on the test set matched across methods are carried out (Table~\ref{tab:op3_unseen_compounds_b}). Every method receives an external descriptor only for the 30 test compounds that had an embedding under all models, and uses a randomly initialized fallback for the remaining 5. The fallback is a standard learnable compound embedding, selected per compound by a frozen indicator, and is trained end-to-end in every configuration. Test rows, splits and seeds are unchanged.

Training is performed with 10 different seeds, and per-seed predictions are scored against the held-out OP3 test set using the evaluation metrics proposed in the OP3 benchmark paper~\cite{szalata_benchmark_2024}, including mean row-wise RMSE, MAE, Spearman correlation and cosine similarity. The 10 seeds therefore quantify optimization variance on this single fixed test set. The experiments are repeated across embeddings obtained from different LPM-like model checkpoints, allowing us to assess how the downstream OP3 metrics depend on the training duration of the multi-dataset-trained, multi-dataset-trained-narrowed, and L1000-based variants. We report results primarily from epoch 20 in Table~\ref{tab:op3_unseen_compounds}. Experiments with Morgan fingerprints and chemical model embeddings are repeated across seeds only, since neither has pretraining checkpoints of ours to vary.

Additionally, we perform step-matched comparisons to assess whether the differences between multi-dataset and L1000-based embeddings reflect unequal numbers of pretraining steps. The Chem-PerturBridge (All) multi-dataset model is trained for 255,200 steps, matching 20 epochs of L1000-only pretraining, and the L1000-based model is trained for 2,089,120 steps, matching 20 epochs of multi-dataset pretraining. Both variants use the original batch size of 16,384. The learning-rate decay factors are adjusted for the different numbers of steps per epoch to approximately 0.779 for the multi-dataset model and 0.996 for the L1000-only model.

We also evaluate covariate ablations in both multi-dataset and L1000-based variants, removing the dataset covariate, dose and time jointly, or all three covariates by zeroing and freezing their corresponding input layers. Ablation embeddings are extracted at pretraining epoch 20. Step-matched and ablation evaluations are performed with both fixed and trainable compound embeddings under native-coverage and coverage-matched downstream setups. The results are reported in Tables~\ref{tab:op3_unseen_compounds},~\ref{tab:op3_unseen_compounds_b},~\ref{tab:op3_ablation_covariates}.

We also quantify uncertainty over the held-out compounds themselves, using the same perturbagen-clustered bootstrap procedure as the agreement benchmark (Appendix~\ref{app:cross_dataset_bootstrap_uncertainty}): 95\% BCa confidence intervals, a nested mean with cluster = $\mathrm{compound}$ (small molecule) and inner strata over $\mathrm{context}$ (cell type), 2{,}000 draws, on seed-averaged row scores. Each $(\mathrm{context},\,\mathrm{compound})$ pair contributes a paired difference against the descriptor-free baseline, signed so that positive values favor the method: $\Delta=\mathrm{method}-\mathrm{descriptor\text{-}free}$ for Spearman and cosine, $\Delta=\mathrm{descriptor\text{-}free}-\mathrm{method}$ for MRRMSE and MAE. The win rate is the fraction of pairs on which the method beats the baseline, averaged under the same clustering. Each method is compared only with the descriptor-free baseline.

\textbf{Results.} Table~\ref{tab:op3_unseen_compounds}
shows that the Chem-PerturBridge-pretrained (All) embeddings, fine-tuned during OP3 training, outperform L1000-only embeddings,
Morgan fingerprints, chemical language model embeddings, and the descriptor-free OP3 baseline on all four metrics, consistently across seeds. Narrowing pretraining to L1000 molecules drops Spearman relative to Chem-PerturBridge (All) while maintaining an MRRMSE advantage over the L1000-based model.

Part of the gain over the L1000-only model reflects differences in compound coverage, since the methods differ in vocabulary size. Under matched coverage (Table~\ref{tab:op3_unseen_compounds_b}), Chem-PerturBridge pretraining still leads on all 4 metrics, but the margin over the L1000-only model is roughly halved on the correlation metrics and falls by about a third on MRRMSE.

Under unmatched coverage (Table~\ref{tab:op3_bootstrap}), LPM on Chem-PerturBridge (All) with tuning on the OP3 train set is the only method whose gains have confidence intervals excluding zero on three metrics: Spearman, cosine and MRRMSE; under matched coverage (Table~\ref{tab:op3_bootstrap_b}) no method shows a resolved gain. The losses are more stable: every fixed-embedding variant, the frozen LPM embeddings included, falls significantly below the descriptor-free baseline on MRRMSE or MAE under both coverage settings.

Step-matched pretraining experiments show a little improvement of L1000-based embeddings on MRRMSE and MAE metrics, while reducing the number of steps during multi-dataset model pretraining weakens performance; the multi-dataset model's trainable embeddings retain better mean scores on all four metrics than the 20-epoch L1000-only model under both coverage settings. However, neither trainable step-matched variant shows a resolved gain over the descriptor-free baseline: the bootstrap confidence intervals include zero for all four metrics under both coverage settings.

The covariate ablation experiments show improvements of trainable compound embeddings on the OP3 downstream task for both multi-dataset and L1000-only models (Table~\ref{tab:op3_ablation_covariates}). Dropping dose and time jointly improves results for both models while dropping the dataset covariate strengthens the L1000-based embeddings and slightly weakens the multi-dataset embeddings in comparison to ones obtained from unablated model versions. The configurations with the lowest errors are different: exclusion of dose and time covariates for multi-dataset, exclusion of the dataset covariate for the L1000-only.

Bootstrap intervals in Table~\ref{tab:op3_ablation_bootstrap} show that embeddings from the multi-dataset Chem-PerturBridge (All) model pretrained without dose and time are the only ones that show resolved gains over the descriptor-free baseline on all 4 metrics under both coverage settings. The L1000 ablations are resolved on MRRMSE alone. Win rates and the $\Delta$ values give different rankings: the highest win rates belong mostly to LPM-L1000 variants, while the largest $\Delta$ values belong to the multi-dataset variant. Thus, larger average improvements do not necessarily imply higher win rates.

\begin{table}[t]
\caption{\textbf{OP3 molecule-vocabulary coverage.}
For each descriptor, the total number of unique compounds in its vocabulary; the number of train/test compounds present in that vocabulary; the number of test compounds missing from the vocabulary and receiving a fallback representation. The narrowed Chem-PerturBridge variant is restricted to L1000 molecules and matches the L1000-based vocabulary.
Abbreviations: CB1/CB2, ChemBERTa-1/2; MF-XL, MolFormer-XL; LPM, large perturbation model; CPB, Chem-PerturBridge; nar., narrowed.}
\label{tab:op3_vocab_coverage}
\centering
\footnotesize
\setlength{\tabcolsep}{4pt}
\begin{tabular}{l cccc}
\toprule
Model / descriptor & In vocabulary & Train covered & Test covered & Test fallbacks \\
\midrule
Morgan / CB1 / CB2 / MF-XL & 140 & 105 & 35 & 0 \\
LPM-CPB (all) / LPM-CPB (all) step-matched & 138 & 104 & 34 & 1 \\
LPM-CPB (nar.) & 114 & 84 & 30 & 5 \\
LPM-L1000 / LPM-L1000 step-matched & 114 & 84 & 30 & 5 \\
\bottomrule
\end{tabular}
\end{table}

\begin{table*}[t]
\caption{\textbf{Perturbation prediction on held-out OP3 compounds.}
Mean $\pm$ standard deviation across 10 seeds under a compound-held-out OP3
re-split. The OP3 dataset is excluded from Chem-PerturBridge pretraining for this
evaluation. LPM embeddings pretrained on Chem-PerturBridge (All) and fine-tuned during OP3 training improve over
L1000-only LPM embeddings, Morgan fingerprints, embeddings from chemical language models (ChemBERTa-1, ChemBERTa-2, MolFormer-XL), and the descriptor-free (OP3 default) baseline across all metrics. Pretraining on Chem-PerturBridge and tuning on the OP3 train set but restricted to L1000 molecules, Chem-PerturBridge (narrowed) Trainable, reduces the Spearman gain, indicating that part of it reflects compound coverage. The Trainable column indicates whether the externally initialized compound embedding is updated during training on the OP3 benchmark; the Fixed setting freezes it. Step-matched rows correspond to LPM embeddings from Chem-PerturBridge (All) pretrained for only as many steps as the original L1000 run, and from L1000 pretrained for as many steps as the original All run. \textbf{Bold} marks the best value in each column and methods whose mean $\pm$ one-standard-deviation
interval overlaps the corresponding interval of the best method; rows are ordered by MRRMSE.
Abbreviations: CB1/CB2, ChemBERTa-1/2; MF-XL, MolFormer-XL; LPM, large perturbation model; CPB, Chem-PerturBridge; nar., narrowed.}
\label{tab:op3_unseen_compounds}
\centering
\scriptsize
\setlength{\tabcolsep}{3pt}
\resizebox{\textwidth}{!}{%
\begin{tabular}{lc cccc}
\toprule
Method & Trainable & Spearman $\uparrow$ & Cosine $\uparrow$ & MRRMSE $\downarrow$ & MAE $\downarrow$ \\
\midrule
CB1 & Fixed & $0.1809 \pm 0.0034$ & $0.1945 \pm 0.0036$ & $0.8741 \pm 0.0023$ & $0.6256 \pm 0.0020$ \\
LPM-CPB (all) step-matched & Fixed & $0.2025 \pm 0.0020$ & $0.2198 \pm 0.0020$ & $0.8665 \pm 0.0024$ & $0.6214 \pm 0.0020$ \\
MF-XL & Fixed & $0.2088 \pm 0.0028$ & $0.2289 \pm 0.0029$ & $0.8627 \pm 0.0034$ & $0.6215 \pm 0.0034$ \\
CB2 & Fixed & $0.2186 \pm 0.0030$ & $0.2373 \pm 0.0031$ & $0.8553 \pm 0.0028$ & $0.6131 \pm 0.0026$ \\
Morgan & Fixed & $0.2253 \pm 0.0015$ & $0.2417 \pm 0.0015$ & $0.8547 \pm 0.0021$ & $0.6139 \pm 0.0022$ \\
CB2 & Yes & $0.2335 \pm 0.0023$ & $0.2513 \pm 0.0023$ & $0.8460 \pm 0.0015$ & $0.6052 \pm 0.0016$ \\
LPM-L1000 & Fixed & $0.2372 \pm 0.0014$ & $0.2545 \pm 0.0013$ & $0.8452 \pm 0.0020$ & $0.6019 \pm 0.0019$ \\
LPM-CPB (nar.) & Fixed & $0.2221 \pm 0.0023$ & $0.2391 \pm 0.0023$ & $0.8439 \pm 0.0023$ & $0.6035 \pm 0.0021$ \\
LPM-CPB (all) & Fixed & $0.2539 \pm 0.0016$ & $0.2719 \pm 0.0016$ & $0.8435 \pm 0.0040$ & $0.6103 \pm 0.0035$ \\
MF-XL & Yes & $0.2344 \pm 0.0029$ & $0.2537 \pm 0.0030$ & $0.8402 \pm 0.0018$ & $0.5979 \pm 0.0013$ \\
LPM-L1000 step-matched & Fixed & $0.2335 \pm 0.0027$ & $0.2501 \pm 0.0028$ & $0.8371 \pm 0.0011$ & $0.5952 \pm 0.0011$ \\
CB1 & Yes & $0.2495 \pm 0.0014$ & $0.2649 \pm 0.0016$ & $0.8279 \pm 0.0012$ & $0.5851 \pm 0.0013$ \\
Descriptor-free & -- & $0.2533 \pm 0.0018$ & $0.2688 \pm 0.0018$ & $0.8161 \pm 0.0004$ & $0.5723 \pm 0.0005$ \\
LPM-L1000 & Yes & $0.2583 \pm 0.0015$ & $0.2748 \pm 0.0014$ & $0.8115 \pm 0.0010$ & $0.5724 \pm 0.0009$ \\
Morgan & Yes & $0.2638 \pm 0.0011$ & $0.2806 \pm 0.0011$ & $0.8115 \pm 0.0004$ & $0.5716 \pm 0.0008$ \\
LPM-L1000 step-matched & Yes & $0.2583 \pm 0.0016$ & $0.2743 \pm 0.0014$ & $0.8097 \pm 0.0007$ & $0.5710 \pm 0.0008$ \\
LPM-CPB (nar.) & Yes & $0.2548 \pm 0.0015$ & $0.2710 \pm 0.0014$ & $0.8089 \pm 0.0006$ & $0.5707 \pm 0.0006$ \\
LPM-CPB (all) step-matched & Yes & $0.2644 \pm 0.0015$ & $0.2809 \pm 0.0015$ & $0.8083 \pm 0.0006$ & $0.5701 \pm 0.0009$ \\
LPM-CPB (all) & Yes & $\mathbf{0.2745 \pm 0.0011}$ & $\mathbf{0.2895 \pm 0.0013}$ & $\mathbf{0.8034 \pm 0.0006}$ & $\mathbf{0.5676 \pm 0.0009}$ \\
\bottomrule
\end{tabular}%
}
\end{table*}

\begin{table*}[t]
\caption{\textbf{Perturbation prediction on held-out OP3 compounds. Coverage-matched results.}
Mean $\pm$ standard deviation across 10 seeds under a compound-held-out OP3
re-split. The OP3 dataset is excluded from Chem-PerturBridge pretraining for this
evaluation. Every method gets an external descriptor only for the 30 test compounds that have an embedding under all models, and uses a randomly initialized fallback for the remaining 5. The fallback is a standard learnable compound embedding, trained end-to-end in every configuration. The Trainable column indicates whether the externally initialized compound embedding is updated during training on the OP3 benchmark; the Fixed setting freezes it. Matching coverage lowers the Chem-PerturBridge (All) Trainable scores relative to Table~\ref{tab:op3_unseen_compounds}, but it still leads on all metrics. Step-matched rows correspond to LPM embeddings from Chem-PerturBridge (All) pretrained for only as many steps as the original L1000 run, and from L1000 pretrained for as many steps as the original All run. \textbf{Bold} marks the best value in each column and methods whose mean $\pm$ one-standard-deviation
interval overlaps the corresponding interval of the best method; rows are ordered by MRRMSE.
Abbreviations: CB1/CB2, ChemBERTa-1/2; MF-XL, MolFormer-XL; LPM, large perturbation model; CPB, Chem-PerturBridge; nar., narrowed.}
\label{tab:op3_unseen_compounds_b}
\centering
\scriptsize
\setlength{\tabcolsep}{3pt}
\resizebox{\textwidth}{!}{%
\begin{tabular}{lc cccc}
\toprule
Method & Trainable & Spearman $\uparrow$ & Cosine $\uparrow$ & MRRMSE $\downarrow$ & MAE $\downarrow$ \\
\midrule
CB1 & Fixed & $0.1979 \pm 0.0031$ & $0.2151 \pm 0.0033$ & $0.8755 \pm 0.0029$ & $0.6313 \pm 0.0025$ \\
LPM-CPB (all) step-matched & Fixed & $0.2048 \pm 0.0030$ & $0.2235 \pm 0.0030$ & $0.8645 \pm 0.0018$ & $0.6204 \pm 0.0017$ \\
MF-XL & Fixed & $0.2100 \pm 0.0025$ & $0.2301 \pm 0.0025$ & $0.8636 \pm 0.0027$ & $0.6217 \pm 0.0026$ \\
CB2 & Fixed & $0.2043 \pm 0.0029$ & $0.2233 \pm 0.0032$ & $0.8613 \pm 0.0024$ & $0.6183 \pm 0.0023$ \\
Morgan & Fixed & $0.2056 \pm 0.0024$ & $0.2204 \pm 0.0024$ & $0.8610 \pm 0.0022$ & $0.6167 \pm 0.0021$ \\
CB2 & Yes & $0.2260 \pm 0.0024$ & $0.2438 \pm 0.0025$ & $0.8493 \pm 0.0021$ & $0.6069 \pm 0.0021$ \\
LPM-CPB (all) & Fixed & $0.2484 \pm 0.0010$ & $0.2666 \pm 0.0011$ & $0.8473 \pm 0.0032$ & $0.6131 \pm 0.0029$ \\
LPM-L1000 & Fixed & $0.2372 \pm 0.0014$ & $0.2545 \pm 0.0013$ & $0.8452 \pm 0.0020$ & $0.6019 \pm 0.0019$ \\
LPM-CPB (nar.) & Fixed & $0.2221 \pm 0.0023$ & $0.2391 \pm 0.0023$ & $0.8439 \pm 0.0023$ & $0.6035 \pm 0.0021$ \\
MF-XL & Yes & $0.2370 \pm 0.0025$ & $0.2563 \pm 0.0025$ & $0.8409 \pm 0.0017$ & $0.5979 \pm 0.0014$ \\
LPM-L1000 step-matched & Fixed & $0.2335 \pm 0.0027$ & $0.2501 \pm 0.0028$ & $0.8371 \pm 0.0011$ & $0.5952 \pm 0.0011$ \\
CB1 & Yes & $0.2536 \pm 0.0014$ & $0.2695 \pm 0.0015$ & $0.8276 \pm 0.0010$ & $0.5851 \pm 0.0011$ \\
Descriptor-free & -- & $0.2533 \pm 0.0018$ & $0.2688 \pm 0.0018$ & $0.8161 \pm 0.0004$ & $0.5723 \pm 0.0005$ \\
Morgan & Yes & $0.2580 \pm 0.0013$ & $0.2739 \pm 0.0013$ & $0.8147 \pm 0.0005$ & $0.5734 \pm 0.0007$ \\
LPM-L1000 & Yes & $0.2583 \pm 0.0015$ & $0.2748 \pm 0.0014$ & $0.8115 \pm 0.0010$ & $0.5724 \pm 0.0009$ \\
LPM-L1000 step-matched & Yes & $0.2583 \pm 0.0016$ & $0.2743 \pm 0.0014$ & $0.8097 \pm 0.0007$ & $\mathbf{0.5710 \pm 0.0008}$ \\
LPM-CPB (all) step-matched & Yes & $0.2602 \pm 0.0012$ & $0.2774 \pm 0.0012$ & $0.8093 \pm 0.0006$ & $0.5713 \pm 0.0008$ \\
LPM-CPB (nar.) & Yes & $0.2548 \pm 0.0015$ & $0.2710 \pm 0.0014$ & $0.8089 \pm 0.0006$ & $\mathbf{0.5707 \pm 0.0006}$ \\
LPM-CPB (all) & Yes & $\mathbf{0.2670 \pm 0.0008}$ & $\mathbf{0.2825 \pm 0.0009}$ & $\mathbf{0.8059 \pm 0.0005}$ & $\mathbf{0.5696 \pm 0.0008}$ \\
\bottomrule
\end{tabular}%
}
\end{table*}

\begin{table*}[t]
\caption{\textbf{Covariate ablations for LPM embeddings on held-out OP3 compounds.} Mean $\pm$ standard deviation across 10 seeds under a compound-held-out OP3 re-split. The Covariates removed column indicates which Chem-PerturBridge covariates (dataset, dose+time, or dataset+dose+time) were ablated during LPM compound-embedding pretraining. The top block uses each method's native OP3 compound coverage; the bottom block is coverage-matched. The Trainable column indicates whether the externally initialized compound embedding is updated during training on the OP3 benchmark; the Fixed setting freezes it. \textbf{Bold} marks the best value in each column and methods whose mean $\pm$ one-standard-deviation
interval overlaps the corresponding interval of the best method; rows are ordered by MRRMSE. Abbreviations: LPM, large perturbation model; CPB, Chem-PerturBridge; nar., narrowed.}
\label{tab:op3_ablation_covariates}
\centering
\scriptsize
\setlength{\tabcolsep}{3pt}
\resizebox{\textwidth}{!}{%
\begin{tabular}{lcr cccc}
\toprule
Method & Trainable & Covariates removed & Spearman $\uparrow$ & Cosine $\uparrow$ & MRRMSE $\downarrow$ & MAE $\downarrow$ \\
\midrule
\multicolumn{7}{l}{\textit{native coverage}} \\
\addlinespace[2pt]
LPM-CPB (all) & Fixed & dose+time & $0.2255 \pm 0.0030$ & $0.2447 \pm 0.0031$ & $0.8621 \pm 0.0026$ & $0.6221 \pm 0.0023$ \\
LPM-CPB (all) & Fixed & dataset+dose+time & $0.2115 \pm 0.0032$ & $0.2279 \pm 0.0035$ & $0.8496 \pm 0.0023$ & $0.6094 \pm 0.0020$ \\
LPM-L1000 & Fixed & -- & $0.2372 \pm 0.0014$ & $0.2545 \pm 0.0013$ & $0.8452 \pm 0.0020$ & $0.6019 \pm 0.0019$ \\
LPM-CPB (all) & Fixed & dataset & $0.2288 \pm 0.0023$ & $0.2473 \pm 0.0023$ & $0.8443 \pm 0.0032$ & $0.6090 \pm 0.0027$ \\
LPM-CPB (nar.) & Fixed & -- & $0.2221 \pm 0.0023$ & $0.2391 \pm 0.0023$ & $0.8439 \pm 0.0023$ & $0.6035 \pm 0.0021$ \\
LPM-CPB (all) & Fixed & -- & $0.2539 \pm 0.0016$ & $0.2719 \pm 0.0016$ & $0.8435 \pm 0.0040$ & $0.6103 \pm 0.0035$ \\
LPM-L1000 & Fixed & dataset+dose+time & $0.2259 \pm 0.0023$ & $0.2442 \pm 0.0022$ & $0.8317 \pm 0.0024$ & $0.5925 \pm 0.0023$ \\
LPM-L1000 & Fixed & dataset & $0.2479 \pm 0.0018$ & $0.2661 \pm 0.0017$ & $0.8302 \pm 0.0016$ & $0.5923 \pm 0.0015$ \\
LPM-L1000 & Fixed & dose+time & $0.2431 \pm 0.0024$ & $0.2627 \pm 0.0024$ & $0.8219 \pm 0.0019$ & $0.5826 \pm 0.0018$ \\
Descriptor-free & -- & -- & $0.2533 \pm 0.0018$ & $0.2688 \pm 0.0018$ & $0.8161 \pm 0.0004$ & $0.5723 \pm 0.0005$ \\
LPM-L1000 & Yes & -- & $0.2583 \pm 0.0015$ & $0.2748 \pm 0.0014$ & $0.8115 \pm 0.0010$ & $0.5724 \pm 0.0009$ \\
LPM-CPB (nar.) & Yes & -- & $0.2548 \pm 0.0015$ & $0.2710 \pm 0.0014$ & $0.8089 \pm 0.0006$ & $0.5707 \pm 0.0006$ \\
LPM-L1000 & Yes & dose+time & $0.2670 \pm 0.0011$ & $0.2833 \pm 0.0011$ & $0.8049 \pm 0.0005$ & $0.5682 \pm 0.0006$ \\
LPM-CPB (all) & Yes & dataset & $0.2731 \pm 0.0014$ & $0.2890 \pm 0.0014$ & $0.8041 \pm 0.0005$ & $0.5682 \pm 0.0008$ \\
LPM-L1000 & Yes & dataset+dose+time & $0.2662 \pm 0.0013$ & $0.2825 \pm 0.0011$ & $0.8041 \pm 0.0005$ & $0.5665 \pm 0.0005$ \\
LPM-CPB (all) & Yes & -- & $0.2745 \pm 0.0011$ & $0.2895 \pm 0.0013$ & $0.8034 \pm 0.0006$ & $0.5676 \pm 0.0009$ \\
LPM-CPB (all) & Yes & dataset+dose+time & $\mathbf{0.2810 \pm 0.0012}$ & $\mathbf{0.2989 \pm 0.0013}$ & $0.8011 \pm 0.0005$ & $\mathbf{0.5649 \pm 0.0005}$ \\
LPM-L1000 & Yes & dataset & $0.2657 \pm 0.0015$ & $0.2835 \pm 0.0014$ & $\mathbf{0.8005 \pm 0.0007}$ & $\mathbf{0.5648 \pm 0.0006}$ \\
LPM-CPB (all) & Yes & dose+time & $\mathbf{0.2790 \pm 0.0009}$ & $0.2961 \pm 0.0008$ & $\mathbf{0.7995 \pm 0.0006}$ & $\mathbf{0.5639 \pm 0.0007}$ \\
\midrule
\multicolumn{7}{l}{\textit{coverage-matched}} \\
\addlinespace[2pt]
LPM-CPB (all) & Fixed & dose+time & $0.2327 \pm 0.0022$ & $0.2523 \pm 0.0023$ & $0.8556 \pm 0.0018$ & $0.6166 \pm 0.0016$ \\
LPM-CPB (all) & Fixed & dataset & $0.2216 \pm 0.0025$ & $0.2404 \pm 0.0024$ & $0.8512 \pm 0.0032$ & $0.6127 \pm 0.0028$ \\
LPM-CPB (all) & Fixed & dataset+dose+time & $0.2216 \pm 0.0029$ & $0.2389 \pm 0.0029$ & $0.8475 \pm 0.0026$ & $0.6082 \pm 0.0023$ \\
LPM-CPB (all) & Fixed & -- & $0.2484 \pm 0.0010$ & $0.2666 \pm 0.0011$ & $0.8473 \pm 0.0032$ & $0.6131 \pm 0.0029$ \\
LPM-L1000 & Fixed & -- & $0.2372 \pm 0.0014$ & $0.2545 \pm 0.0013$ & $0.8452 \pm 0.0020$ & $0.6019 \pm 0.0019$ \\
LPM-CPB (nar.) & Fixed & -- & $0.2221 \pm 0.0023$ & $0.2391 \pm 0.0023$ & $0.8439 \pm 0.0023$ & $0.6035 \pm 0.0021$ \\
LPM-L1000 & Fixed & dataset+dose+time & $0.2259 \pm 0.0023$ & $0.2442 \pm 0.0022$ & $0.8317 \pm 0.0024$ & $0.5925 \pm 0.0023$ \\
LPM-L1000 & Fixed & dataset & $0.2479 \pm 0.0018$ & $0.2661 \pm 0.0017$ & $0.8302 \pm 0.0016$ & $0.5923 \pm 0.0015$ \\
LPM-L1000 & Fixed & dose+time & $0.2431 \pm 0.0024$ & $0.2627 \pm 0.0024$ & $0.8219 \pm 0.0019$ & $0.5826 \pm 0.0018$ \\
Descriptor-free & -- & -- & $0.2533 \pm 0.0018$ & $0.2688 \pm 0.0018$ & $0.8161 \pm 0.0004$ & $0.5723 \pm 0.0005$ \\
LPM-L1000 & Yes & -- & $0.2583 \pm 0.0015$ & $0.2748 \pm 0.0014$ & $0.8115 \pm 0.0010$ & $0.5724 \pm 0.0009$ \\
LPM-CPB (nar.) & Yes & -- & $0.2548 \pm 0.0015$ & $0.2710 \pm 0.0014$ & $0.8089 \pm 0.0006$ & $0.5707 \pm 0.0006$ \\
LPM-CPB (all) & Yes & dataset & $0.2656 \pm 0.0010$ & $0.2821 \pm 0.0012$ & $0.8065 \pm 0.0004$ & $0.5702 \pm 0.0008$ \\
LPM-CPB (all) & Yes & -- & $0.2670 \pm 0.0008$ & $0.2825 \pm 0.0009$ & $0.8059 \pm 0.0005$ & $0.5696 \pm 0.0008$ \\
LPM-L1000 & Yes & dose+time & $0.2670 \pm 0.0011$ & $0.2833 \pm 0.0011$ & $0.8049 \pm 0.0005$ & $0.5682 \pm 0.0006$ \\
LPM-L1000 & Yes & dataset+dose+time & $0.2662 \pm 0.0013$ & $0.2825 \pm 0.0011$ & $0.8041 \pm 0.0005$ & $0.5665 \pm 0.0005$ \\
LPM-CPB (all) & Yes & dataset+dose+time & $\mathbf{0.2728 \pm 0.0008}$ & $\mathbf{0.2911 \pm 0.0010}$ & $0.8039 \pm 0.0004$ & $0.5673 \pm 0.0006$ \\
LPM-L1000 & Yes & dataset & $0.2657 \pm 0.0015$ & $0.2835 \pm 0.0014$ & $\mathbf{0.8005 \pm 0.0007}$ & $\mathbf{0.5648 \pm 0.0006}$ \\
LPM-CPB (all) & Yes & dose+time & $\mathbf{0.2738 \pm 0.0010}$ & $\mathbf{0.2914 \pm 0.0011}$ & $\mathbf{0.8004 \pm 0.0005}$ & $\mathbf{0.5648 \pm 0.0006}$ \\
\bottomrule
\end{tabular}%
}
\end{table*}

\begin{table*}[t]
\caption{\textbf{Bootstrap intervals for held-out OP3 compound predictions.}
Values correspond to the unmatched-coverage evaluation in Table~\ref{tab:op3_unseen_compounds}.
Each cell gives the paired $\Delta$ against the descriptor-free baseline (first line) and the
win rate, the nested mean of per-unit wins for the method (second line, \textcolor{gray}{gray}). Positive $\Delta$ and
win rates above $0.5$ favor the method. Row scores are seed-averaged before bootstrapping.
Intervals are 95\% BCa bootstrap confidence intervals computed from 2{,}000 iterations,
using the 35 held-out test compounds as resampling clusters and cell types as inner strata. $\dagger$ marks $\Delta$ intervals excluding zero;
\textbf{bold} marks those that do so in favor of the method.
The Trainable column indicates whether the externally initialized compound embedding is
updated during training on the OP3 benchmark; the Fixed setting freezes it. Rows are ordered by $\Delta$ MRRMSE, and the order matches Table~\ref{tab:op3_unseen_compounds}. Step-matched rows correspond to LPM embeddings from Chem-PerturBridge (All) pretrained for only as many steps as the original L1000 run, and from L1000 pretrained for as many steps as the original All run. Each method is compared with the descriptor-free
baseline. Abbreviations: CB1/CB2, ChemBERTa-1/2; MF-XL, MolFormer-XL; LPM, large perturbation model;
CPB, Chem-PerturBridge; nar., narrowed.}
\label{tab:op3_bootstrap}
\centering
\tiny
\setlength{\tabcolsep}{1pt}
\renewcommand{\arraystretch}{0.95}
\begin{tabularx}{\textwidth}{@{}>{\raggedright\arraybackslash}p{2.6cm}>{\centering\arraybackslash}p{0.7cm}*{4}{>{\raggedright\arraybackslash}X}@{}}
\toprule
Method & Trainable & $\Delta$ Spearman \newline \textcolor{gray}{win rate} & $\Delta$ Cosine \newline \textcolor{gray}{win rate} & $\Delta$ MRRMSE \newline \textcolor{gray}{win rate} & $\Delta$ MAE 
\newline \textcolor{gray}{win rate} \\
\midrule
CB1 & Fixed & -0.0724 [-0.1317, -0.0156]$^{\dagger}$ \newline \textcolor{gray}{0.297 [0.193, 0.434]} & -0.0743 [-0.1329, -0.0152]$^{\dagger}$ \newline \textcolor{gray}{0.311 [0.207, 0.464]} & -0.0581 [-0.1135, -0.0273]$^{\dagger}$ \newline \textcolor{gray}{0.282 [0.173, 0.414]} & -0.0535 [-0.0939, -0.0288]$^{\dagger}$ \newline \textcolor{gray}{0.232 [0.139, 0.353]} \\
\addlinespace[1pt]
LPM-CPB (all) step-matched & Fixed & -0.0507 [-0.0961, -0.0100]$^{\dagger}$ \newline \textcolor{gray}{0.391 [0.282, 0.506]} & -0.0490 [-0.0961, -0.0088]$^{\dagger}$ \newline \textcolor{gray}{0.377 [0.265, 0.500]} & -0.0506 [-0.0993, -0.0226]$^{\dagger}$ \newline \textcolor{gray}{0.319 [0.217, 0.436]} & -0.0493 [-0.0881, -0.0270]$^{\dagger}$ \newline \textcolor{gray}{0.319 [0.207, 0.443]} \\
\addlinespace[1pt]
MF-XL & Fixed & -0.0442 [-0.0964, 0.0046] \newline \textcolor{gray}{0.362 [0.246, 0.500]} & -0.0396 [-0.0960, 0.0102] \newline \textcolor{gray}{0.348 [0.236, 0.457]} & -0.0466 [-0.0904, -0.0182]$^{\dagger}$ \newline \textcolor{gray}{0.363 [0.254, 0.486]} & -0.0493 [-0.0861, -0.0243]$^{\dagger}$ \newline \textcolor{gray}{0.341 [0.227, 0.457]} \\
\addlinespace[1pt]
CB2 & Fixed & -0.0344 [-0.0817, 0.0179] \newline \textcolor{gray}{0.378 [0.264, 0.507]} & -0.0312 [-0.0849, 0.0206] \newline \textcolor{gray}{0.384 [0.284, 0.507]} & -0.0392 [-0.0715, -0.0140]$^{\dagger}$ \newline \textcolor{gray}{0.355 [0.236, 0.486]} & -0.0409 [-0.0705, -0.0190]$^{\dagger}$ \newline \textcolor{gray}{0.312 [0.203, 0.436]} \\
\addlinespace[1pt]
Morgan & Fixed & -0.0280 [-0.0798, 0.0092] \newline \textcolor{gray}{0.492 [0.358, 0.636]} & -0.0271 [-0.0802, 0.0161] \newline \textcolor{gray}{0.477 [0.354, 0.601]} & -0.0388 [-0.0811, -0.0054]$^{\dagger}$ \newline \textcolor{gray}{0.362 [0.240, 0.500]} & -0.0418 [-0.0756, -0.0195]$^{\dagger}$ \newline \textcolor{gray}{0.282 [0.188, 0.413]} \\
\addlinespace[1pt]
CB2 & Yes & -0.0197 [-0.0562, 0.0189] \newline \textcolor{gray}{0.449 [0.333, 0.562]} & -0.0174 [-0.0574, 0.0250] \newline \textcolor{gray}{0.442 [0.322, 0.557]} & -0.0300 [-0.0612, -0.0055]$^{\dagger}$ \newline \textcolor{gray}{0.370 [0.250, 0.494]} & -0.0330 [-0.0591, -0.0152]$^{\dagger}$ \newline \textcolor{gray}{0.341 [0.210, 0.472]} \\
\addlinespace[1pt]
LPM-L1000 & Fixed & -0.0163 [-0.0533, 0.0170] \newline \textcolor{gray}{0.449 [0.336, 0.574]} & -0.0145 [-0.0507, 0.0183] \newline \textcolor{gray}{0.463 [0.348, 0.579]} & -0.0294 [-0.0707, -0.0053]$^{\dagger}$ \newline \textcolor{gray}{0.463 [0.333, 0.585]} & -0.0299 [-0.0667, -0.0103]$^{\dagger}$ \newline \textcolor{gray}{0.406 [0.292, 0.537]} \\
\addlinespace[1pt]
LPM-CPB (nar.) & Fixed & -0.0314 [-0.0596, -0.0011]$^{\dagger}$ \newline \textcolor{gray}{0.441 [0.321, 0.560]} & -0.0299 [-0.0604, 0.0008] \newline \textcolor{gray}{0.434 [0.316, 0.563]} & -0.0280 [-0.0583, -0.0085]$^{\dagger}$ \newline \textcolor{gray}{0.384 [0.266, 0.521]} & -0.0315 [-0.0582, -0.0154]$^{\dagger}$ \newline \textcolor{gray}{0.333 [0.214, 0.473]} \\
\addlinespace[1pt]
LPM-CPB (all) & Fixed & +0.0007 [-0.0305, 0.0358] \newline \textcolor{gray}{0.500 [0.371, 0.629]} & +0.0032 [-0.0299, 0.0384] \newline \textcolor{gray}{0.521 [0.400, 0.643]} & -0.0275 [-0.0590, 0.0013] \newline \textcolor{gray}{0.399 [0.278, 0.529]} & -0.0381 [-0.0659, -0.0171]$^{\dagger}$ \newline \textcolor{gray}{0.319 [0.203, 0.449]} \\
\addlinespace[1pt]
MF-XL & Yes & -0.0187 [-0.0515, 0.0192] \newline \textcolor{gray}{0.471 [0.353, 0.587]} & -0.0150 [-0.0520, 0.0222] \newline \textcolor{gray}{0.492 [0.369, 0.622]} & -0.0241 [-0.0534, -0.0022]$^{\dagger}$ \newline \textcolor{gray}{0.442 [0.321, 0.573]} & -0.0257 [-0.0502, -0.0085]$^{\dagger}$ \newline \textcolor{gray}{0.427 [0.300, 0.552]} \\
\addlinespace[1pt]
LPM-L1000 step-matched & Fixed & -0.0199 [-0.0555, 0.0133] \newline \textcolor{gray}{0.449 [0.329, 0.564]} & -0.0187 [-0.0525, 0.0151] \newline \textcolor{gray}{0.478 [0.356, 0.610]} & -0.0211 [-0.0457, -0.0019]$^{\dagger}$ \newline \textcolor{gray}{0.420 [0.287, 0.565]} & -0.0231 [-0.0437, -0.0070]$^{\dagger}$ \newline \textcolor{gray}{0.406 [0.275, 0.564]} \\
\addlinespace[1pt]
CB1 & Yes & -0.0038 [-0.0214, 0.0141] \newline \textcolor{gray}{0.499 [0.391, 0.603]} & -0.0039 [-0.0243, 0.0161] \newline \textcolor{gray}{0.514 [0.400, 0.619]} & -0.0119 [-0.0366, 0.0008] \newline \textcolor{gray}{0.435 [0.318, 0.565]} & -0.0129 [-0.0333, -0.0025]$^{\dagger}$ \newline \textcolor{gray}{0.391 [0.270, 0.521]} \\
\addlinespace[1pt]
LPM-L1000 & Yes & +0.0049 [-0.0176, 0.0277] \newline \textcolor{gray}{0.543 [0.418, 0.667]} & +0.0059 [-0.0165, 0.0278] \newline \textcolor{gray}{0.564 [0.425, 0.688]} & +0.0045 [-0.0091, 0.0161] \newline \textcolor{gray}{0.593 [0.474, 0.711]} & -0.0002 [-0.0123, 0.0070] \newline \textcolor{gray}{0.600 [0.470, 0.716]} \\
\addlinespace[1pt]
Morgan & Yes & +0.0106 [-0.0002, 0.0313] \newline \textcolor{gray}{0.578 [0.459, 0.686]} & \textbf{+0.0119 [0.00003, 0.0317]}$^{\dagger}$ \newline \textcolor{gray}{0.586 [0.464, 0.690]} & +0.0046 [-0.0030, 0.0147] \newline \textcolor{gray}{0.579 [0.456, 0.704]} & +0.0007 [-0.0056, 0.0078] \newline \textcolor{gray}{0.514 [0.368, 0.648]} \\
\addlinespace[1pt]
LPM-L1000 step-matched & Yes & +0.0049 [-0.0139, 0.0242] \newline \textcolor{gray}{0.564 [0.449, 0.668]} & +0.0055 [-0.0138, 0.0250] \newline \textcolor{gray}{0.579 [0.471, 0.687]} & +0.0064 [-0.0026, 0.0188] \newline \textcolor{gray}{0.557 [0.443, 0.679]} & +0.0012 [-0.0050, 0.0092] \newline \textcolor{gray}{0.521 [0.373, 0.643]} \\
\addlinespace[1pt]
LPM-CPB (nar.) & Yes & +0.0013 [-0.0161, 0.0220] \newline \textcolor{gray}{0.542 [0.424, 0.652]} & +0.0021 [-0.0176, 0.0228] \newline \textcolor{gray}{0.543 [0.421, 0.671]} & +0.0071 [-0.0024, 0.0183] \newline \textcolor{gray}{0.572 [0.448, 0.687]} & +0.0015 [-0.0057, 0.0093] \newline \textcolor{gray}{0.536 [0.418, 0.647]} \\
\addlinespace[1pt]
LPM-CPB (all) step-matched & Yes & +0.0110 [-0.0093, 0.0381] \newline \textcolor{gray}{0.520 [0.386, 0.636]} & +0.0120 [-0.0092, 0.0412] \newline \textcolor{gray}{0.557 [0.419, 0.676]} & +0.0078 [-0.0025, 0.0196] \newline \textcolor{gray}{0.564 [0.436, 0.673]} & +0.0021 [-0.0052, 0.0093] \newline \textcolor{gray}{0.571 [0.456, 0.681]} \\
\addlinespace[1pt]
LPM-CPB (all) & Yes & \textbf{+0.0212 [0.0065, 0.0420]}$^{\dagger}$ \newline \textcolor{gray}{0.644 [0.521, 0.752]} & \textbf{+0.0207 [0.0043, 0.0416]}$^{\dagger}$ \newline \textcolor{gray}{0.616 [0.483, 0.728]} & \textbf{+0.0127 [0.0022, 0.0288]}$^{\dagger}$ \newline \textcolor{gray}{0.601 [0.464, 0.707]} & +0.0046 [-0.0021, 0.0134] \newline \textcolor{gray}{0.564 [0.429, 0.697]} \\
\bottomrule
\end{tabularx}
\end{table*}

\begin{table*}[t]
\caption{\textbf{Bootstrap intervals for held-out OP3 compound predictions. Coverage-matched results.}
Values correspond to the coverage-matched evaluation in Table~\ref{tab:op3_unseen_compounds_b}.
Each cell gives the paired $\Delta$ against the descriptor-free baseline (first line) and the
win rate, the nested mean of per-unit wins for the method (second line, \textcolor{gray}{gray}). Positive $\Delta$ and
win rates above $0.5$ favor the method. Row scores are seed-averaged before bootstrapping.
Intervals are 95\% BCa bootstrap confidence intervals computed from 2{,}000 iterations,
using the 35 held-out test compounds as resampling clusters and cell types as inner strata. $\dagger$ marks $\Delta$ intervals excluding zero;
\textbf{bold} marks those that do so in favor of the method.
The Trainable column indicates whether the externally initialized compound embedding is
updated during training on the OP3 benchmark; the Fixed setting freezes it. Rows are ordered by $\Delta$ MRRMSE, and the order matches Table~\ref{tab:op3_unseen_compounds_b}. Step-matched rows correspond to LPM embeddings from Chem-PerturBridge (All) pretrained for only as many steps as the original L1000 run, and from L1000 pretrained for as many steps as the original All run. Each method is compared with the descriptor-free
baseline.
Abbreviations: CB1/CB2, ChemBERTa-1/2; MF-XL, MolFormer-XL; LPM, large perturbation model;
CPB, Chem-PerturBridge; nar., narrowed.}
\label{tab:op3_bootstrap_b}
\centering
\tiny
\setlength{\tabcolsep}{1pt}
\renewcommand{\arraystretch}{0.95}
\begin{tabularx}{\textwidth}{@{}>{\raggedright\arraybackslash}p{2.6cm}>{\centering\arraybackslash}p{0.7cm}*{4}{>{\raggedright\arraybackslash}X}@{}}
\toprule
Method & Trainable & $\Delta$ Spearman \newline \textcolor{gray}{win rate} & $\Delta$ Cosine \newline \textcolor{gray}{win rate} & $\Delta$ MRRMSE \newline \textcolor{gray}{win rate} & $\Delta$ MAE 
\newline \textcolor{gray}{win rate} \\
\midrule
CB1 & Fixed & -0.0554 [-0.1132, -0.0044]$^{\dagger}$ \newline \textcolor{gray}{0.326 [0.207, 0.471]} & -0.0537 [-0.1131, 0.0030] \newline \textcolor{gray}{0.347 [0.239, 0.500]} & -0.0596 [-0.1146, -0.0275]$^{\dagger}$ \newline \textcolor{gray}{0.311 [0.203, 0.440]} & -0.0592 [-0.1025, -0.0347]$^{\dagger}$ \newline \textcolor{gray}{0.232 [0.150, 0.340]} \\
\addlinespace[1pt]
LPM-CPB (all) step-matched & Fixed & -0.0483 [-0.0909, -0.0109]$^{\dagger}$ \newline \textcolor{gray}{0.399 [0.279, 0.522]} & -0.0452 [-0.0884, -0.0062]$^{\dagger}$ \newline \textcolor{gray}{0.384 [0.271, 0.521]} & -0.0486 [-0.0964, -0.0232]$^{\dagger}$ \newline \textcolor{gray}{0.319 [0.207, 0.442]} & -0.0483 [-0.0857, -0.0267]$^{\dagger}$ \newline \textcolor{gray}{0.319 [0.210, 0.443]} \\
\addlinespace[1pt]
MF-XL & Fixed & -0.0431 [-0.0962, 0.0056] \newline \textcolor{gray}{0.341 [0.228, 0.486]} & -0.0384 [-0.0956, 0.0107] \newline \textcolor{gray}{0.370 [0.254, 0.486]} & -0.0476 [-0.0919, -0.0202]$^{\dagger}$ \newline \textcolor{gray}{0.370 [0.257, 0.493]} & -0.0495 [-0.0851, -0.0244]$^{\dagger}$ \newline \textcolor{gray}{0.348 [0.239, 0.464]} \\
\addlinespace[1pt]
CB2 & Fixed & -0.0488 [-0.0942, 0.0063] \newline \textcolor{gray}{0.334 [0.232, 0.457]} & -0.0453 [-0.0958, 0.0094] \newline \textcolor{gray}{0.370 [0.266, 0.491]} & -0.0453 [-0.0761, -0.0213]$^{\dagger}$ \newline \textcolor{gray}{0.297 [0.195, 0.414]} & -0.0461 [-0.0743, -0.0259]$^{\dagger}$ \newline \textcolor{gray}{0.239 [0.145, 0.356]} \\
\addlinespace[1pt]
Morgan & Fixed & -0.0477 [-0.1111, -0.0054]$^{\dagger}$ \newline \textcolor{gray}{0.456 [0.329, 0.602]} & -0.0483 [-0.1062, 0.0017] \newline \textcolor{gray}{0.449 [0.322, 0.584]} & -0.0451 [-0.0874, -0.0168]$^{\dagger}$ \newline \textcolor{gray}{0.348 [0.229, 0.486]} & -0.0447 [-0.0773, -0.0232]$^{\dagger}$ \newline \textcolor{gray}{0.283 [0.183, 0.407]} \\
\addlinespace[1pt]
CB2 & Yes & -0.0272 [-0.0636, 0.0127] \newline \textcolor{gray}{0.449 [0.332, 0.564]} & -0.0249 [-0.0644, 0.0186] \newline \textcolor{gray}{0.435 [0.318, 0.551]} & -0.0333 [-0.0651, -0.0117]$^{\dagger}$ \newline \textcolor{gray}{0.398 [0.275, 0.529]} & -0.0347 [-0.0601, -0.0179]$^{\dagger}$ \newline \textcolor{gray}{0.340 [0.214, 0.471]} \\
\addlinespace[1pt]
LPM-CPB (all) & Fixed & -0.0048 [-0.0372, 0.0276] \newline \textcolor{gray}{0.478 [0.350, 0.609]} & -0.0021 [-0.0352, 0.0330] \newline \textcolor{gray}{0.500 [0.379, 0.630]} & -0.0313 [-0.0606, -0.0025]$^{\dagger}$ \newline \textcolor{gray}{0.370 [0.257, 0.507]} & -0.0409 [-0.0678, -0.0207]$^{\dagger}$ \newline \textcolor{gray}{0.276 [0.171, 0.400]} \\
\addlinespace[1pt]
LPM-L1000 & Fixed & -0.0163 [-0.0533, 0.0170] \newline \textcolor{gray}{0.449 [0.336, 0.574]} & -0.0145 [-0.0507, 0.0183] \newline \textcolor{gray}{0.463 [0.348, 0.579]} & -0.0294 [-0.0707, -0.0053]$^{\dagger}$ \newline \textcolor{gray}{0.463 [0.333, 0.585]} & -0.0299 [-0.0667, -0.0103]$^{\dagger}$ \newline \textcolor{gray}{0.406 [0.292, 0.537]} \\
\addlinespace[1pt]
LPM-CPB (nar.) & Fixed & -0.0314 [-0.0596, -0.0011]$^{\dagger}$ \newline \textcolor{gray}{0.441 [0.321, 0.560]} & -0.0299 [-0.0604, 0.0008] \newline \textcolor{gray}{0.434 [0.316, 0.563]} & -0.0280 [-0.0583, -0.0085]$^{\dagger}$ \newline \textcolor{gray}{0.384 [0.266, 0.521]} & -0.0315 [-0.0582, -0.0154]$^{\dagger}$ \newline \textcolor{gray}{0.333 [0.214, 0.473]} \\
\addlinespace[1pt]
MF-XL & Yes & -0.0161 [-0.0476, 0.0221] \newline \textcolor{gray}{0.450 [0.336, 0.566]} & -0.0123 [-0.0488, 0.0247] \newline \textcolor{gray}{0.507 [0.393, 0.630]} & -0.0249 [-0.0547, -0.0043]$^{\dagger}$ \newline \textcolor{gray}{0.493 [0.371, 0.617]} & -0.0257 [-0.0522, -0.0082]$^{\dagger}$ \newline \textcolor{gray}{0.449 [0.326, 0.569]} \\
\addlinespace[1pt]
LPM-L1000 step-matched & Fixed & -0.0199 [-0.0555, 0.0133] \newline \textcolor{gray}{0.449 [0.329, 0.564]} & -0.0187 [-0.0525, 0.0151] \newline \textcolor{gray}{0.478 [0.356, 0.610]} & -0.0211 [-0.0457, -0.0019]$^{\dagger}$ \newline \textcolor{gray}{0.420 [0.287, 0.565]} & -0.0231 [-0.0437, -0.0070]$^{\dagger}$ \newline \textcolor{gray}{0.406 [0.275, 0.564]} \\
\addlinespace[1pt]
CB1 & Yes & +0.0003 [-0.0177, 0.0181] \newline \textcolor{gray}{0.550 [0.443, 0.651]} & +0.0007 [-0.0202, 0.0207] \newline \textcolor{gray}{0.558 [0.450, 0.664]} & -0.0116 [-0.0376, 0.0009] \newline \textcolor{gray}{0.464 [0.340, 0.594]} & -0.0129 [-0.0345, -0.0029]$^{\dagger}$ \newline \textcolor{gray}{0.413 [0.292, 0.543]} \\
\addlinespace[1pt]
Morgan & Yes & +0.0047 [-0.0074, 0.0261] \newline \textcolor{gray}{0.528 [0.405, 0.651]} & +0.0052 [-0.0082, 0.0255] \newline \textcolor{gray}{0.564 [0.442, 0.680]} & +0.0014 [-0.0061, 0.0081] \newline \textcolor{gray}{0.565 [0.443, 0.688]} & -0.0011 [-0.0073, 0.0040] \newline \textcolor{gray}{0.521 [0.371, 0.662]} \\
\addlinespace[1pt]
LPM-L1000 & Yes & +0.0049 [-0.0176, 0.0277] \newline \textcolor{gray}{0.543 [0.418, 0.667]} & +0.0059 [-0.0165, 0.0278] \newline \textcolor{gray}{0.564 [0.425, 0.688]} & +0.0045 [-0.0091, 0.0161] \newline \textcolor{gray}{0.593 [0.474, 0.711]} & -0.0002 [-0.0123, 0.0070] \newline \textcolor{gray}{0.600 [0.470, 0.716]} \\
\addlinespace[1pt]
LPM-L1000 step-matched & Yes & +0.0049 [-0.0139, 0.0242] \newline \textcolor{gray}{0.564 [0.449, 0.668]} & +0.0055 [-0.0138, 0.0250] \newline \textcolor{gray}{0.579 [0.471, 0.687]} & +0.0064 [-0.0026, 0.0188] \newline \textcolor{gray}{0.557 [0.443, 0.679]} & +0.0012 [-0.0050, 0.0092] \newline \textcolor{gray}{0.521 [0.373, 0.643]} \\
\addlinespace[1pt]
LPM-CPB (all) step-matched & Yes & +0.0068 [-0.0165, 0.0360] \newline \textcolor{gray}{0.542 [0.421, 0.660]} & +0.0085 [-0.0139, 0.0371] \newline \textcolor{gray}{0.550 [0.426, 0.673]} & +0.0068 [-0.0035, 0.0188] \newline \textcolor{gray}{0.571 [0.448, 0.680]} & +0.0010 [-0.0066, 0.0085] \newline \textcolor{gray}{0.513 [0.392, 0.616]} \\
\addlinespace[1pt]
LPM-CPB (nar.) & Yes & +0.0013 [-0.0161, 0.0220] \newline \textcolor{gray}{0.542 [0.424, 0.652]} & +0.0021 [-0.0176, 0.0228] \newline \textcolor{gray}{0.543 [0.421, 0.671]} & +0.0071 [-0.0024, 0.0183] \newline \textcolor{gray}{0.572 [0.448, 0.687]} & +0.0015 [-0.0057, 0.0093] \newline \textcolor{gray}{0.536 [0.418, 0.647]} \\
\addlinespace[1pt]
LPM-CPB (all) & Yes & +0.0137 [-0.0052, 0.0345] \newline \textcolor{gray}{0.586 [0.455, 0.700]} & +0.0138 [-0.0054, 0.0360] \newline \textcolor{gray}{0.550 [0.414, 0.679]} & +0.0102 [-0.0013, 0.0262] \newline \textcolor{gray}{0.550 [0.418, 0.674]} & +0.0026 [-0.0050, 0.0120] \newline \textcolor{gray}{0.492 [0.362, 0.631]} \\
\bottomrule
\end{tabularx}
\end{table*}

\begin{table*}[t]
\caption{\textbf{Bootstrap intervals for LPM covariate ablations on held-out OP3 compounds.}
Each cell gives the paired $\Delta$ against the descriptor-free baseline (first line) and the
win rate, the nested mean of per-unit wins for the method (second line, \textcolor{gray}{gray}). Positive $\Delta$ and
win rates above $0.5$ favor the method. Row scores are seed-averaged before bootstrapping.
Intervals are 95\% BCa bootstrap confidence intervals computed from 2{,}000 iterations,
using the 35 held-out test compounds as resampling clusters and cell types as inner strata. $\dagger$ marks $\Delta$ intervals excluding zero;
\textbf{bold} marks those that do so in favor of the method.
The Covariates removed column indicates which Chem-PerturBridge covariates (dataset, dose+time, or dataset+dose+time) were ablated during LPM compound-embedding pretraining.
The top block uses each method's native OP3 compound coverage; the bottom block is coverage-matched.
The Trainable column indicates whether the externally initialized compound embedding is
updated during training on the OP3 benchmark; the Fixed setting freezes it. Rows within each block are ordered by $\Delta$ MRRMSE.
Abbreviations: LPM, large perturbation model; CPB, Chem-PerturBridge; nar., narrowed.}
\label{tab:op3_ablation_bootstrap}
\centering
\tiny
\setlength{\tabcolsep}{1pt}
\renewcommand{\arraystretch}{0.90}
\resizebox{0.97\textwidth}{!}{%
\begin{tabularx}{1.12\textwidth}{@{}>{\raggedright\arraybackslash}p{1.55cm}>{\centering\arraybackslash}p{0.75cm}>{\raggedright\arraybackslash}p{1.85cm}*{4}{>{\raggedright\arraybackslash}X}@{}}
\toprule
Method & Trainable & Covariates removed & $\Delta$ Spearman \newline \textcolor{gray}{win rate} & $\Delta$ Cosine \newline \textcolor{gray}{win rate} & $\Delta$ MRRMSE \newline \textcolor{gray}{win rate} & $\Delta$ MAE \newline \textcolor{gray}{win rate} \\
\midrule
\multicolumn{7}{@{}l@{}}{\textit{native coverage}} \\
\addlinespace[2pt]
LPM-CPB (all) & Fixed & dose+time & -0.0277 [-0.0801, 0.0247] \newline \textcolor{gray}{0.464 [0.341, 0.609]} & -0.0241 [-0.0788, 0.0286] \newline \textcolor{gray}{0.464 [0.323, 0.613]} & -0.0461 [-0.0904, -0.0134]$^{\dagger}$ \newline \textcolor{gray}{0.327 [0.215, 0.464]} & -0.0500 [-0.0865, -0.0258]$^{\dagger}$ \newline \textcolor{gray}{0.283 [0.170, 0.420]} \\
\addlinespace[1pt]
LPM-CPB (all) & Fixed & dataset+dose+time & -0.0416 [-0.0958, 0.0176] \newline \textcolor{gray}{0.406 [0.291, 0.539]} & -0.0408 [-0.0983, 0.0200] \newline \textcolor{gray}{0.399 [0.279, 0.536]} & -0.0334 [-0.0736, 0.0020] \newline \textcolor{gray}{0.320 [0.208, 0.472]} & -0.0372 [-0.0730, -0.0129]$^{\dagger}$ \newline \textcolor{gray}{0.334 [0.217, 0.469]} \\
\addlinespace[1pt]
LPM-L1000 & Fixed & -- & -0.0163 [-0.0533, 0.0170] \newline \textcolor{gray}{0.449 [0.336, 0.574]} & -0.0145 [-0.0507, 0.0183] \newline \textcolor{gray}{0.463 [0.348, 0.579]} & -0.0294 [-0.0707, -0.0053]$^{\dagger}$ \newline \textcolor{gray}{0.463 [0.333, 0.585]} & -0.0299 [-0.0667, -0.0103]$^{\dagger}$ \newline \textcolor{gray}{0.406 [0.292, 0.537]} \\
\addlinespace[1pt]
LPM-CPB (all) & Fixed & dataset & -0.0245 [-0.0730, 0.0164] \newline \textcolor{gray}{0.435 [0.305, 0.565]} & -0.0216 [-0.0697, 0.0232] \newline \textcolor{gray}{0.449 [0.321, 0.580]} & -0.0284 [-0.0697, 0.0041] \newline \textcolor{gray}{0.399 [0.271, 0.537]} & -0.0368 [-0.0708, -0.0105]$^{\dagger}$ \newline \textcolor{gray}{0.391 [0.264, 0.514]} \\
\addlinespace[1pt]
LPM-CPB (nar.) & Fixed & -- & -0.0314 [-0.0596, -0.0011]$^{\dagger}$ \newline \textcolor{gray}{0.441 [0.321, 0.560]} & -0.0299 [-0.0604, 0.0008] \newline \textcolor{gray}{0.434 [0.316, 0.563]} & -0.0280 [-0.0583, -0.0085]$^{\dagger}$ \newline \textcolor{gray}{0.384 [0.266, 0.521]} & -0.0315 [-0.0582, -0.0154]$^{\dagger}$ \newline \textcolor{gray}{0.333 [0.214, 0.473]} \\
\addlinespace[1pt]
LPM-CPB (all) & Fixed & -- & +0.0007 [-0.0305, 0.0358] \newline \textcolor{gray}{0.500 [0.371, 0.629]} & +0.0032 [-0.0299, 0.0384] \newline \textcolor{gray}{0.521 [0.400, 0.643]} & -0.0275 [-0.0590, 0.0013] \newline \textcolor{gray}{0.399 [0.278, 0.529]} & -0.0381 [-0.0659, -0.0171]$^{\dagger}$ \newline \textcolor{gray}{0.319 [0.203, 0.449]} \\
\addlinespace[1pt]
LPM-L1000 & Fixed & dataset+dose+time & -0.0274 [-0.0669, 0.0075] \newline \textcolor{gray}{0.470 [0.364, 0.578]} & -0.0247 [-0.0741, 0.0107] \newline \textcolor{gray}{0.491 [0.387, 0.594]} & -0.0157 [-0.0348, 0.0025] \newline \textcolor{gray}{0.427 [0.304, 0.557]} & -0.0204 [-0.0371, -0.0078]$^{\dagger}$ \newline \textcolor{gray}{0.390 [0.271, 0.513]} \\
\addlinespace[1pt]
LPM-L1000 & Fixed & dataset & -0.0054 [-0.0381, 0.0281] \newline \textcolor{gray}{0.478 [0.357, 0.598]} & -0.0027 [-0.0353, 0.0313] \newline \textcolor{gray}{0.486 [0.361, 0.603]} & -0.0142 [-0.0553, 0.0161] \newline \textcolor{gray}{0.478 [0.345, 0.607]} & -0.0202 [-0.0501, -0.00002]$^{\dagger}$ \newline \textcolor{gray}{0.456 [0.332, 0.600]} \\
\addlinespace[1pt]
LPM-L1000 & Fixed & dose+time & -0.0100 [-0.0520, 0.0255] \newline \textcolor{gray}{0.486 [0.357, 0.600]} & -0.0059 [-0.0546, 0.0336] \newline \textcolor{gray}{0.493 [0.359, 0.621]} & -0.0058 [-0.0267, 0.0100] \newline \textcolor{gray}{0.500 [0.353, 0.621]} & -0.0104 [-0.0271, 0.0027] \newline \textcolor{gray}{0.428 [0.298, 0.571]} \\
\addlinespace[1pt]
LPM-L1000 & Yes & -- & +0.0049 [-0.0176, 0.0277] \newline \textcolor{gray}{0.543 [0.418, 0.667]} & +0.0059 [-0.0165, 0.0278] \newline \textcolor{gray}{0.564 [0.425, 0.688]} & +0.0045 [-0.0091, 0.0161] \newline \textcolor{gray}{0.593 [0.474, 0.711]} & -0.0002 [-0.0123, 0.0070] \newline \textcolor{gray}{0.600 [0.470, 0.716]} \\
\addlinespace[1pt]
LPM-CPB (nar.) & Yes & -- & +0.0013 [-0.0161, 0.0220] \newline \textcolor{gray}{0.542 [0.424, 0.652]} & +0.0021 [-0.0176, 0.0228] \newline \textcolor{gray}{0.543 [0.421, 0.671]} & +0.0071 [-0.0024, 0.0183] \newline \textcolor{gray}{0.572 [0.448, 0.687]} & +0.0015 [-0.0057, 0.0093] \newline \textcolor{gray}{0.536 [0.418, 0.647]} \\
\addlinespace[1pt]
LPM-L1000 & Yes & dose+time & +0.0138 [-0.0049, 0.0309] \newline \textcolor{gray}{0.594 [0.473, 0.703]} & +0.0146 [-0.0057, 0.0338] \newline \textcolor{gray}{0.615 [0.500, 0.732]} & \textbf{+0.0112 [0.00009, 0.0260]}$^{\dagger}$ \newline \textcolor{gray}{0.608 [0.485, 0.729]} & +0.0041 [-0.0037, 0.0120] \newline \textcolor{gray}{0.572 [0.421, 0.688]} \\
\addlinespace[1pt]
LPM-CPB (all) & Yes & dataset & \textbf{+0.0198 [0.0020, 0.0469]}$^{\dagger}$ \newline \textcolor{gray}{0.550 [0.419, 0.674]} & \textbf{+0.0201 [0.0010, 0.0466]}$^{\dagger}$ \newline \textcolor{gray}{0.550 [0.421, 0.671]} & \textbf{+0.0120 [0.0002, 0.0295]}$^{\dagger}$ \newline \textcolor{gray}{0.550 [0.420, 0.677]} & +0.0040 [-0.0045, 0.0140] \newline \textcolor{gray}{0.543 [0.394, 0.674]} \\
\addlinespace[1pt]
LPM-L1000 & Yes & dataset+dose+time & +0.0128 [-0.0079, 0.0338] \newline \textcolor{gray}{0.629 [0.500, 0.736]} & +0.0137 [-0.0111, 0.0353] \newline \textcolor{gray}{0.644 [0.528, 0.755]} & \textbf{+0.0120 [0.0017, 0.0252]}$^{\dagger}$ \newline \textcolor{gray}{0.622 [0.492, 0.751]} & +0.0058 [-0.0021, 0.0126] \newline \textcolor{gray}{0.622 [0.497, 0.742]} \\
\addlinespace[1pt]
LPM-CPB (all) & Yes & -- & \textbf{+0.0212 [0.0065, 0.0420]}$^{\dagger}$ \newline \textcolor{gray}{0.644 [0.521, 0.752]} & \textbf{+0.0207 [0.0043, 0.0416]}$^{\dagger}$ \newline \textcolor{gray}{0.616 [0.483, 0.728]} & \textbf{+0.0127 [0.0022, 0.0288]}$^{\dagger}$ \newline \textcolor{gray}{0.601 [0.464, 0.707]} & +0.0046 [-0.0021, 0.0134] \newline \textcolor{gray}{0.564 [0.429, 0.697]} \\
\addlinespace[1pt]
LPM-CPB (all) & Yes & dataset+dose+time & \textbf{+0.0278 [0.0097, 0.0505]}$^{\dagger}$ \newline \textcolor{gray}{0.608 [0.500, 0.707]} & \textbf{+0.0301 [0.0116, 0.0541]}$^{\dagger}$ \newline \textcolor{gray}{0.593 [0.498, 0.702]} & \textbf{+0.0150 [0.0030, 0.0319]}$^{\dagger}$ \newline \textcolor{gray}{0.609 [0.493, 0.725]} & +0.0073 [-0.0021, 0.0177] \newline \textcolor{gray}{0.587 [0.463, 0.716]} \\
\addlinespace[1pt]
LPM-L1000 & Yes & dataset & +0.0124 [-0.0095, 0.0447] \newline \textcolor{gray}{0.536 [0.427, 0.645]} & +0.0148 [-0.0080, 0.0444] \newline \textcolor{gray}{0.536 [0.427, 0.646]} & \textbf{+0.0157 [0.0042, 0.0330]}$^{\dagger}$ \newline \textcolor{gray}{0.608 [0.470, 0.729]} & +0.0074 [-0.0010, 0.0173] \newline \textcolor{gray}{0.593 [0.461, 0.714]} \\
\addlinespace[1pt]
LPM-CPB (all) & Yes & dose+time & \textbf{+0.0258 [0.0100, 0.0502]}$^{\dagger}$ \newline \textcolor{gray}{0.587 [0.471, 0.704]} & \textbf{+0.0274 [0.0102, 0.0567]}$^{\dagger}$ \newline \textcolor{gray}{0.594 [0.464, 0.710]} & \textbf{+0.0166 [0.0058, 0.0355]}$^{\dagger}$ \newline \textcolor{gray}{0.537 [0.399, 0.668]} & \textbf{+0.0083 [0.0012, 0.0192]}$^{\dagger}$ \newline \textcolor{gray}{0.543 [0.414, 0.676]} \\

\midrule
\multicolumn{7}{@{}l@{}}{\textit{coverage-matched}} \\
\addlinespace[2pt]

LPM-CPB (all) & Fixed & dose+time & -0.0206 [-0.0719, 0.0279] \newline \textcolor{gray}{0.464 [0.336, 0.614]} & -0.0165 [-0.0689, 0.0360] \newline \textcolor{gray}{0.486 [0.357, 0.624]} & -0.0397 [-0.0836, -0.0087]$^{\dagger}$ \newline \textcolor{gray}{0.356 [0.236, 0.499]} & -0.0445 [-0.0791, -0.0212]$^{\dagger}$ \newline \textcolor{gray}{0.297 [0.186, 0.435]} \\
\addlinespace[1pt]
LPM-CPB (all) & Fixed & dataset & -0.0317 [-0.0765, 0.0115] \newline \textcolor{gray}{0.413 [0.286, 0.546]} & -0.0284 [-0.0767, 0.0167] \newline \textcolor{gray}{0.435 [0.293, 0.573]} & -0.0352 [-0.0730, -0.0048]$^{\dagger}$ \newline \textcolor{gray}{0.384 [0.264, 0.521]} & -0.0406 [-0.0723, -0.0157]$^{\dagger}$ \newline \textcolor{gray}{0.348 [0.232, 0.479]} \\
\addlinespace[1pt]
LPM-CPB (all) & Fixed & dataset+dose+time & -0.0315 [-0.0853, 0.0259] \newline \textcolor{gray}{0.399 [0.279, 0.536]} & -0.0298 [-0.0882, 0.0256] \newline \textcolor{gray}{0.399 [0.274, 0.537]} & -0.0313 [-0.0744, -0.00001]$^{\dagger}$ \newline \textcolor{gray}{0.320 [0.200, 0.457]} & -0.0359 [-0.0766, -0.0129]$^{\dagger}$ \newline \textcolor{gray}{0.327 [0.203, 0.464]} \\
\addlinespace[1pt]
LPM-CPB (all) & Fixed & -- & -0.0048 [-0.0372, 0.0276] \newline \textcolor{gray}{0.478 [0.350, 0.609]} & -0.0021 [-0.0352, 0.0330] \newline \textcolor{gray}{0.500 [0.379, 0.630]} & -0.0313 [-0.0606, -0.0025]$^{\dagger}$ \newline \textcolor{gray}{0.370 [0.257, 0.507]} & -0.0409 [-0.0678, -0.0207]$^{\dagger}$ \newline \textcolor{gray}{0.276 [0.171, 0.400]} \\
\addlinespace[1pt]
LPM-L1000 & Fixed & -- & -0.0163 [-0.0533, 0.0170] \newline \textcolor{gray}{0.449 [0.336, 0.574]} & -0.0145 [-0.0507, 0.0183] \newline \textcolor{gray}{0.463 [0.348, 0.579]} & -0.0294 [-0.0707, -0.0053]$^{\dagger}$ \newline \textcolor{gray}{0.463 [0.333, 0.585]} & -0.0299 [-0.0667, -0.0103]$^{\dagger}$ \newline \textcolor{gray}{0.406 [0.292, 0.537]} \\
\addlinespace[1pt]
LPM-CPB (nar.) & Fixed & -- & -0.0314 [-0.0596, -0.0011]$^{\dagger}$ \newline \textcolor{gray}{0.441 [0.321, 0.560]} & -0.0299 [-0.0604, 0.0008] \newline \textcolor{gray}{0.434 [0.316, 0.563]} & -0.0280 [-0.0583, -0.0085]$^{\dagger}$ \newline \textcolor{gray}{0.384 [0.266, 0.521]} & -0.0315 [-0.0582, -0.0154]$^{\dagger}$ \newline \textcolor{gray}{0.333 [0.214, 0.473]} \\
\addlinespace[1pt]
LPM-L1000 & Fixed & dataset+dose+time & -0.0274 [-0.0669, 0.0075] \newline \textcolor{gray}{0.470 [0.364, 0.578]} & -0.0247 [-0.0741, 0.0107] \newline \textcolor{gray}{0.491 [0.387, 0.594]} & -0.0157 [-0.0348, 0.0025] \newline \textcolor{gray}{0.427 [0.304, 0.557]} & -0.0204 [-0.0371, -0.0078]$^{\dagger}$ \newline \textcolor{gray}{0.390 [0.271, 0.513]} \\
\addlinespace[1pt]
LPM-L1000 & Fixed & dataset & -0.0054 [-0.0381, 0.0281] \newline \textcolor{gray}{0.478 [0.357, 0.598]} & -0.0027 [-0.0353, 0.0313] \newline \textcolor{gray}{0.486 [0.361, 0.603]} & -0.0142 [-0.0553, 0.0161] \newline \textcolor{gray}{0.478 [0.345, 0.607]} & -0.0202 [-0.0501, -0.00002]$^{\dagger}$ \newline \textcolor{gray}{0.456 [0.332, 0.600]} \\
\addlinespace[1pt]
LPM-L1000 & Fixed & dose+time & -0.0100 [-0.0520, 0.0255] \newline \textcolor{gray}{0.486 [0.357, 0.600]} & -0.0059 [-0.0546, 0.0336] \newline \textcolor{gray}{0.493 [0.359, 0.621]} & -0.0058 [-0.0267, 0.0100] \newline \textcolor{gray}{0.500 [0.353, 0.621]} & -0.0104 [-0.0271, 0.0027] \newline \textcolor{gray}{0.428 [0.298, 0.571]} \\
\addlinespace[1pt]
LPM-L1000 & Yes & -- & +0.0049 [-0.0176, 0.0277] \newline \textcolor{gray}{0.543 [0.418, 0.667]} & +0.0059 [-0.0165, 0.0278] \newline \textcolor{gray}{0.564 [0.425, 0.688]} & +0.0045 [-0.0091, 0.0161] \newline \textcolor{gray}{0.593 [0.474, 0.711]} & -0.0002 [-0.0123, 0.0070] \newline \textcolor{gray}{0.600 [0.470, 0.716]} \\
\addlinespace[1pt]
LPM-CPB (nar.) & Yes & -- & +0.0013 [-0.0161, 0.0220] \newline \textcolor{gray}{0.542 [0.424, 0.652]} & +0.0021 [-0.0176, 0.0228] \newline \textcolor{gray}{0.543 [0.421, 0.671]} & +0.0071 [-0.0024, 0.0183] \newline \textcolor{gray}{0.572 [0.448, 0.687]} & +0.0015 [-0.0057, 0.0093] \newline \textcolor{gray}{0.536 [0.418, 0.647]} \\
\addlinespace[1pt]
LPM-CPB (all) & Yes & dataset & +0.0123 [-0.0095, 0.0395] \newline \textcolor{gray}{0.514 [0.379, 0.644]} & +0.0133 [-0.0097, 0.0408] \newline \textcolor{gray}{0.492 [0.350, 0.636]} & +0.0096 [-0.0027, 0.0246] \newline \textcolor{gray}{0.478 [0.344, 0.609]} & +0.0020 [-0.0062, 0.0123] \newline \textcolor{gray}{0.492 [0.355, 0.629]} \\
\addlinespace[1pt]
LPM-CPB (all) & Yes & -- & +0.0137 [-0.0052, 0.0345] \newline \textcolor{gray}{0.586 [0.455, 0.700]} & +0.0138 [-0.0054, 0.0360] \newline \textcolor{gray}{0.550 [0.414, 0.679]} & +0.0102 [-0.0013, 0.0262] \newline \textcolor{gray}{0.550 [0.418, 0.674]} & +0.0026 [-0.0050, 0.0120] \newline \textcolor{gray}{0.492 [0.362, 0.631]} \\
\addlinespace[1pt]
LPM-L1000 & Yes & dose+time & +0.0138 [-0.0049, 0.0309] \newline \textcolor{gray}{0.594 [0.473, 0.703]} & +0.0146 [-0.0057, 0.0338] \newline \textcolor{gray}{0.615 [0.500, 0.732]} & \textbf{+0.0112 [0.00009, 0.0260]}$^{\dagger}$ \newline \textcolor{gray}{0.608 [0.485, 0.729]} & +0.0041 [-0.0037, 0.0120] \newline \textcolor{gray}{0.572 [0.421, 0.688]} \\
\addlinespace[1pt]
LPM-L1000 & Yes & dataset+dose+time & +0.0128 [-0.0079, 0.0338] \newline \textcolor{gray}{0.629 [0.500, 0.736]} & +0.0137 [-0.0111, 0.0353] \newline \textcolor{gray}{0.644 [0.528, 0.755]} & \textbf{+0.0120 [0.0017, 0.0252]}$^{\dagger}$ \newline \textcolor{gray}{0.622 [0.492, 0.751]} & +0.0058 [-0.0021, 0.0126] \newline \textcolor{gray}{0.622 [0.497, 0.742]} \\
\addlinespace[1pt]
LPM-CPB (all) & Yes & dataset+dose+time & +0.0196 [-0.0025, 0.0416] \newline \textcolor{gray}{0.579 [0.464, 0.702]} & \textbf{+0.0223 [0.0016, 0.0496]}$^{\dagger}$ \newline \textcolor{gray}{0.571 [0.448, 0.686]} & +0.0122 [-0.0004, 0.0298] \newline \textcolor{gray}{0.594 [0.464, 0.707]} & +0.0050 [-0.0042, 0.0157] \newline \textcolor{gray}{0.543 [0.414, 0.671]} \\
\addlinespace[1pt]
LPM-L1000 & Yes & dataset & +0.0124 [-0.0095, 0.0447] \newline \textcolor{gray}{0.536 [0.427, 0.645]} & +0.0148 [-0.0080, 0.0444] \newline \textcolor{gray}{0.536 [0.427, 0.646]} & \textbf{+0.0157 [0.0042, 0.0330]}$^{\dagger}$ \newline \textcolor{gray}{0.608 [0.470, 0.729]} & +0.0074 [-0.0010, 0.0173] \newline \textcolor{gray}{0.593 [0.461, 0.714]} \\
\addlinespace[1pt]
LPM-CPB (all) & Yes & dose+time & \textbf{+0.0206 [0.0005, 0.0454]}$^{\dagger}$ \newline \textcolor{gray}{0.587 [0.443, 0.703]} & \textbf{+0.0227 [0.0030, 0.0504]}$^{\dagger}$ \newline \textcolor{gray}{0.580 [0.445, 0.703]} & \textbf{+0.0157 [0.0044, 0.0357]}$^{\dagger}$ \newline \textcolor{gray}{0.544 [0.400, 0.677]} & \textbf{+0.0074 [0.0001, 0.0184]}$^{\dagger}$ \newline \textcolor{gray}{0.529 [0.391, 0.657]} \\
\bottomrule
\end{tabularx}}%
\end{table*}

\subsection{Molecule-holdout multi-output LPM evaluation}
\label{app:expanded_molecule_holdout_lpm}
We performed an expanded molecule-holdout evaluation to test whether the
harmonized resource improves direct perturbation-profile prediction across
source datasets. This experiment complements the OP3 downstream embedding
evaluation in Section~\ref{sec:repr}. Instead of exporting compound embeddings
to a separate OP3 predictor, we evaluate the multi-output LPM itself under
per-dataset molecule-holdout splits across the full Chem-PerturBridge resource.

The experiments start from the same harmonized AnnData and condition-level
effect assets used by the representation-learning pipeline. These assets are
converted into a multi-output LPM training table in which each row corresponds
to one perturbation sample rather than one sample-gene pair.

Splits are defined at the molecule level within each dataset. Validation and
test molecules are selected from molecules shared with at least one other
dataset, so that a molecule held out in one target dataset
can still remain available for training in another dataset. The target split
size is up to 10\% of shared molecules for validation and up to 10\% for test.
For datasets with very few shared molecules, we
divide the available shared held-out samples between validation and test evenly. This design measures transfer of compound-level information across datasets rather than zero-shot generalization to unseen molecules.

\begin{table*}[t]
\caption{\textbf{Molecule-holdout split summary for the multi-output LPM evaluation.}
Validation and test molecules are selected within each dataset from molecules
shared with at least one other dataset. VCPI-0002 has no test
molecules under this split because only one shared molecule was available; it is
therefore reported only in the test summaries.}
\label{tab:expanded_lpm_split_summary}

\centering

\scriptsize

\setlength{\tabcolsep}{3pt}

\resizebox{\textwidth}{!}{%

\begin{tabular}{l l r r r r r r r}

\toprule

Dataset

& Strategy

& Total molecules

& Shared molecules

& Val molecules

& Test molecules

& Train samples

& Val samples

& Test samples \\

\midrule

CIGS MCE

& ten\_percent\_shared\_molecules

& 11{,}160

& 3{,}402

& 340

& 340

& 20{,}493

& 660

& 666 \\

CIGS TCM

& ten\_percent\_shared\_molecules

& 1{,}820

& 990

& 99

& 99

& 6{,}353

& 385

& 379 \\

Ginkgo GDPx2

& ten\_percent\_shared\_molecules

& 86

& 72

& 7

& 7

& 1{,}728

& 168

& 168 \\

VCPI-0001

& low\_shared\_sample\_balanced

& 2{,}272

& 5

& 3

& 2

& 13{,}601

& 18

& 12 \\

VCPI-0002

& low\_shared\_sample\_balanced

& 1{,}488

& 1

& 1

& 0

& 8{,}921

& 6

& 0 \\

LINCS Phase I

& ten\_percent\_shared\_molecules

& 20{,}162

& 1{,}942

& 194

& 194

& 174{,}492

& 6{,}096

& 5{,}589 \\

LINCS Phase II

& ten\_percent\_shared\_molecules

& 1{,}791

& 1{,}425

& 142

& 142

& 82{,}472

& 8{,}440

& 8{,}017 \\

Novartis DRUG-seq

& ten\_percent\_shared\_molecules

& 3{,}819

& 1{,}456

& 145

& 145

& 14{,}116

& 580

& 580 \\

DILImap train

& ten\_percent\_shared\_molecules

& 250

& 221

& 22

& 22

& 820

& 86

& 87 \\

OP3

& ten\_percent\_shared\_molecules

& 105

& 104

& 10

& 10

& 335

& 40

& 40 \\

sci-Plex

& ten\_percent\_shared\_molecules

& 187

& 177

& 17

& 17

& 1{,}967

& 226

& 206 \\

Tahoe-100M

& ten\_percent\_shared\_molecules

& 378

& 340

& 34

& 34

& 44{,}102

& 4{,}799

& 4{,}866 \\

\bottomrule

\end{tabular}%
}
\end{table*}

We modified the LPM-style predictor from a gene-conditional scalar regressor
into a sample-conditional multi-output regressor for efficiency, given a large number of genes for many of the datasets. This follows a convention common in other perturbation prediction models~\cite{lotfollahi_predicting_2023, piran_disentanglement_2024}. In the original format, each
row corresponded to one sample-gene pair: the model received embeddings for the
dataset, context, compound, and readout gene, together with dose and time, and
predicted a scalar log-fold change for that gene. A full expression profile
therefore required one forward pass per output gene. In the multi-output
version, each row corresponds to one perturbation sample. The model receives
dataset, context, compound, dose, and time inputs, but not the output gene
identity, and returns a vector of predicted log-fold changes over the global
gene vocabulary in a single forward pass. Since datasets differ in measured
genes, every target vector is accompanied by an observed-gene mask; training
loss and validation/test RMSE are computed only over observed entries.

\begin{table*}[t]
\caption{\textbf{Model variants in the expanded molecule-holdout LPM evaluation.}
All variants use the same multi-output architecture and differ only in training
scope, initialization, fine-tuning, and molecule representation.}
\label{tab:expanded_lpm_variants}
\centering
\footnotesize
\begin{tabular}{p{3.0cm} p{10.5cm}}
\toprule
Variant & Description \\
\midrule
All & Train one multi-output LPM jointly on all datasets. \\
FT frozen mol & Initialize from the best all-data checkpoint for the target dataset, then fine-tune on the target dataset with molecule embeddings frozen. \\
Scratch & Train only on the target dataset from random initialization. \\
All Morgan fixed & Train an all-data model using fixed Morgan fingerprints as molecule embeddings. \\
FT Morgan fixed & Initialize target fine-tuning from the all-data fixed-Morgan model. \\
Scratch Morgan fixed & Train target-only with fixed Morgan fingerprints. \\
All Morgan learned & Initialize molecule embeddings from Morgan fingerprints, then allow them to update during all-data training. \\
FT Morgan learned fixmol & Initialize from the Morgan-initialized all-data checkpoint and fine-tune on the target dataset with the updated molecule embeddings frozen. \\
\bottomrule
\end{tabular}
\end{table*}

For every run, test RMSE is reported from the checkpoint with the best validation
RMSE. For all-data source models, we train 10
source seeds and, for each dataset and source seed, select the epoch with the
best dataset-specific validation RMSE. For fine-tuning models, we first choose
the source checkpoint across source seeds and epochs with the best validation
RMSE for the target dataset, then launch 10 target fine-tuning seeds from that
checkpoint. For target-only models, we train 10 seeds from random
initialization and select the best-validation checkpoint for each seed.
VCPI-0002 has no test split under the current molecule-holdout split and is
therefore excluded from test summary. The 10 seeds quantify optimization variance on a single fixed split per dataset.

Across the expanded molecule-holdout benchmark, Chem-PerturBridge-trained
models are highly competitive, but the strongest strategy depends on the target
dataset. The most consistently strong approach is to initialize molecule
embeddings from Morgan fingerprints, train the multi-output LPM across
Chem-PerturBridge, and then fine-tune on the target dataset while freezing the
learned molecule embeddings. This variant is best or near-best on 8 of the 11
datasets with test splits. More broadly, a method that uses all
Chem-PerturBridge datasets at some stage gives the lowest or tied-lowest mean
test RMSE in 10 of the 11 datasets under the reported means.
These results support the use of Chem-PerturBridge as a training corpus for
transferable perturbation models. At the same time, the benchmark exposes
important failure modes: target-only or fixed-Morgan baselines remain strongest
for some datasets, and GDPx2 shows little evidence that learned transferable
embeddings improve over random molecule embeddings (target-only training).

To test whether the differences between the consistently strong variant and the baseline exceed sampling variation, we compare FT Morgan learned fixmol against Scratch under two sources of uncertainty, seed-level tests per dataset and resampling over held-out molecules. Firstly, we use seed-level Welch tests on the 10 per-seed RMSEs, Holm-corrected across the 11 datasets. These are unpaired, since a seed does not link a scratch run to one started from a pretrained checkpoint. They result in 5 significant gains and 3 losses (Table~\ref{tab:lpm_molecule_welch_holm}), on CIGS MCE, Ginkgo GDPx2 and sci-Plex. Secondly, we resample held-out molecules. Per dataset we compute unit RMSE at $(\mathrm{compound},\,\mathrm{context},\,\mathrm{dose},\,\mathrm{time})$, average over seeds, and take $\Delta = \mathrm{RMSE(Scratch)}-\mathrm{RMSE(FT)}$, so positive values favor fine-tuning. Intervals are computed with the same perturbagen-clustered bootstrap strategy as the agreement benchmark (Appendix~\ref{app:cross_dataset_bootstrap_uncertainty}): 95\% BCa on a nested mean, cluster = $\mathrm{compound}$, strata = $\mathrm{context}\times\mathrm{time}$, 2{,}000 draws. Under this analysis, 4 datasets improve, 1 shows a loss, and 6 are indistinguishable (Table~\ref{tab:lpm_molecule_bootstrap}), counting VCPI-0001, where the interval is nominally nonzero but rests on two held-out molecules. Only CIGS MCE remains a loss, so two of the three seed-level losses reflect the seed-level precision on small deltas rather than a disadvantage that holds across molecules. Taking the two analyses together, fine-tuning helps beyond both seed and molecule variance on LINCS Phase I, LINCS Phase II, Novartis DRUG-seq, and OP3, and hurts under both on CIGS MCE. These tests cover one pairwise contrast; the remaining variants in Table~\ref{tab:expanded_lpm_test_rmse} are compared by mean and standard deviation only.

\begin{table*}[t]
\caption{\textbf{Per-dataset Welch $t$-tests on 10-seed test RMSE, Holm-corrected across datasets.}
Comparison is Scratch vs.\ FT (Morgan, learned fixmol). $\Delta$ is $\mathrm{mean\,RMSE}(\mathrm{Scratch})-\mathrm{mean\,RMSE}(\mathrm{FT})$, so positive values favor fine-tuning; $\Delta$ is computed before rounding.
Tests are unpaired two-sided Welch $t$-tests;
$p$-values are Holm-adjusted across the 11 datasets ($\alpha=0.05$); n.s. - not significant.}
\label{tab:lpm_molecule_welch_holm}
\centering
\footnotesize
\setlength{\tabcolsep}{3pt}
\begin{tabular}{lccccc}
\toprule
Dataset & Scratch RMSE & FT RMSE & $\Delta$ & $p_{\mathrm{Holm}}$ & Outcome \\
\midrule
LINCS Phase I & 0.6326 & 0.6214 & 0.0112 & 2.55$\times$10$^{-12}$ & FT better \\
LINCS Phase II & 0.4886 & 0.4814 & 0.0072 & 1.54$\times$10$^{-17}$ & FT better \\
Novartis DRUG-seq & 0.5408 & 0.5371 & 0.0037 & 6.61$\times$10$^{-10}$ & FT better \\
OP3 & 0.3478 & 0.3431 & 0.0048 & 7.20$\times$10$^{-7}$ & FT better \\
Tahoe-100M & 0.7380 & 0.7354 & 0.0027 & 5.97$\times$10$^{-4}$ & FT better \\
\midrule
CIGS TCM & 0.4081 & 0.4077 & 0.0004 & 0.172 & n.s. \\
DILImap train & 0.6095 & 0.6095 & 0.0001 & 0.891 & n.s. \\
VCPI-0001 & 0.5707 & 0.5706 & 0.0001 & 0.095 & n.s. \\
\midrule
sci-Plex & 0.6845 & 0.6899 & -0.0054 & 9.57$\times$10$^{-4}$ & Scratch better \\
Ginkgo GDPx2 & 0.4331 & 0.4380 & -0.0049 & 0.0034 & Scratch better \\
CIGS MCE & 0.5714 & 0.5750 & -0.0036 & 0.0017 & Scratch better \\
\bottomrule
\end{tabular}
\end{table*}

\begin{table*}[t]
\caption{\textbf{Bootstrap intervals for the Scratch vs.\ FT Morgan learned fixmol test RMSE comparison.}
Values correspond to the molecule-level analysis in Appendix~\ref{app:expanded_molecule_holdout_lpm}. Per dataset, unit RMSE is seed-averaged at $(\mathrm{compound},\,\mathrm{context},\,\mathrm{dose},\,\mathrm{time})$; $\Delta$ is the nested mean of $\mathrm{RMSE}(\mathrm{Scratch})-\mathrm{RMSE}(\mathrm{FT})$, so positive values favor fine-tuning. Win rate is the nested mean of unit wins for fine-tuning under the same clustering. Intervals are 95\% BCa bootstrap confidence intervals computed from 2{,}000 iterations, clustered by PubChem CID with inner strata over context and time. $n_{\mathrm{cmpd}}$ is the number of PubChem CID clusters with held-out test molecules. VCPI-0001 has two held-out molecules, so its interval is percentile rather than BCa, and we treat it as uninformative. Intervals are unadjusted across the 11 datasets; n.s. - not significant.}
\label{tab:lpm_molecule_bootstrap}
\centering
\footnotesize
\setlength{\tabcolsep}{3pt}
\begin{tabular}{lrccc}
\toprule
Dataset & $n_{\mathrm{cmpd}}$ &
$\Delta$ [95\% BCa] & Win rate [95\% BCa] & Outcome \\
\midrule
LINCS Phase I & 194 & 0.0090 [0.0066, 0.0112] & 0.623 [0.582, 0.656] & FT better \\
LINCS Phase II & 142 & 0.0084 [0.0046, 0.0139] & 0.568 [0.512, 0.623] & FT better \\
Novartis DRUG-seq & 145 & 0.0028 [0.0021, 0.0038] & 0.821 [0.781, 0.852] & FT better \\
OP3 & 10 & 0.0041 [0.0007, 0.0088] & 0.725 [0.475, 0.875] & FT better \\
\midrule
Tahoe-100M & 34 & 0.0025 [-0.0053, 0.0065] & 0.598 [0.520, 0.664] & n.s. \\
CIGS TCM & 99 & 0.0002 [-0.0002, 0.0005] & 0.474 [0.415, 0.531] & n.s. \\
DILImap train & 22 & 0.0001 [-0.0006, 0.0007] & 0.517 [0.368, 0.640] & n.s. \\
sci-Plex & 17 & -0.0010 [-0.0088, 0.0021] & 0.469 [0.240, 0.529] & n.s. \\
Ginkgo GDPx2 & 7 & -0.0016 [-0.0244, 0.0064] & 0.679 [0.399, 0.827] & n.s. \\
VCPI-0001 & 2 & 0.0001 [0.0000, 0.0001] & 0.583 [0.500, 0.667] & uninformative \\
\midrule
CIGS MCE & 340 & -0.0036 [-0.0108, -0.0013] & 0.393 [0.359, 0.428] & Scratch better \\
\bottomrule
\end{tabular}
\end{table*}

\section{Resource requirements}
\label{app:resource_requirements}

All experiments were run on a shared SLURM-managed HPC cluster.
Tables~\ref{tab:CPU_resources} and~\ref{tab:GPU_resources} report the
compute footprint of the GPU- and CPU-bound workflows used in this
work.

The CPU table covers the data-preparation and DEG-analysis pipeline.
For each (workflow, dataset) pair it reports the number of CPU nodes
used, the per-node CPU and RAM allocation, the peak RAM observed
during the job, the size of the input and
output files, the requested SLURM time limit, and the maximum
observed wall-clock time. CPU jobs ran on nodes with Intel Xeon
Gold, Intel Xeon Platinum, or AMD EPYC CPUs. Rows labeled
``DEG (group / sep)'' carry two values per cell, corresponding to
the group all replicates and separate replicates DEG configurations. Cells marked ``per-dataset'' indicate that the
value differs across datasets, and they are not broken down individually, 
since per-dataset sizes for the upstream / downstream pipelines are already reported in the table.

The GPU table covers the model-training workflows: LPM-style
training on multiple datasets and on L1000 separately, and OP3
benchmarking. For each workflow it reports the number of GPU nodes
used, the per-node CPU and RAM allocation, the size of the input
and output files, and the wall-clock time of training pipelines.
GPU jobs used one GPU per node on a mix of NVIDIA V100, H100, and
A100\,MIG hardware.

Note that the requested allocations (RAM and time limit) are
deliberately higher than the observed peaks: we provision a
safety buffer so jobs are not killed by SLURM for out-of-memory
or wall-time-limit violations.

\begin{table*}[t]
\caption{CPU compute footprint for data processing. DILImap is split into train and train+validation variants. Jobs ran on nodes with Intel Xeon Gold, Intel Xeon Platinum, or AMD EPYC CPUs. ``DEG (group / sep)'' rows show values for the group-all-replicates / separate-replicates DEG configurations. ``per-dataset''~=~varies by dataset.}
\label{tab:CPU_resources}
\centering
\scriptsize
\setlength{\tabcolsep}{3pt}
\begin{tabular}{@{}l l c c c c c c c c@{}}
\toprule
         &         &       & \multicolumn{2}{c}{Per node} & Peak RSS & \multicolumn{2}{c}{Files (GB)} & \multicolumn{2}{c}{Time (h)} \\
\cmidrule(lr){4-5} \cmidrule(lr){7-8} \cmidrule(lr){9-10}
Workflow & Dataset & \# nodes& CPUs & RAM (GB) & (GB) & In & Out & Lim & Max \\
\midrule
Prepare pseudo-/bulk    & CIGS-MCE            & 1   & 4      & 500      & 11.5         & 0     & 1.8        & 10  & 2.8\\
                          & CIGS-TCM            & 1   & 4      & 500      & 1.6          & 0     & 0.6        & 10  & 0.5\\
                          & DILImap train       & 1   & 2      & 100      & 1.8          & 0     & 0.7        & 2   & 0.2\\
                          & DILImap train\_val  & 1   & 2      & 150      & 2.1          & 0     & 0.8        & 2   & 0.2\\
                          & GDPx2               & 1   & 2      & 250      & 4.1          & 0     & 0.8        & 8   & 0.1\\
                          & L1000 Phase I       & 1   & 2      & 500      & 16.3         & 0     & 112.0& 10  & 1.1\\
                          & L1000 Phase II      & 1   & 2      & 250      & 3.0          & 0     & 30.0& 10  & 0.7\\
                          & Novartis/DRUG-seq   & 1   & 4      & 500      & 87.8         & 0     & 33.0& 10  & 2.2\\
                          & OP3                 & 1   & 2      & 100      & 0.6          & 0     & 0.2        & 5   & 0.4\\
                          & sci-Plex            & 1   & 2      & 100      & 25.8         & 0     & 2.7        & 1   & 0.4\\
                          & Tahoe-100M          & 14  & 2      & 700      & 616.0& 0     & 336.0& 16  & 2.4\\
                          & VCPI-0001           & 1   & 2      & 250      & 45.0         & 0     & 11.1       & 8   & 1.3\\
                          & VCPI-0002           & 1   & 2      & 250      & 22.0         & 0     & 7.3        & 8   & 1.0\\
\addlinespace
Process pseudobulk        & --                  & 1   & 2      & 500      & 82.0& per‑-dataset  & per-‑dataset       & 2   & 1.3\\
\addlinespace
DEG (group / sep)         & CIGS-MCE            & 26  & 2 / 2  & 50 / 50  & 1.6 / 3.6& 0.7   & 6.6 / 18.0& 10  & 0.8 / 6.3\\
                          & CIGS-TCM            & 10  & 2 / 2  & 50 / 50  & 1.3 / 3.0& 0.2   & 2.0 / 5.2& 10  & 0.5 / 5.1\\
                          & DILImap train       & 2   & 2 / 10 & 50 / 100 & 6.9 / 20.4& 0.4   & 4.6 / 13.0& 48  & 3.6 / 34.9\\
                          & DILImap train\_val  & 2   & 2 / 10 & 50 / 100 & 7.3 / 21.4& 0.5   & 4.7 / 13.0& 48  & 3.4 / 27.0\\
                          & GDPx2               & 12  & 2 / 2  & 50 / 50  & 1.5 / 5.2& 0.7   & 3.0 / 14.0& 3   & 0.1 / 0.5\\
                          & L1000 Phase I       & 310 & 2 / 2  & 50 / 50  & 1.2 / 2.8& 23.0& 21.0 / 63.0& 10  & 0.1 / 5.5\\
                          & L1000 Phase II      & 142 & 2 / 2  & 50 / 50  & 1.0 / 2.7& 5.2   & 11.0 / 28.0& 10  & 0.1 / 3.5\\
                          & Novartis/DRUG-seq   & 22  & 2 / 10 & 50 / 100 & 5.3 / 17.7& 3.1   & 31.0 / 105& 48  & 5.7 / 27.0\\
                          & OP3                 & 4   & 2 / 2  & 50 / 50  & 0.7 / 1.4& 0.04  & 0.4 / 1.1  & 3   & 0.02 / 0.05   \\
                          & sci-Plex            & 3   & 2 / 2  & 50 / 50  & 5.2 / 8.2& 0.3   & 3.4 / 5.6  & 6   & 0.7 / 2.5\\
                          & Tahoe-100M          & 50  & 2 / 2  & 50 / 50  & 15.5 / 16.6  & 8.6   & 181.0 / 204.0& 10  & 8.8 / 6.5\\
                          & VCPI-0001           & 11  & 2 / 10 & 50 / 100 & 9.8 / 17.0& 2.8   & 26.0 / 49.0& 24  & 8.6 / 23.5\\
                          & VCPI-0002           & 8   & 2 / 10 & 50 / 100 & 7.8 / 15.7& 1.9   & 18.0 / 31.0& 24  & 7.1 / 20.7\\
\addlinespace
DEG aggregation           & --                  & 1   & 2      & 500      & 166.0& per-‑dataset  & per-‑dataset       & 2   & 1.6\\
Enrich pseudobulk .var    & --                  & 1   & 4      & 250      & 174.0& per‑-dataset  & per-‑dataset       & 12  & 0.8\\
\bottomrule
\end{tabular}

\end{table*}

\begin{table*}[t]
\caption{GPU compute footprint for OP3 evaluation. One GPU is used per node in every configuration. LPM/OP3-based training workflows ran on a mix of NVIDIA V100, H100, and A100\,MIG GPUs.}
\label{tab:GPU_resources}
\centering
\footnotesize
\setlength{\tabcolsep}{3pt}
\begin{tabular}{@{}l c c c c c c@{}}
\toprule
         &       & \multicolumn{2}{c}{Per node} & \multicolumn{2}{c}{Files (GB)} & Wall \\
\cmidrule(lr){3-4} \cmidrule(lr){5-6}
Workflow & \# nodes & CPUs & RAM (GB) & In & Out & time (h) \\
\midrule
LPM-based, multi-dataset               & 16 & 8 & 64 & 26.0   & 3.2 & 21.0  \\
LPM-based, L1000                      & 16 & 8 & 64 & 2.7  & 1.0 & 2.5 \\
OP3, per embedding, 10 seeds & 10 & 4 & 64 & 0.3  & 0.5 & 0.7 \\
\bottomrule
\end{tabular}

\end{table*}

The expanded molecule-holdout multi-output LPM experiments used single-node
H100 80GB jobs rather than the earlier 16-node distributed setup.   CPU RAM is the maximum
resident set size recorded by SLURM for the batch step.

\begin{table*}[t]
\caption{Observed compute footprint for molecule-holdout
multi-output LPM experiments.}
\label{tab:expanded_lpm_training_resources}
\centering
\scriptsize
\setlength{\tabcolsep}{3pt}
\resizebox{\textwidth}{!}{%
\begin{tabular}{lrrrrr}

\toprule

Workflow

& Completed jobs

& Total elapsed job-hours

& GPU-hours

& CPU-hours

& Peak CPU RAM \\

\midrule

All-data source model

& 10

& 17.36

& 34.71

& 555.4

& 103.3 GB \\

All-data source model, fixed Morgan molecule embeddings

& 10

& 22.04

& 44.07

& 705.1

& 94.9 GB \\

All-data source model, Morgan-initialized learnable molecule embeddings

& 10

& 17.27

& 34.53

& 552.5

& 93.0 GB \\

Target fine-tuning from all-data source, frozen molecule embeddings

& 120

& 30.04

& 50.00

& 961.2

& 84.3 GB \\

Target fine-tuning from fixed-Morgan source

& 120

& 30.79

& 51.60

& 985.1

& 84.8 GB \\

Target fine-tuning from Morgan-initialized source, frozen updated molecule embeddings

& 120

& 28.89

& 48.51

& 924.5

& 83.8 GB \\

Target-only scratch training

& 120

& 31.23

& 52.84

& 999.4

& 83.5 GB \\

Target-only scratch training with fixed Morgan molecule embeddings

& 120

& 42.10

& 74.85

& 1{,}347.3

& 84.5 GB \\

\textbf{All training jobs}

& \textbf{630}

& \textbf{219.71}

& \textbf{391.11}

& \textbf{7{,}030.6}

& \textbf{103.3 GB} \\

\bottomrule

\end{tabular}%

}

\end{table*}


\clearpage

\end{document}